\documentclass{adobe_research}

\usepackage[utf8]{inputenc}
\usepackage{CJKutf8}
\usepackage{amsmath}
\usepackage{floatrow}
\usepackage{array}
\usepackage{makecell}
\usepackage{threeparttable}
\usepackage{pifont}
\usepackage{amssymb}
\usepackage{xspace}
\usepackage{tikz}
\usetikzlibrary{arrows.meta,shapes.geometric}
\usepackage{enumitem}
\graphicspath{{data/}}

\newcommand{\twvp}{\texttt{Anchor-CoT}\xspace}
\newcommand{\docount}{\texttt{DocCount}\xspace}
\def\eg{\emph{e.g}.}
\def\etc{\emph{etc}}

\definecolor{darkblue}{rgb}{0.0, 0.0, 0.55}
\hypersetup{colorlinks=true, citecolor=darkblue, linkcolor=darkblue, urlcolor=darkblue}
\newcommand{\adopd}{\texttt{ADOPD}\xspace}
\newcommand{\adopdf}{\texttt{ADOPD 2024}\xspace}
\newcommand{\adopds}{\texttt{ADOPD 2026}\xspace}
\newcommand{\doctobox}{\mbox{{Doc2Box}}\xspace}
\newcommand{\doctomask}{\mbox{{Doc2Mask}}\xspace}
\newcommand{\doctotag}{\mbox{{Doc2Tag}}\xspace}

\title{\textcolor{adobered}{Thinking with Anchors:} Grounded and Efficient Document Reasoning}
\author[1,*]{Sichen Zhu}
\author[1,*]{Yuchen Zhu}
\author[2]{Wenzhuo Xu}
\author[3]{Jason Kuen}
\author[3]{Wanrong Zhu}
\author[3]{Jing Shi}
\author[4]{Xuan Shen}
\author[5]{Quanyi Wang}
\author[6]{Yiwei Wang}
\author[6]{Yujun Cai}
\author[7]{Bing Shuai}
\author[7]{Qin Zhang}
\author[1]{Yongxin Chen}
\author[8]{Shilong Liu}
\author[1,\dagger]{Molei Tao}
\author[3,\dagger]{Jiuxiang Gu}
\affiliation[1]{Georgia Tech}
\affiliation[2]{CMU}
\affiliation[3]{Adobe}
\affiliation[4]{ZJU}
\affiliation[5]{NUIST}
\affiliation[6]{SEU}
\affiliation[7]{Physion Labs}
\affiliation[8]{Columbia University}
\contribution[*]{Core contributors}
\contribution[\dagger]{Project lead}

\newcommand{\paperabstract}{%
Existing document understanding benchmarks have largely focused on locating page elements, yet real-world document intelligence requires models to reason jointly about region semantics, spatial relations, and visual structure.
We present \adopds, a reasoning-oriented extension of \adopd that turns page decomposition into spatially grounded document understanding.
\adopds enriches page anchors inherited from \adopdf dataset with human-cleaned captions, semantic tags, and generated chain-of-thought (CoT) traces grounded to document regions.
Instead of treating boxes, masks, and tags as independent supervision signals, we cast text blocks, visual entities, semantic labels, bounding boxes, and polygon masks as a shared vocabulary of \emph{visual anchors}.
This representation supports three connected capabilities.
First, region-level semantic tagging asks models to identify document element types from both page context and local appearance, revealing long-tail semantic failures that standard layout benchmarks often hide.
Second, unified vision-language grounding generates text regions and visual entities together with coordinates or polygonal outlines, transforming detection and segmentation outputs into structured anchors that can be reused by downstream reasoning systems.
Third, current state-of-the-art models still struggle with dense counting tasks evaluated on \docount, a benchmark derived from \adopds, highlighting the need for the Thinking-with-Anchors pipeline in document semantic understanding.
By connecting page decomposition to verifiable visual-anchor reasoning, \adopds provides a task framework that moves document understanding beyond localization toward anchor-grounded document intelligence.
}
\abstract{\paperabstract}

\date{August 1st, 2026}
\adobedata[Project Page]{\href{http://sichenzhu.github.io/thinking-with-anchors/}{sichenzhu.github.io/thinking-with-anchors/}}
\adobedata[Github]{\href{http://github.com/SichenZhu/ADOPD2026}{github.com/SichenZhu/ADOPD2026} | \href{http://github.com/SichenZhu/DocCount}{github.com/SichenZhu/DocCount}}
\adobedata[HuggingFace]{\href{http://huggingface.co/collections/adopd/thinking-with-anchors}{huggingface.co/collections/adopd/thinking-with-anchors}}

\begin{document}
\begin{CJK*}{UTF8}{gbsn}
\maketitle

\vspace{-1em}
\section{Introduction}
Appearing as posters, menus, advertisements, infographics, etc., documents are a primary interface for real-world information exchange. Automating document understanding has motivated research on document analysis and OCR systems~\citep{chaudhuri2007digital,smith2007overview}, as well as real-world applications ranging from information extraction and accessibility to business-process automation, document question answering, and multimodal assistants~\citep{huang2022layoutlmv3,kim2021donut,mathew2021docvqa,Mathur_2023_AAAI}.
Recent datasets and benchmarks have enabled important progress in layout analysis and document understanding~\citep{zhong2019publaynet,pfitzmann2022doclaynet,Cheng_2023_CVPR}, largely by defining page elements that can be detected, segmented, or parsed.
However, many downstream uses require more than a list of localized regions: a system may need to associate a price with the correct product block, connect a chart to its legend and caption, distinguish decorative regions from informative content, or justify an answer by pointing to specific visual evidence on the page.
In these settings, page elements must function as grounded anchors that carry semantic roles, spatial intent, and meaningful relations to neighboring regions.
This perspective exposes the gap between page-decomposition-centered supervision and the evidence-oriented reasoning required by practical document AI.

Current methods only partially fulfill this requirement.
Specialized document parsers, detectors, and segmenters can localize text blocks or visual entities, but their outputs are often treated as coarse page-decomposition results rather than as grounded evidence with semantic roles and cross-region spatial relations.
Open-vocabulary detectors and unified generative models make grounding more flexible via conditioning on language or emitting boxes and polygons as sequences~\citep{groundingdino2024,pix2seq2022,florence22024,kosmos22023}. 
However, such practice is vulnerable when applied to dense document pages: when every text block, entity, and polygon vertex is serialized autoregressively, latency and formatting errors increase with the number of input anchors.
Recent work on multi-token prediction, multi-head decoding, masked generation, and diffusion language models suggests a complementary direction in which dense outputs can be predicted in larger parallel chunks while preserving language-level control~\citep{mtp2024,medusa2024,maskgit2022,llada2025,dream2025}. This provides a brand-new angle to treat the visual grounding task as both a fine-grounding and a semantic reasoning problem.

Motivated by the aforementioned gap and new perspectives on visual grounding, we build \adopds to turn dense document annotations into a testbed for grounded document reasoning. Starting from the original \adopdf dataset~\citep{adopd2024}, which contains 120k visually diverse document images across more than 1000 document types, we enrich the inherited page element annotations with human-cleaned captions, semantic tags, and chain-of-thought (CoT) data. We organize OCR text boxes, human-labeled entity polygons, semantic labels, and polygon outlines around a shared abstraction: the \emph{visual anchor}.

A \emph{visual anchor} is a document region represented not only by its geometry but also by its semantic role, functions, and potential relations to other regions on the page. This abstraction builds on prior work that reasons over visual primitives~\citep{lu2026think}.
By proposing a visually anchored dataset and an additional anchor-grounded evaluation benchmark, our goal is to study how vision-language models (VLMs) can go beyond detection and segmentation to ground, tag, compose, and verify document anchors as intermediate evidence for reasoning.

This paper develops \adopdf from the visual-anchor point of view along the following directions.
First, we collect visual anchors over \adopds: human-cleaned captions, semantic tags for entity polygons, and OCR text boxes. We also generated CoT traces grounded to those visual anchors (Sec.~\ref{sec:data_collection}).
Second, we evaluate document localization on collected visual anchors (Sec.~\ref{sec:decomposition_experiments}), focusing on conventional detection/segmentation tasks.
Third, we evaluate entity-level semantic tagging (Sec.~\ref{sec:tagging_experiments}), where models must use both local appearance and full-page context to assign semantic roles to localized regions. %
Fourth, we derived \docount from \adopds, a dense document-counting benchmark with Thinking-with-Anchors reasoning traces, and evaluated state-of-the-art VLMs in the zero-shot setting (Sec.~\ref{sec:reason_corpus}).

In summary, \adopds reframes document page decomposition as grounded reasoning over visual anchors.
By connecting semantic region tagging, unified grounding of text and visual entities, and verified grounded narration, it studies how document models can produce structured visual evidence that is spatially precise, semantically meaningful, and usable by downstream reasoning systems.
This shifts \adopd from a dataset for decomposing document pages into a benchmark for building, evaluating, and diagnosing grounded document reasoning.

\vspace{-0.5em}
\section{Background}
\label{sec:related}
\vspace{-0.5em}
\noindent\textbf{From pages to evidence.} Document understanding has long combined page segmentation, OCR, layout analysis, and task-specific extraction. Early systems used rule-based grouping, projection profiles, connected components, or handcrafted geometry for page segmentation and text-line extraction~\citep{ouwayed2012general,lee2019page}, while modern datasets such as PubLayNet~\citep{zhong2019publaynet}, DocBank~\citep{li2006docbank}, DocLayNet~\citep{pfitzmann2022doclaynet}, M$^6$Doc~\citep{Cheng_2023_CVPR}, IIIT-AR-13K~\citep{mondal2020iiit}, and the original \adopdf page-decomposition dataset~\citep{adopd2024} support learning-based layout detection and segmentation. Document-language pretraining and OCR-free models further brings together text, layout, images for parsing and extraction, and question answering~\citep{huang2022layoutlmv3,selfdoc2021,gu2021unidoc,tang2023unifying,kim2021donut,mathew2021docvqa,Mathur_2023_AAAI}. These works define important tasks, but they often evaluate either low-level decomposition or final answers without specifying the intermediate evidence layer. \adopds addresses this gap by making boxes, polygon masks, and semantic tags explicit visual anchors that can be evaluated as structured evidence for reasoning.

\noindent\textbf{Grounding as an interface.} Large vision-language models bring language-level reasoning to images through visual instruction tuning, multimodal chain-of-thought, and grounded generation~\citep{llava2023,qwenvl2023,multimodalcot2023,kosmos22023,shikra2023,ferret2024,florence22024,unifiedio22024}. Their interface is attractive for documents because region names, coordinates, counts, spatial order, and justifications can all be represented in one symbolic output space. 
Recent work has also used visual counting as a diagnostic of grounding and reasoning in MLLMs. HoloCount~\citep{deng2026holocount} introduces semantic counting, analytical counting, and robustness testing tasks. It reveals substantial degradation from basic perception to analytical and robustness settings, underscoring that fluent multimodal responses do not guarantee numerically reliable visual grounding. HoloCount primarily works in the natural image domain and evaluates the final count using exact-match accuracy. \adopds is complementary in the document domain and exposes the intermediate boxes, masks, and semantic tags that support such outputs and can be evaluated independently. 
Work on thinking with those visual anchors makes this idea more explicit by inserting spatial markers into reasoning, reducing the ambiguity of purely textual references. We derived a dense document counting benchmark, \docount, from the \adopds dataset. \docount inherits the benefit of rich anchors and executable checks that tie reasoning back to verifiable visual evidence.

\noindent\textbf{Efficient anchor generation}.The cost of autoregressive generation of visual anchors such as vertex coordinates motivates efficient alternatives on both the language and vision sides. In language modeling, multi-token prediction and multi-head decoding reduce sequential steps~\citep{mtp2024,medusa2024}, while masked, diffusion, and block-diffusion models replace strict left-to-right decoding with iterative or blockwise refinement~\citep{maskgit2022,llada2025,dream2025,arriola2025bd3lm,cheng2025sdar,wu2025fastdllmv2}; \textsc{FLARE}~\citep{zhu2026flare} extends this direction to hybrid language-model backbones. In vision, YOLO-style detectors~\citep{ge2021yolox}, DETR/RF-DETR~\citep{carion2020end,rfdetr2024}, GroundingDINO~\citep{groundingdino2024}, and SAM-style segmenters~\citep{kirillov2023segany,ravi2024sam2} already produce boxes and masks through highly parallel computation with usable confidence scores. These models are strong anchor producers, but efficiency alone does not decide which regions belong together, which are semantically meaningful, or how they support an answer. Our framework treats efficient generation and perception as complementary to reasoning: fast models provide dense candidates, while \adopds evaluates how those candidates become typed, grounded evidence.

\noindent\textbf{Composable perception and reasoning.} Agentic visual systems connect perception modules with language-level decisions by calling tools, writing programs, or reasoning over marked inputs. VISPROG~\citep{visprog2023} and ViperGPT~\citep{vipergpt2023} compose vision modules through executable programs, MM-ReAct~\citep{mmreact2023} routes a language model through vision experts, and Set-of-Mark prompting~\citep{setofmark2023} lets a multimodal model refer to numbered visual evidence without directly regressing coordinates. These systems show that visual reasoning can be separated from low-level perception, but they are usually evaluated on natural images or general multimodal tasks with fewer regions and weaker requirements on calibrated grounding. In one downstream experiment, we proposed an agentic self-refinement workflow that studies agentic grouping over detector candidates, and our grounded verifier checks whether the resulting reasoning remains spatially precise and semantically accountable.

\section{Data Construction}
\vspace{-0.8em}
\label{sec:data_collection}

\begin{figure*}[!h]
\centering
\vspace{-0.5em}
\includegraphics[width=\linewidth]{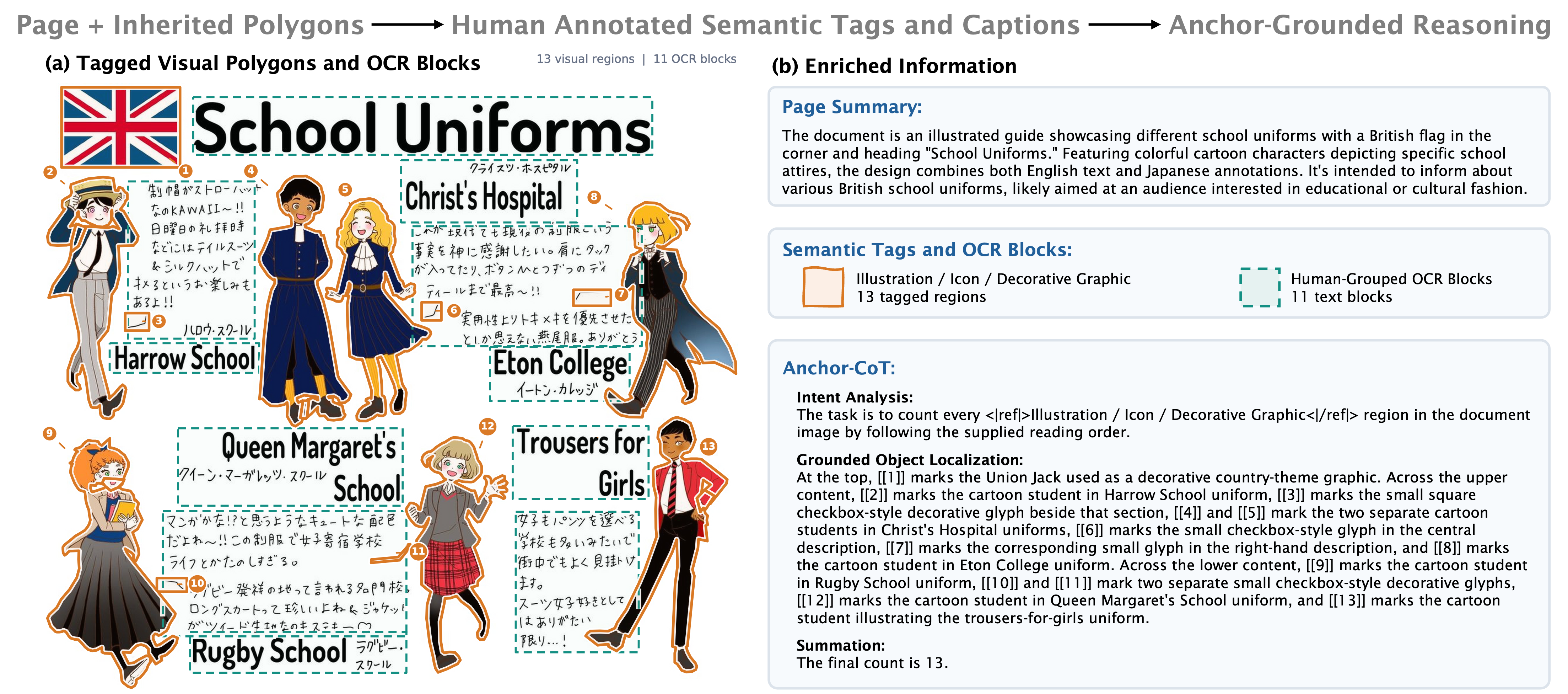}
\vspace{-1em}
\caption{Overview of the \adopds data enrichment. Starting from inherited \adopdf page anchors, the new data round adds human-cleaned captions, semantic tags over regions, and generated CoT traces whose answers are grounded to visual anchors such as boxes and polygons.}
\vspace{-1em}
\label{fig:data_enrichment_overview}
\end{figure*}

The original \adopdf corpus~\citep{adopd2024} contains dense page geometry for ${\sim}$120k visually diverse documents, including human-drawn entity polygons and OCR text blocks. \adopds builds on top of these inherited anchors by adding three additional layers: human-cleaned captions, fine-grained semantic tags, and generated CoT traces grounded to visual anchors (Figure~\ref{fig:data_enrichment_overview}).

\providecommand{\tableyes}{\ding{51}}
\providecommand{\tablepartial}{\ensuremath{\triangle}}
\providecommand{\tableno}{\textemdash}

\providecommand{\benchmarkdatasetcolwidth}{3.75cm}
\providecommand{\benchmarksizecolwidth}{2.15cm}
\providecommand{\benchmarkprovenancecolwidth}{1.3cm}
\providecommand{\benchmarkboxcolwidth}{1.3cm}
\providecommand{\benchmarkpolygoncolwidth}{1.72cm}
\providecommand{\benchmarkocrcolwidth}{1.72cm}
\providecommand{\benchmarksemanticcolwidth}{1.72cm}
\providecommand{\benchmarkspatialcolwidth}{1.72cm}
\providecommand{\benchmarkfeaturecolsep}{3pt}

\begin{table*}[t]
  \centering
  \begin{threeparttable}
    \caption[]{%
      Comparison of document-layout datasets.
      \textbf{Size} is the complete dataset size (training, validation, and test
      combined), reported in the dataset's native unit: pages, images, documents
      (docs), or annotated block instances (inst.).
      \textbf{Provenance} describes how the annotations were produced:
      \textbf{H} denotes manual annotation, \textbf{HV} human verification or
      correction, \textbf{A} automatic or weak-supervision annotation, and
      \textbf{S} synthetic generation.
      \textbf{Box} denotes native axis-aligned boxes, rotated boxes, or
      polygons with at most four vertices.
      \textbf{Arbitrary polygon} denotes native per-instance boundaries with
      more than four vertices. Converting a box into a four-point polygon does
      not count.
      \textbf{Region-linked OCR} denotes transcription explicitly associated
      with a localized region instance.
      \textbf{Semantic class} denotes a semantic category assigned to a
      localized region.
      \textbf{Grounded spatial tasks} denotes released annotations requiring
      localization or spatial/relational reasoning over identified regions.
      Coordinates alone do not count.
      \tableyes{} indicates explicit native support, \tablepartial{} indicates
      subset-only, indirect, inherited, or structurally related support, and
      \tableno{} indicates that the feature is absent.}
    \label{tab:selected-document-benchmarks}

    \scriptsize
    \setlength{\tabcolsep}{\benchmarkfeaturecolsep}
    \renewcommand{\arraystretch}{1.12}
    \begin{tabular}{@{}>{\raggedright\arraybackslash}p{\benchmarkdatasetcolwidth}@{}
      >{\centering\arraybackslash}p{\benchmarksizecolwidth}
      >{\centering\arraybackslash}p{\benchmarkprovenancecolwidth}
      >{\centering\arraybackslash}p{\benchmarkboxcolwidth}
      >{\centering\arraybackslash}p{\benchmarkpolygoncolwidth}
      >{\centering\arraybackslash}p{\benchmarkocrcolwidth}
      >{\centering\arraybackslash}p{\benchmarksemanticcolwidth}
      >{\centering\arraybackslash}p{\benchmarkspatialcolwidth}@{}}
      \toprule
      Dataset
      & Size
      & \rotatebox{0}{\makecell[c]{Provenance}}
      & \rotatebox{0}{\makecell[r]{Box}}
      & \rotatebox{0}{\makecell[c]{Arbitrary\\polygon}}
      & \rotatebox{0}{\makecell[c]{Region-linked\\OCR}}
      & \rotatebox{0}{\makecell[c]{Semantic\\class}}
      & \rotatebox{0}{\makecell[c]{Grounded\\spatial tasks}} \\
      \midrule
      PubLayNet~\cite{publaynet}
        & 358,353 pages & A & \tableyes & \tablepartial & \tableno & \tableyes & \tableno \\
      DocBank~\cite{docbank}
        & 500K pages & A & \tableyes & \tableno & \tableyes & \tableyes & \tableno \\
      IIIT-AR-13K~\cite{iiitar13k}
        & 13K pages & H & \tableyes & \tableno & \tableno & \tableyes & \tableno \\
      DocLayNet~\cite{doclaynet}
        & 80,863 pages & H & \tableyes & \tableno & \tablepartial & \tableyes & \tableno \\
      M\textsuperscript{6}Doc~\cite{m6doc}
        & 9,080 pages & H & \tableyes & \tableyes & \tableno & \tableyes & \tableno \\
      DocGenome~\cite{docgenome}
        & 500K docs & A+HV & \tableyes & \tableno & \tableyes & \tableyes & \tableyes \\
      PALdb~\cite{paldb}
        & 441K pages & A+HV & \tableyes & \tableno & \tableyes & \tableyes & \tableno \\
      {Diachronic Doc}~\cite{diachronicdocument}
        & 7,254 pages & H & \tableyes & \tableno & \tableno & \tableyes & \tableno \\
      DocSynth-300K~\cite{doclayoutyolo}
        & 300K images & S & \tableyes & \tableno & \tableno & \tableyes & \tableno \\
      DocStructBench~\cite{doclayoutyolo}
        & 11,314 images & H & \tableyes & \tableno & \tableno & \tableyes & \tableno \\
      GraphDoc~\cite{graphdoc}
        & 80K images & H+A & \tableyes & \tableno & \tableyes & \tableyes & \tableyes \\
      IndicDLP~\cite{indicdlp}
        & 119,806 images & H+HV & \tableyes & \tableno & \tableno & \tableyes & \tableno \\
      MonkeyDoc~\cite{monkeyocr}
        & 3.9M inst. & H+A+S & \tableyes & \tableno & \tableyes & \tableyes & \tablepartial \\
      SCAN~\cite{scan}
        & 24,577 pages & H & \tableyes & \tableno & \tableno & \tableyes & \tableno \\
      DocAtlas~\cite{docatlas}
        & 365,862 pages & A+S & \tableyes & \tableno & \tableyes & \tableyes & \tablepartial \\
      \midrule
      \adopds (ours)
        & 120K pages & H+HV & \tableyes & \tableyes & \tableyes & \tableyes & \tableyes \\
      \bottomrule
    \end{tabular}

  \end{threeparttable}
\end{table*}

\vspace{-0.8em}
\subsection{Entity-Level Semantic Tagging}
\label{sec:data_tagging}
\vspace{-0.2em}
Human-drawn entity polygons and OCR text-block locations from the original \adopdf dataset gives each page a dense geometric decomposition. These anchors are useful for document understanding, but they are not sufficient for high-level reasoning: a model may know where a region is without knowing whether it functions as a photograph, a brand logo, a background image, a decorative motif, or a text role such as a title or caption. Clean page-level captions are also critical. In \adopds, we therefore first rewrite the original captions with human annotation, asking annotators to summarize the document type, visual style, salient content, and high-level intent so that the global description is better aligned with human interpretation.

However, global caption alone remains too coarse for fine-grained reasoning over page elements. We introduce entity-level semantic tagging on top of the visual anchors. The annotation unit is an existing entity polygon or OCR text block, and annotators assign a semantic role only after reading and understanding the full document context. For visual masks, this includes judging whether the region belongs to the foreground or background and then assigning a document-specific tag such as photograph, illustration, brand logo, background image, chart, icon, or color block. For OCR blocks, the tag captures text function, such as title, body text, header, footer, or image caption. Each tagged anchor thus becomes a reusable, semantically meaningful evidence unit of the form \textit{geometry} + \textit{modality} + \textit{semantic role}.

This setting differs from conventional tagging tasks where the label space is usually clearer in advance, such as NLP sequence tagging and named-entity recognition~\citep{tjong2003conll}, or object-centric vision taxonomies in PASCAL VOC and MS COCO~\citep{DBLP:journals/ijcv/EveringhamGWWZ10,DBLP:conf/eccv/LinMBHPRDZ14}. Document elements are shaped by layout, design intent, reading order, and page context, so their semantic roles cannot be reduced to a fixed object list. We therefore define the 30-class tagging taxonomy through a human-in-the-loop process: candidate tags are proposed from the data, annotators apply them to real pages, and ambiguous boundaries are refined through review. Despite such effort, document tagging is inevitably long-tailed and subjective. Annotators may differ in design knowledge, visual judgment, or interpretation of tags. Thus, we further apply disambiguation rules and post-processing checks. The complete taxonomy, foreground/background layer definition, disambiguation rules, worked examples, and resulting long-tail tag distribution are provided in Appendix~\ref{sec:annotation_guidelines}. The comparison between \adopds and other existing document datasets is shown in Table.~\ref{tab:selected-document-benchmarks}.

\vspace{-0.3em}
\subsection{Thinking with Anchors}
\label{sec:data_twp}
The final enrichment layer turns the tagged visual anchors of Sec.~\ref{sec:data_tagging} into grounded, step-by-step reasoning traces, so that each page is paired not only with \emph{where} and \emph{what} its regions are but also with \emph{how} to reason over them. We instantiate this layer on dense counting, a task that remains challenging for multimodal large language models precisely because it requires grounding and semantic understanding at once rather than a single free-form answer. Each example starts from a document page plus polygon masks with human-annotated semantic classes, such as \texttt{Photograph}, \texttt{Table}, \texttt{Brand Logo}, etc. Following the Thinking-with-Visual-Primitives paradigm \citep{lu2026think}, we construct counting questions whose answers require grounding related regions on the document, rather than relying solely on free-form caption knowledge. Class labels are further selected during data preparation. A VLM verifier is applied to validate the labels of the polygon masks, followed by a manual human check. The CoT generator is prompted to write the natural-language reasoning for each grounded anchors in the samples, providing us with a CoT-style supervision that is linguistically natural but geometrically controlled as well. This procedure yields \docount, a document-counting benchmark on which we evaluate 13 state-of-the-art VLMs in the zero-shot setting (Sec.~\ref{sec:reason_corpus}). Table.~\ref{tab:docount_benchmark_comparison} shows the comparison between \docount and existing counting-related benchmarks.

\begin{table*}[!h]
\centering
\caption{Comparison of evaluation protocols in representative
visual-counting benchmarks. 
\textbf{\# Samples} reports the official test split in the datasets, or the full released evaluation collection for evaluation-only benchmarks.
\textbf{Counted Object} means the visual target whose instance count contributes to the final answer, before any requested arithmetic calculation that may appear in the query.
\textbf{Granularity} means whether the counted targets are semantically defined or only labeled with their class name. 
\textbf{Grounding} means instance-level spatial ground truth annotations available for objects that contribute to the final answer, including points, boxes, or polygons. 
\textbf{Grounded-reasoning} means whether an explicit reasoning representation exists for every instance-level spatial target in the evaluation collection.
}
\label{tab:docount_benchmark_comparison}
\scriptsize
\setlength{\tabcolsep}{1.5pt}
\renewcommand{\arraystretch}{1.16}
\begin{tabular}{@{}
  >{\raggedright\arraybackslash}p{0.175\textwidth}
  >{\raggedright\arraybackslash}p{0.165\textwidth}
  >{\raggedright\arraybackslash}p{0.085\textwidth}
  >{\raggedright\arraybackslash}p{0.165\textwidth}
  >{\raggedright\arraybackslash}p{0.165\textwidth}
  >{\raggedright\arraybackslash}p{0.145\textwidth}
  >{\raggedright\arraybackslash}p{0.075\textwidth}
  @{}}
\toprule
\textbf{Benchmark} &
\textbf{Image Domain} &
\textbf{\# Samples} &
\textbf{Counted Object} &
\textbf{Granularity} &
\textbf{Grounding} &
\textbf{Grounded-reasoning} \\
\midrule
{ShanghaiTech \citep{shanghaitech}} &
Natural scenes &
498 &
Person &
Fixed class &
Points &
No \\

{NWPU-Crowd \citep{nwpucrowd}} &
Natural scenes &
1,500 &
Person &
Fixed class &
Points, Boxes &
No \\

{JHU-CROWD++ \citep{jhucrowd}} &
Natural crowd scenes &
1,600 &
Person &
Fixed class &
Points &
No \\

{CARPK \citep{carpk}} &
Aerial parking lots &
459 &
Car &
Fixed class &
Boxes &
No \\

{UCF-QNRF \citep{ucfqnrf}} &
Dense natural scenes &
334 &
Person &
Fixed class &
Points &
No \\

{FSC-147 \citep{fsc147}} &
Natural images &
1,190 &
Target objects\textsuperscript{1} &
Exampler boxes\textsuperscript{1} &
Points &
No \\

{CountBench \citep{countbench}} &
General images &
540 &
Target labels &
Class name &
None &
No \\

{PixMo-Count \citep{pixmocount}} &
General images &
540 &
Target labels &
Class name &
None &
No \\

{CountQA \citep{countqa}} &
Natural scenes &
1,528 &
Target labels &
Class name&
None &
No \\

{HoloCount \citep{holocount}} &
General images\textsuperscript{2}&
2,480 &
Target labels &
Class name\textsuperscript{3}&
None &
No \\

\midrule
\docount (ours) &
Document pages &
442 &
Semantic document region &
Semantically defined classes &
Polygon-grounded &
Yes\textsuperscript{4} \\
\bottomrule
\end{tabular}

\vspace{2pt}
\begin{minipage}{\textwidth}
\scriptsize
\textit{Notes.}
\textsuperscript{1}The model is given three examples with bounding boxes and is asked to count all objects visually similar to those examples.
\textsuperscript{2} HoloCount also includes adversarial synthetic images.
\textsuperscript{3} Depending on the task query, HoloCount may include attribute constraints, regions of interest (ROIs), coordinates, exclusions, set operations, and categorical name in the input prompt.
\textsuperscript{4}\docount provides grounded-CoT, namely, \twvp{}, together with the final scalar count. 
\end{minipage}
\end{table*}

\vspace{-0.5em}
\section{Experiments}
\label{sec:experiments_2026}

This section is organized around three experimental threads, based on the same finalized 80k/20k/20k train/validation/test split. The split construction and balance diagnostics are detailed in Appendix~\ref{sec:data_split}. First, we evaluate document decomposition and detection, comparing specialized non-VLM detectors/segmenters with the VLM-based LocateAnything \cite{wang2026locateanything} grounder and an agentic grouping refinement. Second, we study entity-level semantic tagging, in which the goal is to assign taxonomy labels defined in Sec.~\ref{sec:data_collection} to localized document regions. Third, we evaluate VLMs' zero-shot performance on \docount, our document-counting benchmark.

\subsection{Visual Anchor Localization}
\label{sec:decomposition_experiments}

\begin{table}[h]
\centering
\scriptsize
\setlength{\tabcolsep}{4pt}
\caption{Main results for \doctobox detection and \doctomask segmentation tasks.}
\label{tab:adopd2026_main_results}
\begin{tabular*}{\linewidth}{@{\extracolsep{\fill}}l r c c c r r r r r r r@{}}
\toprule
& & \multicolumn{3}{c}{Zero-shot (ZS)} & \multicolumn{6}{c}{Fine-tuned (FT)} & \\
\cmidrule(lr){3-5}\cmidrule(lr){6-11}
Model & \#Par. & AP & AP$_{50}$ & AP$_{75}$ &
AP & AP$_{50}$ & AP$_{75}$ &
AP$_S$ & AP$_M$ & AP$_L$ & mF1 \\
\midrule
\multicolumn{12}{l}{\emph{Box detection}} \\
YOLOv12-M     & 20.2\,M & 0.1 & 0.3 & 0.1 & 69.0 & 82.1 & 75.0 & 30.0 & 55.2 & 75.1 & 77.0 \\
YOLOv12-X     & 59.1\,M & 0.1 & 0.3 & 0.1 & 69.1 & 82.0 & 75.1 & 29.7 & 55.1 & 75.2 & 77.0 \\
RF-DETR-Nano  & 30.5\,M & 0.0 & 0.1 & 0.0 & 50.2 & 72.3 & 53.9 & 9.8 & 31.6 & 58.8 & 69.7 \\
RF-DETR-Large & 33.9\,M & 0.1 & 0.3 & 0.2 & 63.9 & 81.3 & 69.9 & 22.8 & 48.1 & 71.4 & 75.5 \\
LocateAnything-3B & 3.83\,B & -- & -- & -- & 40.2 & 56.8 & 43.4 & 9.3 & 26.3 & 47.8 & 59.4 \\
\midrule
\multicolumn{12}{l}{\emph{Mask segmentation}} \\
YOLOv12-Seg-M    & ${\sim}$23.6\,M & 1.9 & 3.7 & 1.6 & 62.7 & 78.9 & 74.4 & 29.0 & 57.3 & 74.2 & 76.2 \\
YOLOv12-Seg-X    & ${\sim}$62.8\,M & 2.2 & 4.0 & 2.0 & 64.4 & 80.5 & 76.7 & 30.0 & 59.5 & 76.8 & 77.1 \\
RF-DETR-Seg-Nano & 33.6\,M & 1.3 & 2.8 & 0.9 & 52.9 & 71.0 & 58.4 & 3.6 & 26.3 & 63.6 & 70.2 \\
RF-DETR-Seg-2XL  & 38.6\,M & 1.3 & 2.6 & 0.9 & 69.5 & 81.9 & 75.8 & 20.9 & 51.6 & 77.6 & 76.8 \\
SAM3             & 848\,M & 2.3 & 5.0 & 1.7 & 57.2 & 77.7 & 58.0 & 27.7 & 45.4 & 62.0 & 75.5 \\
SAM3.1           & 848\,M & 2.5 & 5.4 & 2.0 & 59.4 & 78.7 & 60.6 & 27.2 & 45.5 & 64.7 & 76.0 \\
LocateAnything-3B-DP & 3.83\,B & -- & -- & -- & 32.5 & 47.4 & 34.0 & 5.7 & 20.3 & 38.0 & 57.1 \\
\bottomrule
\end{tabular*}
\end{table}

\subsubsection{Experimental Setup}
\label{sec:decomposition_setup}

\paragraph{Tasks.}

We evaluate two tag-agnostic page localization tasks based on OCR bounding boxes (\doctobox) and visual entity masks (\doctomask). In \doctobox detection, every text box region is treated as a single class, \textit{entity}. In \doctomask segmentation, every foreground visual entity mask is also treated as an \textit{entity}. Although these localization tasks do not require the model to distinguish semantic region tags, they are not intended to make the task trivial. The overall goal is to measure whether a model can recover the reusable visual anchors that downstream semantic and reasoning tasks depend on, such as OCR boxes and visual entity masks. In documents, failures such as missing text blocks, merged visual regions, or over-split entities can break overall document understanding. Following \adopdf~\citep{adopd2024}, we decouple localization from tagging: \doctobox detection and \doctomask segmentation identify where visual anchors are and whether they are decomposed at the right level of granularity, while \doctotag task evaluates which semantic labels should be assigned to the localized regions. This separation allows us to diagnose geometric decomposition errors separately from semantic tagging errors.

\paragraph{Modeling.}

We evaluate six model families on these two tag-agnostic page localization tasks. The specialized detection and segmentation model families are fine-tuned on the 80k train split. The same models are also evaluated zero-shot when an off-the-shelf checkpoint is available for comparison. The non-VLM baselines include YOLOv12 
for text box detection and YOLOv12-Seg for mask segmentation~\citep{tian2025yolov12}; 
RF-DETR
for text box detection and RF-DETR-Seg for mask segmentation~\citep{carion2020end,rfdetr2024};
and promptable segmentation foundation models SAM3/SAM3.1
~\citep{kirillov2023segany,ravi2024sam2,carion2025sam3}.
We also evaluate LocateAnything-3B, a 3.83B-parameter VLM-based generative grounder that emits boxes and polygonal masks~\citep{wang2026locateanything}. For detailed training and inference setup, please refer to \ref{training_setup_exp1}. 

\paragraph{Evaluation metrics.}

For the localization task, the standard evaluation metrics are COCO-style Average Precision (AP), which averages AP over ten IoU thresholds, and Average Recall (AR) computed on the single \textit{entity} class. In \doctobox detection, we evaluate box AP/AR against annotated OCR blocks. In \doctomask segmentation, we evaluate mask AP/AR against foreground entity masks. %
In addition, we also report AP$_{50}$ (IoU threshold 0.5), AP$_{75}$ (IoU threshold 0.75), AP$_S$ (regions smaller than $32^2  \,  \text{pixel}^2$), AP$_M$ (regions between $32^2$ to $96^2  \,  \text{pixel}^2$), AP$_L$ (regions greater than $96^2 \, \text{pixel}^2$). 
These metrics reward both recall and spatial precision, penalizing the decomposition errors that this experiment thread is designed to expose, such as missing anchors, over-merged regions, over-split boxes, and loose masks. 
Unlike other non-VLM detectors and segmentors, LocateAnything will not output a deterministic confidence score by default. We thus use a pseudo-confidence score simulated from the generation positions of the emitted boxes. Such a pseudo-confidence score is used only for the AP-style metrics, which require detections to be ranked. Although we tried our best to adopt a pseudo-confidence score to align with the requirements of a confidence-score-based metric, this metric may not fairly evaluate confidence-free models such as LocateAnything, especially in cross-model comparisons.
We also include mF1 as an auxiliary diagnostic metric. LocateAnything follows the confidence-free F1@IoU Mean used for generative box outputs~\citep{wang2026locateanything}, whereas non-VLM detectors and segmenters use confidence-swept IoU$=0.5$ F1. Given such model-specific calculation, mF1 should be interpreted within its stated definition rather than as a single, universal ranking score for cross-model comparison.

\subsubsection{Analysis}
\label{sec:decomposition_analysis}

From Table~\ref{tab:adopd2026_main_results}, we can see that \adopds creates a genuine transfer gap between the pretraining image domain and the document domain for segmentation and detection models. Zero-shot evaluations are near failure, while fine-tuning the same architectures lifts AP by $50$--$70$ points. This indicates that natural-image category priors and promptable segmentation priors in existing models' training procedure do not directly transfer to OCR boxes (\doctobox) or visual entity masks (\doctomask) localization task.

\begin{figure*}[!h]
\vspace{-0.5em}
\centering
\begin{subfigure}{0.24\textwidth}\includegraphics[width=\linewidth]{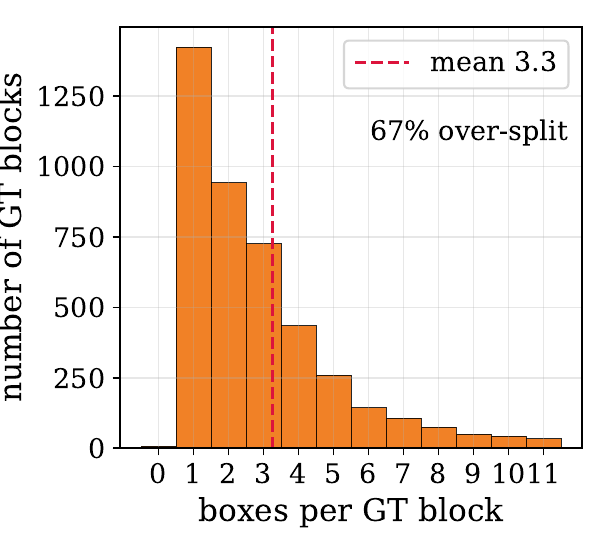}\caption{Boxes per GT block.}\label{fig:agentic_oversplit}\end{subfigure}\hfill
\begin{subfigure}{0.24\textwidth}\includegraphics[width=\linewidth]{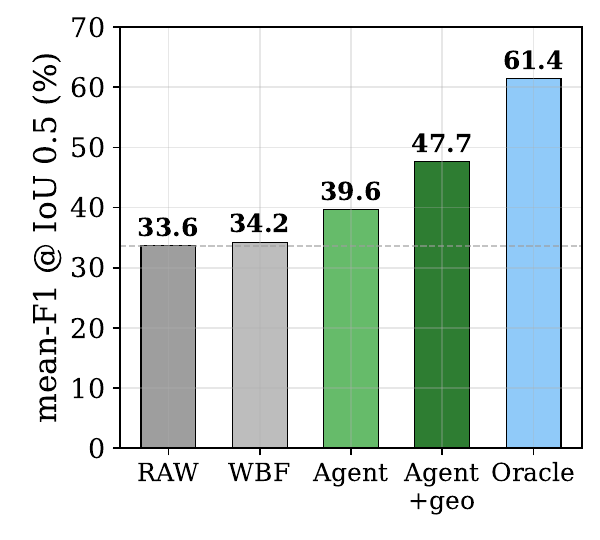}\caption{mean-F1 by method.}\label{fig:agentic_bar}\end{subfigure}\hfill
\begin{subfigure}{0.24\textwidth}\includegraphics[width=\linewidth]{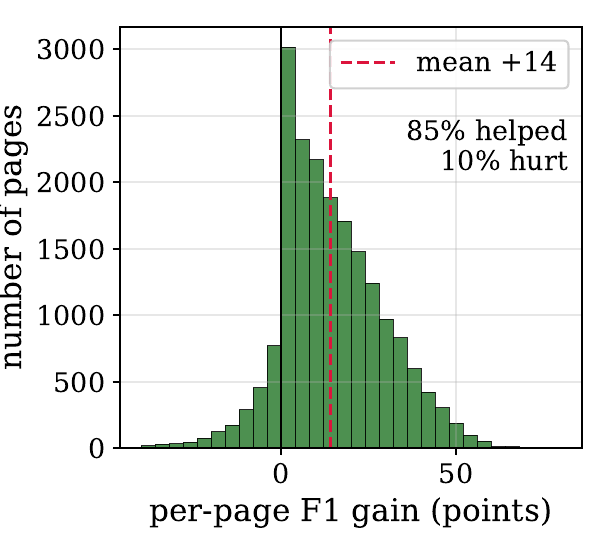}\caption{Per-page gain.}\label{fig:agentic_gainhist}\end{subfigure}\hfill
\begin{subfigure}{0.24\textwidth}\includegraphics[width=\linewidth]{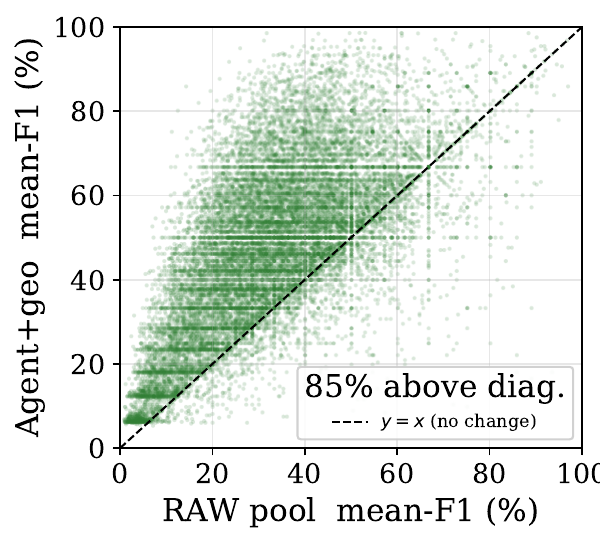}\caption{RAW vs.\ Agent+geo.}\label{fig:agentic_scatter}\end{subfigure}
\caption{Agentic refinement over the 20k-pages validation split for RF-DETR-Large proposed text boxes pool. Panel (a) shows why the agentic workflow is needed: detector boxes per ground-truth block have a long over-split tail. Panel (b) separates the sources of improvement: WBF alone barely changes mean-F1, semantic grouping by agent gives the first lift, and the agent with geometric over-merge guard further improves the mean-F1. Panels (c) and (d) show that the improvement is broad across the validation split rather than driven by a few pages.}
\label{fig:agentic_analysis}
\end{figure*}

The main box-detection error stems from over-fragmentation. Each ground truth
text box block is covered by $3.3$ predicted detector boxes on
average, and $66\%$ receive at least two predicted boxes (Fig.~\ref{fig:agentic_analysis}a). A GT-guided perfect merge can lift
mean-F1 by a large margin (Fig.~\ref{fig:agentic_analysis}b). We therefore treat the detected box pool as proposals and ask a reasoning VLM to group the numbered boxes only through an agentic workflow. Fig.~\ref{fig:agentic_workflow} shows the agentic workflow.

\begin{figure}[!h]
\centering
\scriptsize
\begin{tcolorbox}[
  title={Agentic grouping and geometric validation of predicted text boxes},
  colback=blue!3,
  colframe=blue!35!black,
  colbacktitle=blue!10,
  coltitle=black,
  boxrule=0.5pt,
  arc=1mm,
  left=4pt,
  right=4pt,
  top=3pt,
  bottom=3pt,
  fonttitle=\bfseries
]

\textbf{Visual input.} Page image with fused detector boxes drawn in color and numbered $0,\ldots,N{-}1$ ($N\le30$).

\vspace{2pt}
\textbf{User prompt.} \emph{The image is a document page with detected boxes drawn and numbered. Detectors often over-split one real text block into several boxes. Group the numbered boxes so each group is one text block a human would annotate together. Use visual layout: same column, contiguous lines/items, shared paragraph/list/table context. Do not invent boxes. Output all groups as a JSON list of lists; leave unrelated boxes as singletons.}

\vspace{2pt}
\textbf{Reason-then-parse protocol.}

\hspace{0.4em}\textbf{Reason over marks.} Send the marked page and prompt; stop generation at \texttt{</think>} and save the bounded reasoning trace $t$.

\hspace{0.4em}\textbf{Parse groups.} Reuse the same marked page and prefill the assistant with $t$ followed by \texttt{FINAL: [[}, forcing JSON continuation, e.g.\ \texttt{[[1,2,3],[4],[5,6]]}.

\vspace{2pt}
\textbf{Post-check.} Each returned group is merged to one union box only if a deterministic geometric guard accepts it; otherwise the member boxes are kept separate.
\end{tcolorbox}
\caption{Prompted agent workflow for self-refinement. The VLM reasons over numbered detector boxes and then reuses the saved trace to produce parseable JSON groups; no coordinate regression is performed.}
\label{fig:agentic_workflow}
\end{figure}

This agentic refinement stage requires no additional training and does not use ground-truth annotations when deciding which boxes to merge. RF-DETR and YOLOv12 are run with a low confidence threshold of (0.05) to retain a high-recall collection of candidate text boxes. Overlapping predictions from the two detectors are combined using Weighted Box Fusion (WBF), remaining duplicates are removed by Non-Maximum Suppression (NMS) at IoU (0.6). The remaining boxes are drawn on the original page and assigned integer identifiers. Keye-VL-2.0-30B \cite{team2026kwai} is asked to examine the numbered candidate boxes and, based on the page’s visual layout and text organization, explain which boxes likely belong to the same coherent text block (Fig.~\ref{fig:agentic_workflow} \textbf{Reason over marks}). The explanation is then included in a second model request, prompting the model to complete only a final list of box-identifier groups (Fig.~\ref{fig:agentic_workflow} \textbf{Parse groups}). For each proposed group, we compute the smallest enclosing rectangle of its member boxes rather than directly predicting coordinates. A deterministic geometric check rejects groups whose enclosing rectangle contains excessive empty space, covers most of the page, or contains more than 20 boxes. When a group is rejected, its original boxes are retained separately instead of being merged (Fig.~\ref{fig:agentic_workflow} \textbf{Post-check}).
The full algorithm and numeric table are in Appendix~\ref{sec:agentic_algorithm} and Appendix Table~\ref{tab:agentic_refine}.

The agentic workflow recovers a large fraction of ground truth text boxes without any additional training. 
On the same fixed pool of predicted text boxes (RF-DETR-Large, denoted as RAW), geometry-based fusion (WBF) increases mean-F1 by $0.6$, agent grouping raises the performance by $6.0$ ($33.6$ to $39.6$), and the agentic workflow with geometric over-merge guard reaches $47.7$, closing $49\%$ of the $61.4$ merge-oracle ceiling (Fig.~\ref{fig:agentic_analysis}b). Representative qualitative examples are shown in Appendix Fig.~\ref{fig:agentic_examples}.

From Fig.~\ref{fig:agentic_analysis}d, $85\%$ of per-page points lie above the diagonal of the RAW- vs. Agent$+$geo- graph, which means that $85\%$ of pages achieve improvements. We use mean-F1@IoU0.5 for this experiment because the target output is one clean box per text block region without a direct confidence score, and AP can decrease even when the final decomposition is more useful for downstream document processing.

\subsection{Entity-Level Semantic Tagging}
\label{sec:tagging_experiments}

After the document regions are localized, entity-level tagging asks the model to decide what semantic role the localized region plays in the document. This isolates semantic interpretation from geometric decomposition. We evaluate this setting with a 12-class region taxonomy derived from the annotation scheme in Sec.~\ref{sec:data_tagging}, comparing three zero-shot open VLM families (Qwen2.5-VL, InterVL3, Gemma4-12B) against a fine-tuned LocateAnything-3B tagger.

\paragraph{Evaluation Metrics} The evaluation is controlled around the same localized region. Each open VLM sees the full page with a red-box marker, pixel coordinates, and the candidate 12-class label list, while LocateAnything uses its native \texttt{<ref>$\cdot$</ref><box>$\cdot$</box>} input format. We report two region presentations: a \emph{single-image} view, where the model sees the marked full page, and a \emph{dual-image} view, where the marked page is paired with a crop around the target region. All rows in Table~\ref{tab:reason_locany_tag} are class accuracy scored on the same class-balanced validation set (30 regions per class, $n{=}360$, so micro and macro accuracy coincide).

\subsubsection{Analysis}

From the results in Table.~\ref{tab:reason_locany_tag}, we can see that VLMs overuse top classes while almost never recovering the rare classes in the long tail of the label distribution. Fine-tuning is helpful in recovering the rare classes in the long-tail distribution, as we could see from LocateAnything's performance on Background Image, Color Block, and Dialog Box. 

The single-image and dual-image views expose a second dimension of difficulty. Cropping is consistently helpful for zero-shot models, but the gain is modest, which suggests that the primary bottleneck is not simply seeing the region at higher resolution. For the fine-tuned tagger, cropping helps label decisions based on local visual form, such as Color Block, Table, Dialog Box, and Line / Divider, but it hurts label accuracy for page-level contexts, especially Photograph, Text Block, and Icon. This is probably because the functionality of regions in a document may depend on the region's own semantics, as well as on the role of the whole document. A single crop cannot preserve page-level signal.

\label{sec:reason_locany_tag_fix}

The long-tail distribution of classes continues to influence tagging accuracy. Under-exposed classes such as Photograph, Table, and Dialog Box remain below where their visual separability suggests they should be. We conducted experiments with resampling that oversamples rare classes by $\min(\sqrt{N_{\max}/N_c},\,8)$, giving the tail more training exposure. Results confirm the frequency diagnosis by lifting Photograph ($30\rightarrow57$), Table ($63\rightarrow81$), Chart ($74\rightarrow88$), and Dialog Box ($56\rightarrow69$), but it sacrifices some top-class accuracies at the same time (Appendix Table.~\ref{tab:reason_locany_tag_fix}).

We also run an auxiliary VLM-tagger diagnostic on the original 30-class taxonomy to separate representation errors from decoding errors. A Qwen3.5-9B-VL tagger trained on a 15-label set achieves about $60$--$61\%$ accuracy, but its greedy decoder never emits the rare \texttt{Visual Motif/Pattern} label. Rather than allowing unconstrained generation, we then compute the teacher-forced likelihood for each candidate label and select the highest-scoring label as the closed-set prediction. The teacher-forced way means that, when scoring a candidate label, the evaluator supplies that label’s preceding tokens to the model rather than letting the model generate its own preceding tokens. When candidate labels are scored using such teacher-forced likelihood rather than free generation, length-normalizing the score recovers Visual Motif recall from $0$ to $23\%$ with no retraining, and a mild logit adjustment reaches $43\%$ at the cost of top-class accuracy (Appendix Table~\ref{tab:reason_vlm_closedset}). 

\begin{table}[!h]
\centering
\scriptsize
\setlength{\tabcolsep}{4pt}
\caption{Region content-type tagging on the 12-class taxonomy: zero-shot open VLMs vs.\ the fine-tuned LocateAnything-3B tagger.}
\label{tab:reason_locany_tag}
\begin{tabular*}{\linewidth}{@{\extracolsep{\fill}}l cc cc cc cc@{}}
\toprule
 & \multicolumn{6}{c}{Zero-shot} & \multicolumn{2}{c}{LocateAny.\ (FT)} \\
\cmidrule(lr){2-7}\cmidrule(lr){8-9}
 & \multicolumn{2}{c}{Qwen2.5-VL} & \multicolumn{2}{c}{InternVL3} & \multicolumn{2}{c}{Gemma4-12B} & & \\
\cmidrule(lr){2-3}\cmidrule(lr){4-5}\cmidrule(lr){6-7}\cmidrule(lr){8-9}
Class & Single & Dual & Single & Dual & Single & Dual & Single & Dual \\
\midrule
Text Block / Content         & 73.3 & 73.3 & 56.7 & 66.7 & 50.0 & 46.7 & 86.7 & 53.3 \\
Photograph                   & 40.0 & 20.0 & 56.7 & 60.0 & 76.7 & 83.3 & 86.7 & 40.0 \\
Line / Divider               & 16.7 & 16.7 & 13.3 & 40.0 & 46.7 & 60.0 & 76.7 & 90.0 \\
Icon                         & 30.0 & 46.7 & 33.3 & 43.3 & 33.3 & 36.7 & 83.3 & 63.3 \\
Brand Logo                   & 76.7 & 70.0 & 63.3 & 66.7 & 80.0 & 90.0 & 46.7 & 40.0 \\
Table                        & 70.0 & 86.7 & 80.0 & 83.3 & 76.7 & 83.3 & 46.7 & 66.7 \\
Chart / Graph                & 53.3 & 56.7 & 56.7 & 60.0 & 63.3 & 66.7 & 66.7 & 70.0 \\
Color Block                  & 10.0 & 10.0 & 10.0 & 10.0 & 16.7 & 20.0 & 43.3 & 86.7 \\
Dialog Box                   & 36.7 & 30.0 & 13.3 & 10.0 & 26.7 & 36.7 & 40.0 & 60.0 \\
Illustration / Artwork       & 13.3 & 20.0 & 46.7 & 30.0 & 50.0 & 56.7 & 50.0 & 63.3 \\
Decorative / Pattern Graphic &  6.7 &  6.7 &  0.0 &  3.3 &  3.3 &  3.3 & 13.3 & 26.7 \\
Background Image             &  3.3 &  0.0 &  3.3 &  0.0 &  3.3 &  0.0 & 30.0 & 43.3 \\
\midrule
Avg & 35.8 & 36.4 & 36.1 & 39.4 & 43.9 & 48.6 & 55.8 & 58.6 \\
\bottomrule
\end{tabular*}
\end{table}

\subsection{Anchor Thinking}
\label{sec:reason_corpus}

\newcommand{\code}[1]{\texttt{#1}}

To construct \docount, a benchmark that evaluates models' counting ability in a dense setting, we build a pipeline using the \adopds dataset that contains 12 class labels. The data preparation pipeline starts with selecting target labels, simplifying human-labeled polygons, and converting the target regions into both a grounded Anchor-CoT generation and a counting benchmark.

\subsubsection{Construction of Evaluation Dataset}

\noindent \textbf{Target Vocabulary } To more directly assess MLLMs’ dense counting capability, we construct the evaluation set using 4 target labels from the 12-class taxonomy. We select this subset for its especially clear and visually separable class definitions, thereby reducing potential ambiguity in category interpretation during evaluation. \ref{twvp_def_block} attaches the complete class definitions of the following 4 target labels. 

\begin{itemize}[leftmargin=*, itemsep=2pt]
  \item \texttt{Brand Logo}: visual identity mark used to identify a brand, organization, product, service, certification, standard, award, or official program. 
  \item \texttt{Photograph}: realistic or naturalistic image content depicting real-world people, objects, places, products, scenes, textures, and related content. It may appear as a foreground or background image. 
  \item \texttt{Table}: grid-like information organized into rows, columns, or
  cells for comparison or lookup.
  \item \texttt{Chart / Graph}: self-contained quantitative or categorical
  data visualizations, representing data values using visual encodings such as position, length, area, color scale, angle, bars, lines, points, slices, heatmap cells, contours, map regions, or network nodes and edges.
\end{itemize}

\noindent \textbf{Polygon Simplification } Human-annotated object masks may contain duplicate polygon vertices or redundantly close vertices. To construct a clean polygon mask input for \twvp{} generation, the polygon mask is simplified by deleting consecutive duplicates and collinear vertices, removing the ``least important'' vertex via the Visvalingam-Whyatt algorithm, and selecting the candidate with the fewest vertices that still significantly overlaps the original polygons. If no candidate passes, the cleaned original is kept. This helped us reduce around 23.49\% percent of total vertices in polygon masks. 

\begin{table}[!t]
\centering
\scriptsize
\setlength{\tabcolsep}{3pt}
\caption{Reasoning-on exact-count accuracy on the \docount benchmark (400+ clean-label
documents), with per-class accuracies. Total accuracy is shown in the
rightmost column. Underline marks the second-best result in each column.}
\label{tab:overall}
\begin{tabular*}{\textwidth}{@{\extracolsep{\fill}}lccccc}
\toprule
Model & Brand Logo & Photograph & Table &
Chart / Graph & Acc. \\
\midrule
\multicolumn{6}{l}{\quad \textit{Open-source models}} \\
Kimi-K2.5 (1T) & 63.80 & 87.78 & 58.67 & 66.67 & 72.85 \\
Qwen3.5 (397B-A17B) & 54.60 & \underline{87.22} & \underline{65.33} & 75.00 & \underline{70.81} \\
Qwen3.6 (35B-A3B) & 55.83 & 85.00 & 56.00 & 75.00 & 68.78 \\
GLM-4.6V (106B) & 50.92 & 84.44 & 46.67 & 62.50 & 64.48 \\
Gemma-4 (31B) & 55.21 & 72.78 & 41.33 & \underline{70.83} & 60.86 \\
MiMo-VL (7B) & 46.63 & 75.56 & 48.00 & 58.33 & 59.28 \\
InternVL3.5 (30B-A3B) & 41.10 & 61.11 & 30.67 & 50.00 & 47.96 \\
Cosmos3-Super (64B) & 28.22 & 52.22 & 16.00 & 58.33 & 37.56 \\
\midrule
\multicolumn{6}{l}{\quad \textit{Proprietary models}} \\
Claude-Sonnet-4.5 & \underline{56.44} & 81.67 & 54.67 & 54.17 & 66.29 \\
GPT-5.2 & 46.63 & 82.78 & 66.67 & 75.00 & 66.29 \\
Gemini-2.5-Pro & 50.92 & 80.56 & 42.67 & 66.67 & 62.44 \\
\bottomrule
\end{tabular*}
\end{table}
\noindent \textbf{Strong-VLM Polygon-Label Verification } After polygon simplification, we input the polygon masks for the top-1 class (by count) from each image into a VLM verifier (GPT-5.5) for class label verification. To build a dense counting dataset, only documents in which the top-1 class contains more than 3 objects are retained.
The verifier prompt contains the original image, an overlay image with indexed polygons, and a prompt asking whether each visible polygon belongs to the expected class. The verifier vocabulary includes the four target classes, plus a \texttt{Other/None of the Above} tag. The verifier only checks whether selected polygons are mislabeled, given the class definition. It does not check whether the entire document contains additional unmarked polygons of the same class. Missing instances are handled by the following manual review process.

\noindent \textbf{Manual Missing-Label Review } Document candidates that pass the verifier's check are manually reviewed for missing
top-1-class instances and documents with missing labels are excluded from the evaluation dataset. During the same pass, the human reviewer also sanity-checks the
visible labeled polygons: if a marked polygon clearly does not belong to the
target class, the image is excluded together with missing-label cases. This
pipeline finally produces a conservative but reliable evaluation set for \docount.

\noindent \textbf{\twvp{} Input Construction } The 400+ documents are packaged for \twvp{} generation. Only the verified top-1 class per document is used. For each sample, the CoT generator (GPT 5.5) receives the image, the target class definition, and a list of simplified target polygons as
placeholder assignments:$\code{[[1]]} = [[x,y],\ldots],\, \code{[[2]]} = [[x,y],\ldots],\, \ldots, \,$
the CoT generator is prompted to write natural first-person reasoning with what target label is required to be counted in the document, what each polygon instance is about, and what the final count is to the question. This design preserves the original separation between geometry and language. The CoT generator controls the prose, but the count, instance input order, and polygon
coordinates remain deterministic functions of the verified clean labels. Examples are presented in \ref{sec:reason_qualitative}.
\vspace{-0.5em}
\subsubsection{Evaluation}
The 400+ documents now serve as \docount, a benchmark dataset for dense counting evaluation of current state-of-the-art VLMs. For evaluation experiments in this paper, target polygon masks, polygon overlays, and target object lists are not included in the evaluation prompt. Each model receives only the original image and a question asking how many regions of the target class appear on the document page. The default prompt also includes the target class definition and counting instructions (see \ref{twvp_def_block}) and asks the model to return an integer count inside an answer tag.

The benchmark uses ten different question phrasings to avoid giving every
sample the identical wording. The official metric is exact integer accuracy
over all samples. 

\noindent \textbf{Overall Model Ranking } Table~\ref{tab:overall} reports the updated model counting results.
Kimi-K2.5 is the best performer, followed by Qwen3.6-35B-A3B. At the same time, the task is not saturated: the best model still misses around 25\% of the samples.

\noindent \textbf{Analysis } The counting evaluation shows that current state-of-the-art VLMs still struggle with exact visual enumeration in document pages. The best condition reaches 72.85\%
accuracy, but the task remains far from solved (selected cases in \ref{sec:reason_failure_cases}). The class analysis suggests two kinds of
failure. \docount evaluates VLM's capability from both
perception- and policy-angle: VLMs must interpret definitions consistently
aligned with the annotation policy and understand the whole images in order to give out the final correct answer.

\vspace{-0.5emn}
\section{Conclusion}
\vspace{-0.5em}
This paper documented the \adopds pipeline.
First, we described the \textbf{data enrichment} protocol that adds human-cleaned captions and closed-vocabulary tags on top of \adopdf's inherited entity polygons and OCR text blocks (Sec.~\ref{sec:data_collection}).
Second, we built a model-guided 80k/20k/20k split whose validation and test sets are representative, mutually matched, and moderately challenging, selected from multiple candidates by an ensemble model-challenge calibration (Appendix~\ref{sec:data_split}).
Third, our \textbf{document decomposition and localization} study showed that YOLOv12, RF-DETR, SAM3/SAM3.1, and the VLM-based LocateAnything grounder all face a substantial domain gap on \adopds, while an agentic grouping refinement over predicted text boxes can recover part of the over-fragmentation error without additional training (Sec.~\ref{sec:decomposition_experiments}).
Fourth, our \textbf{entity-level tagging} study showed that zero-shot VLMs struggle with document-specific region semantics, while fine-tuning a LocateAnything-style tagger recovers part of the class label in the long tail distribution (Sec.~\ref{sec:tagging_experiments}). 
Finally, our \textbf{\docount benchmarking} also reveals a current challenge for state-of-the-art VLMs in performing the counting task with definition-following, semantic understanding, and dense grounding on documents (Sec.~\ref{sec:reason_corpus}).

Together, \adopds opens multiple directions for future work: improving small-entity detection under dense layouts, bridging the domain gap for CJK documents, scaling the grounded-narration corpus to the full reasoning-question taxonomy, and training MLLMs that leverage grounded visual anchors to reason about real-world document pages.

\subsection*{Acknowledgement}
This work was supported by Adobe Research Gift Funding.

\FloatBarrier
\clearpage
\bibliographystyle{colm2026_conference}
\bibliography{adopd2026}

@techreport{lu2026think,
  title={Thinking with Visual Primitives},
  author={Lu, Ruijie and Ma, Yiyang and Chen, Xiaokang and Luo, Lingxiao and Wu, Zhiyu and Pan, Zizheng and Liu, Xingchao and Lin, Yutong and Li, Hao and Liu, Wen and Hao, Zhewen and Gao, Xi and Nie, Shaoheng and Wei, Yixuan and Xie, Zhenda and Chen, Ting and Zeng, Gang},
  institution={DeepSeek-AI},
  type={Technical report},
  year={2026}
}

@book{chaudhuri2007digital,
  title={Digital Document Processing: Major Directions and Recent Advances},
  editor={Chaudhuri, Bidyut B.},
  year={2007},
  publisher={Springer London},
  doi={10.1007/978-1-84628-726-8}
}

@inproceedings{zhong2019publaynet,
  title={{PubLayNet}: Largest Dataset Ever for Document Layout Analysis},
  author={Zhong, Xu and Tang, Jianbin and Yepes, Antonio Jimeno},
  booktitle={International Conference on Document Analysis and Recognition (ICDAR)},
  pages={1015--1022},
  year={2019},
  doi={10.1109/ICDAR.2019.00166}
}

@inproceedings{Mathur_2023_AAAI,
  author={Mathur, Puneet and Jain, Rajiv and Gu, Jiuxiang and Dernoncourt, Franck and Manocha, Dinesh and Morariu, Vlad I.},
  title={DocEdit: Language-Guided Document Editing},
  booktitle={AAAI},
  pages={1914--1922},
  year={2023},
  doi={10.1609/aaai.v37i2.25282}
}

@inproceedings{mathew2021docvqa,
  title={{DocVQA}: A Dataset for {VQA} on Document Images},
  author={Mathew, Minesh and Karatzas, Dimosthenis and Jawahar, C. V.},
  booktitle={IEEE/CVF Winter Conference on Applications of Computer Vision (WACV)},
  pages={2200--2209},
  year={2021}
}

@InProceedings{Cheng_2023_CVPR,
    author    = {Cheng, Hiuyi and Zhang, Peirong and Wu, Sihang and Zhang, Jiaxin and Zhu, Qiyuan and Xie, Zecheng and Li, Jing and Ding, Kai and Jin, Lianwen},
    title     = {M6Doc: A Large-Scale Multi-Format, Multi-Type, Multi-Layout, Multi-Language, Multi-Annotation Category Dataset for Modern Document Layout Analysis},
    booktitle = {CVPR},
  year = {2023}
}

@inproceedings{li2006docbank,
  title={{DocBank}: A Benchmark Dataset for Document Layout Analysis},
  author={Li, Minghao and Xu, Yiheng and Cui, Lei and Huang, Shaohan and Wei, Furu and Li, Zhoujun and Zhou, Ming},
  booktitle={Proceedings of the 28th International Conference on Computational Linguistics (COLING)},
  pages={949--960},
  year={2020},
  doi={10.18653/v1/2020.coling-main.82}
}

@inproceedings{pfitzmann2022doclaynet,
  title={{DocLayNet}: A Large Human-Annotated Dataset for Document-Layout Segmentation},
  author={Pfitzmann, Birgit and Auer, Christoph and Dolfi, Michele and Nassar, Ahmed S. and Staar, Peter W. J.},
  booktitle={SIGKDD},
  pages={3743--3751},
  year={2022},
  doi={10.1145/3534678.3539043}
}

@inproceedings{kim2021donut,
  title={{OCR}-Free Document Understanding Transformer},
  author={Kim, Geewook and Hong, Teakgyu and Yim, Moonbin and Nam, JeongYeon and Park, Jinyoung and Yim, Jinyeong and Hwang, Wonseok and Yun, Sangdoo and Han, Dongyoon and Park, Seunghyun},
  booktitle={European Conference on Computer Vision (ECCV)},
  pages={498--517},
  year={2022},
  doi={10.1007/978-3-031-19815-1_29}
}

@inproceedings{smith2007overview,
  title={An Overview of the Tesseract {OCR} Engine},
  author={Smith, Ray},
  booktitle={International Conference on Document Analysis and Recognition (ICDAR)},
  pages={629--633},
  year={2007},
  doi={10.1109/ICDAR.2007.4376991}
}

@inproceedings{selfdoc2021,
  title={SelfDoc: Self-Supervised Document Representation Learning},
  author={Li, Peizhao and Gu, Jiuxiang and Kuen, Jason and Morariu, Vlad I. and Zhao, Handong and Jain, Rajiv and Manjunatha, Varun and Liu, Hongfu},
  booktitle={IEEE/CVF Conference on Computer Vision and Pattern Recognition (CVPR)},
  pages={5652--5660},
  year={2021}
}

@inproceedings{huang2022layoutlmv3,
  title={LayoutLMv3: Pre-training for Document AI with Unified Text and Image Masking},
  author={Huang, Yupan and Lv, Tengchao and Cui, Lei and Lu, Yutong and Wei, Furu},
  booktitle={ACMMM},
  year={2022}
}

@inproceedings{carion2020end,
  title={End-to-End Object Detection with Transformers},
  author={Carion, Nicolas and Massa, Francisco and Synnaeve, Gabriel and Usunier, Nicolas and Kirillov, Alexander and Zagoruyko, Sergey},
  booktitle={European Conference on Computer Vision (ECCV)},
  pages={213--229},
  year={2020},
  publisher={Springer}
}

@article{DBLP:journals/ijcv/EveringhamGWWZ10,
  author    = {Mark Everingham and
               Luc Van Gool and
               Christopher K. I. Williams and
               John M. Winn and
               Andrew Zisserman},
  title     = {The Pascal Visual Object Classes {(VOC)} Challenge},
  journal   = {International Journal of Computer Vision},
  volume    = {88},
  number    = {2},
  pages     = {303--338},
  year      = {2010}
}

@inproceedings{DBLP:conf/eccv/LinMBHPRDZ14,
  author    = {Tsung{-}Yi Lin and
               Michael Maire and
               Serge J. Belongie and
               James Hays and
               Pietro Perona and
               Deva Ramanan and
               Piotr Doll{\'{a}}r and
               C. Lawrence Zitnick},
  editor    = {David J. Fleet and
               Tom{\'{a}}s Pajdla and
               Bernt Schiele and
               Tinne Tuytelaars},
  title     = {Microsoft {COCO:} Common Objects in Context},
  booktitle = {Computer Vision - {ECCV} 2014 - 13th European Conference, Zurich,
               Switzerland, September 6-12, 2014, Proceedings, Part {V}},
  series    = {Lecture Notes in Computer Science},
  volume    = {8693},
  pages     = {740--755},
  publisher = {Springer},
  year      = {2014},
  url       = {https://doi.org/10.1007/978-3-319-10602-1\_48},
  doi       = {10.1007/978-3-319-10602-1\_48},
  bibsource = {dblp computer science bibliography, https://dblp.org}
}

@inproceedings{mondal2020iiit,
  title={IIIT-AR-13K: a new dataset for graphical object detection in documents},
  author={Mondal, Ajoy and Lipps, Peter and Jawahar, CV},
  booktitle={DAS},
  year={2020}
}

@inproceedings{gu2021unidoc,
  title={{UniDoc}: Unified Pretraining Framework for Document Understanding},
  author={Gu, Jiuxiang and Kuen, Jason and Morariu, Vlad I. and Zhao, Handong and Jain, Rajiv and Barmpalios, Nikolaos and Nenkova, Ani and Sun, Tong},
  booktitle={NeurIPS},
  pages={39--50},
  year={2021}
}

@article{ouwayed2012general,
  title={A general approach for multi-oriented text line extraction of handwritten documents},
  author={Ouwayed, Nazih and Bela{\"\i}d, Abdel},
  journal={IJDAR},
  year={2012}
}

@inproceedings{tang2023unifying,
  title={Unifying Vision, Text, and Layout for Universal Document Processing},
  author={Tang, Zineng and Yang, Ziyi and Wang, Guoxin and Fang, Yuwei and Liu, Yang and Zhu, Chenguang and Zeng, Michael and Zhang, Cha and Bansal, Mohit},
  booktitle={IEEE/CVF Conference on Computer Vision and Pattern Recognition (CVPR)},
  pages={19254--19264},
  year={2023}
}

@inproceedings{lee2019page,
  title={Page Segmentation Using a Convolutional Neural Network with Trainable Co-occurrence Features},
  author={Lee, Joonho and Hayashi, Hideaki and Ohyama, Wataru and Uchida, Seiichi},
  booktitle={International Conference on Document Analysis and Recognition (ICDAR)},
  pages={1023--1028},
  year={2019},
  doi={10.1109/ICDAR.2019.00167}
}

@inproceedings{kirillov2023segany,
  title={Segment Anything},
  author={Kirillov, Alexander and Mintun, Eric and Ravi, Nikhila and Mao, Hanzi and Rolland, Chloe and Gustafson, Laura and Xiao, Tete and Whitehead, Spencer and Berg, Alexander C. and Lo, Wan-Yen and Doll{\'a}r, Piotr and Girshick, Ross},
  booktitle={IEEE/CVF International Conference on Computer Vision (ICCV)},
  pages={4015--4026},
  year={2023}
}

@inproceedings{adopd2024,
  title={{AD}o{PD}: A Large-Scale Document Page Decomposition Dataset},
  author={Gu, Jiuxiang and Shi, Xiangxi and Kuen, Jason and Qi, Lu and Zhang, Ruiyi and Liu, Anqi and Nenkova, Ani and Sun, Tong},
  booktitle={International Conference on Learning Representations (ICLR)},
  year={2024},
  url={https://openreview.net/forum?id=x1ptaXpOYa}
}

@article{ge2021yolox,
  title={{YOLOX}: Exceeding {YOLO} Series in 2021},
  author={Ge, Zheng and Liu, Songtao and Wang, Feng and Li, Zeming and Sun, Jian},
  journal={arXiv preprint arXiv:2107.08430},
  year={2021}
}

@article{tian2025yolov12,
  title={{YOLOv12}: Attention-Centric Real-Time Object Detectors},
  author={Tian, Yunjie and Ye, Qixiang and Doermann, David},
  journal={arXiv preprint arXiv:2502.12524},
  year={2025}
}

@article{ravi2024sam2,
  title={{SAM} 2: Segment Anything in Images and Videos},
  author={Ravi, Nikhila and Gabeur, Valentin and Hu, Yuan-Ting and Hu, Ronghang and Ryali, Chaitanya and Ma, Tengyu and Khedr, Haitham and R{\"a}dle, Roman and Rolland, Chloe and Gustafson, Laura and Mintun, Eric and Pan, Junting and Alwala, Kalyan Vasudev and Carion, Nicolas and Wu, Chao-Yuan and Girshick, Ross and Doll{\'a}r, Piotr and Feichtenhofer, Christoph},
  journal={arXiv preprint arXiv:2408.00714},
  year={2024}
}

@article{carion2025sam3,
  title={{SAM} 3: Segment Anything with Concepts},
  author={Carion, Nicolas and Gustafson, Laura and Hu, Yuan-Ting and Debnath, Shoubhik and Hu, Ronghang and Suris, Didac and Ryali, Chaitanya and Alwala, Kalyan Vasudev and Khedr, Haitham and Huang, Andrew and Lei, Jie and Ma, Tengyu and Guo, Baishan and Kalla, Arpit and Marks, Markus and Greer, Joseph and Wang, Meng and Sun, Peize and R{\"a}dle, Roman and Afouras, Triantafyllos and Mavroudi, Effrosyni and Xu, Katherine and Wu, Tsung-Han and Zhou, Yu and Momeni, Liliane and Hazra, Rishi and Ding, Shuangrui and Vaze, Sagar and Porcher, Francois and Li, Feng and Li, Siyuan and Kamath, Aishwarya and Cheng, Ho Kei and Doll{\'a}r, Piotr and Ravi, Nikhila and Saenko, Kate and Zhang, Pengchuan and Feichtenhofer, Christoph},
  journal={arXiv preprint arXiv:2511.16719},
  year={2025}
}

@inproceedings{rfdetr2024,
  title={{RF-DETR}: Neural Architecture Search for Real-Time Detection Transformers},
  author={Robinson, Isaac and Robicheaux, Peter and Popov, Matvei and Ramanan, Deva and Peri, Neehar},
  booktitle={International Conference on Learning Representations (ICLR)},
  year={2026},
  url={https://openreview.net/forum?id=qHm5GePxTh}
}

@article{wang2026locateanything,
  title={LocateAnything: Fast and High-Quality Vision-Language Grounding with Parallel Box Decoding},
  author={Wang, Shihao and Liu, Shilong and Kuang, Yuanguo and Wei, Xinyu and Liu, Yangzhou and Li, Zhiqi and Man, Yunze and Chen, Guo and Tao, Andrew and Liu, Guilin and Kautz, Jan and Zhang, Lei and Yu, Zhiding},
  journal={arXiv preprint arXiv:2605.27365},
  year={2026}
}

@inproceedings{groundingdino2024,
  title={Grounding {DINO}: Marrying {DINO} with Grounded Pre-Training for Open-Set Object Detection},
  author={Liu, Shilong and Zeng, Zhaoyang and Ren, Tianhe and Li, Feng and Zhang, Hao and Yang, Jie and Jiang, Qing and Li, Chunyuan and Yang, Jianwei and Su, Hang and Zhu, Jun and Zhang, Lei},
  booktitle={ECCV},
  year={2024}
}

@inproceedings{pix2seq2022,
  title={Pix2Seq: A Language Modeling Framework for Object Detection},
  author={Chen, Ting and Saxena, Saurabh and Li, Lala and Fleet, David J. and Hinton, Geoffrey},
  booktitle={International Conference on Learning Representations (ICLR)},
  year={2022},
  url={https://openreview.net/forum?id=e42KbIw6Wb}
}

@inproceedings{florence22024,
  title={Florence-2: Advancing a Unified Representation for a Variety of Vision Tasks},
  author={Xiao, Bin and Wu, Haiping and Xu, Weijian and Dai, Xiyang and Hu, Houdong and Lu, Yumao and Zeng, Michael and Liu, Ce and Yuan, Lu},
  booktitle={CVPR},
  year={2024}
}

@article{kosmos22023,
  title={Kosmos-2: Grounding Multimodal Large Language Models to the World},
  author={Peng, Zhiliang and Wang, Wenhui and Dong, Li and Hao, Yaru and Huang, Shaohan and Ma, Shuming and Wei, Furu},
  journal={arXiv preprint arXiv:2306.14824},
  year={2023}
}

@article{shikra2023,
  title={Shikra: Unleashing Multimodal {LLM}'s Referential Dialogue Magic},
  author={Chen, Keqin and Zhang, Zhao and Zeng, Weili and Zhang, Richong and Zhu, Feng and Zhao, Rui},
  journal={arXiv preprint arXiv:2306.15195},
  year={2023}
}

@inproceedings{ferret2024,
  title={Ferret: Refer and Ground Anything Anywhere at Any Granularity},
  author={You, Haoxuan and Zhang, Haotian and Gan, Zhe and Du, Xianzhi and Zhang, Bowen and Wang, Zirui and Cao, Liangliang and Chang, Shih-Fu and Yang, Yinfei},
  booktitle={ICLR},
  year={2024}
}

@inproceedings{unifiedio22024,
  title={Unified-{IO} 2: Scaling Autoregressive Multimodal Models with Vision, Language, Audio, and Action},
  author={Lu, Jiasen and Clark, Christopher and Lee, Sangho and Zhang, Zichen and Khosla, Savya and Marten, Ryan and Hoiem, Derek and Kembhavi, Aniruddha},
  booktitle={CVPR},
  year={2024}
}

@inproceedings{mtp2024,
  title={Better \& Faster Large Language Models via Multi-token Prediction},
  author={Gloeckle, Fabian and Youbi Idrissi, Badr and Rozi{\`e}re, Baptiste and Lopez-Paz, David and Synnaeve, Gabriel},
  booktitle={Proceedings of the 41st International Conference on Machine Learning},
  series={Proceedings of Machine Learning Research},
  volume={235},
  pages={15706--15734},
  publisher={PMLR},
  year={2024},
  url={https://proceedings.mlr.press/v235/gloeckle24a.html}
}

@article{medusa2024,
  title={Medusa: Simple {LLM} Inference Acceleration Framework with Multiple Decoding Heads},
  author={Cai, Tianle and Li, Yuhong and Geng, Zhengyang and Peng, Hongwu and Lee, Jason D. and Chen, Deming and Dao, Tri},
  journal={arXiv preprint arXiv:2401.10774},
  year={2024}
}

@inproceedings{maskgit2022,
  title={Mask{GIT}: Masked Generative Image Transformer},
  author={Chang, Huiwen and Zhang, Han and Jiang, Lu and Liu, Ce and Freeman, William T.},
  booktitle={CVPR},
  year={2022}
}

@article{llada2025,
  title={Large Language Diffusion Models},
  author={Nie, Shen and Zhu, Fengqi and You, Zebin and Zhang, Xiaolu and Ou, Jingyang and Hu, Jun and Zhou, Jun and Lin, Yankai and Wen, Ji-Rong and Li, Chongxuan},
  journal={arXiv preprint arXiv:2502.09992},
  year={2025}
}

@article{dream2025,
  title={Dream 7B: Diffusion Large Language Models},
  author={Ye, Jiacheng and Xie, Zhihui and Zheng, Lin and Gao, Jiahui and Wu, Zirui and Jiang, Xin and Li, Zhenguo and Kong, Lingpeng},
  journal={arXiv preprint arXiv:2508.15487},
  year={2025}
}

@inproceedings{llava2023,
  title={Visual Instruction Tuning},
  author={Liu, Haotian and Li, Chunyuan and Wu, Qingyang and Lee, Yong Jae},
  booktitle={NeurIPS},
  year={2023}
}

@article{qwenvl2023,
  title={{Qwen-VL}: A Versatile Vision-Language Model for Understanding, Localization, Text Reading, and Beyond},
  author={Bai, Jinze and Bai, Shuai and Yang, Shusheng and Wang, Shijie and Tan, Sinan and Wang, Peng and Lin, Junyang and Zhou, Chang and Zhou, Jingren},
  journal={arXiv preprint arXiv:2308.12966},
  year={2023}
}

@article{multimodalcot2023,
  title={Multimodal Chain-of-Thought Reasoning in Language Models},
  author={Zhang, Zhuosheng and Zhang, Aston and Li, Mu and Zhao, Hai and Karypis, George and Smola, Alex},
  journal={arXiv preprint arXiv:2302.00923},
  year={2023}
}

@inproceedings{visprog2023,
  title={Visual Programming: Compositional Visual Reasoning Without Training},
  author={Gupta, Tanmay and Kembhavi, Aniruddha},
  booktitle={CVPR},
  year={2023}
}

@inproceedings{vipergpt2023,
  title={{ViperGPT}: Visual Inference via Python Execution for Reasoning},
  author={Sur{\'i}s, D{\'i}dac and Menon, Sachit and Vondrick, Carl},
  booktitle={ICCV},
  year={2023}
}

@article{mmreact2023,
  title={{MM-REACT}: Prompting ChatGPT for Multimodal Reasoning and Action},
  author={Yang, Zhengyuan and Li, Linjie and Wang, Jianfeng and Lin, Kevin and Azarnasab, Ehsan and Ahmed, Faisal and Liu, Zicheng and Liu, Ce and Zeng, Michael and Wang, Lijuan},
  journal={arXiv preprint arXiv:2303.11381},
  year={2023}
}

@article{setofmark2023,
  title={Set-of-Mark Prompting Unleashes Extraordinary Visual Grounding in {GPT-4V}},
  author={Yang, Jianwei and Zhang, Hao and Li, Feng and Zou, Xueyan and Li, Chunyuan and Gao, Jianfeng},
  journal={arXiv preprint arXiv:2310.11441},
  year={2023}
}

@article{arriola2025bd3lm,
  title={Block Diffusion: Interpolating Between Autoregressive and Diffusion Language Models},
  author={Arriola, Marianne and Gokaslan, Aaron and Chiu, Justin T. and Yang, Zhihan and Qi, Zhixuan and Han, Jiaqi and Sahoo, Subham Sekhar and Kuleshov, Volodymyr},
  journal={arXiv preprint arXiv:2503.09573},
  year={2025}
}

@article{cheng2025sdar,
  title={{SDAR}: A Synergistic Diffusion-AutoRegression Paradigm for Scalable Sequence Generation},
  author={Cheng, Shuang and Bian, Yihan and Liu, Dawei and Zhang, Linfeng and Yao, Qian and Tian, Zhongbo and Wang, Wenhai and Guo, Qipeng and Chen, Kai and Qi, Biqing and Zhou, Bowen},
  journal={arXiv preprint arXiv:2510.06303},
  year={2025}
}

@article{wu2025fastdllmv2,
  title={{Fast-dLLM v2}: Efficient Block-Diffusion {LLM}},
  author={Wu, Chengyue and Zhang, Hao and Xue, Shuchen and Diao, Shizhe and Fu, Yonggan and Liu, Zhijian and Molchanov, Pavlo and Luo, Ping and Han, Song and Xie, Enze},
  journal={arXiv preprint arXiv:2509.26328},
  year={2025}
}

@article{zhu2026flare,
  title={{FLARE}: Diffusion for Hybrid Language Model},
  author={Zhu, Yuchen and Shi, Jing and Ge, Chongjian and Tan, Hao and Xu, Yiran and Zhu, Wanrong and Kuen, Jason and Goswami, Koustava and Jain, Rajiv and Chen, Yongxin and Tao, Molei and Gu, Jiuxiang},
  journal={arXiv preprint arXiv:2606.01774},
  year={2026}
}

@inproceedings{tjong2003conll,
  title={Introduction to the {CoNLL}-2003 Shared Task: Language-Independent Named Entity Recognition},
  author={Tjong Kim Sang, Erik F. and De Meulder, Fien},
  booktitle={Proceedings of the Seventh Conference on Natural Language Learning at {HLT-NAACL}},
  pages={142--147},
  year={2003}
}

@article{deng2026holocount,
  title={HoloCount: A Holistic Visual Counting Benchmark for {MLLMs}},
  author={Deng, Jinhong and Qiao, Limeng and Wan, Guanglu},
  journal={arXiv preprint arXiv:2607.06420},
  year={2026},
  url={https://arxiv.org/abs/2607.06420}
}

@INPROCEEDINGS{shanghaitech,
  author={Zhang, Yingying and Zhou, Desen and Chen, Siqin and Gao, Shenghua and Ma, Yi},
  booktitle={2016 IEEE Conference on Computer Vision and Pattern Recognition (CVPR)}, 
  title={Single-Image Crowd Counting via Multi-Column Convolutional Neural Network}, 
  year={2016},
  volume={},
  number={},
  pages={589-597},
  doi={10.1109/CVPR.2016.70}}

@article{nwpucrowd,
  title={NWPU-crowd: A large-scale benchmark for crowd counting and localization},
  author={Wang, Qi and Gao, Junyu and Lin, Wei and Li, Xuelong},
  journal={IEEE transactions on pattern analysis and machine intelligence},
  volume={43},
  number={6},
  pages={2141--2149},
  year={2020},
  publisher={IEEE}
}

@article{jhucrowd,
  title={Jhu-crowd++: Large-scale crowd counting dataset and a benchmark method},
  author={Sindagi, Vishwanath A and Yasarla, Rajeev and Patel, Vishal M},
  journal={IEEE transactions on pattern analysis and machine intelligence},
  volume={44},
  number={5},
  pages={2594--2609},
  year={2020},
  publisher={IEEE}
}

@inproceedings{carpk,
  title={Drone-based object counting by spatially regularized regional proposal network},
  author={Hsieh, Meng-Ru and Lin, Yen-Liang and Hsu, Winston H},
  booktitle={2017 IEEE International Conference on Computer Vision (ICCV)},
  pages={4165--4173},
  year={2017},
  organization={IEEE}
}

@inproceedings{ucfqnrf,
  title={Composition loss for counting, density map estimation and localization in dense crowds},
  author={Idrees, Haroon and Tayyab, Muhmmad and Athrey, Kishan and Zhang, Dong and Al-Maadeed, Somaya and Rajpoot, Nasir and Shah, Mubarak},
  booktitle={European Conference on Computer Vision},
  pages={544--559},
  year={2018},
  organization={Springer}
}

@InProceedings{fsc147,
    author    = {Ranjan, Viresh and Sharma, Udbhav and Nguyen, Thu and Hoai, Minh},
    title     = {Learning To Count Everything},
    booktitle = {Proceedings of the IEEE/CVF Conference on Computer Vision and Pattern Recognition (CVPR)},
    month     = {June},
    year      = {2021},
    pages     = {3394-3403}
}

@inproceedings{countbench,
  title={Teaching clip to count to ten},
  author={Paiss, Roni and Ephrat, Ariel and Tov, Omer and Zada, Shiran and Mosseri, Inbar and Irani, Michal and Dekel, Tali},
  booktitle={2023 IEEE/CVF International Conference on Computer Vision (ICCV)},
  pages={3147--3157},
  year={2023},
  organization={IEEE}
}

@inproceedings{pixmocount,
  title={Molmo and pixmo: Open weights and open data for state-of-the-art vision-language models},
  author={Deitke, Matt and Clark, Christopher and Lee, Sangho and Tripathi, Rohun and Yang, Yue and Park, Jae Sung and Salehi, Mohammadreza and Muennighoff, Niklas and Lo, Kyle and Soldaini, Luca and others},
  booktitle={2025 IEEE/CVF Conference on Computer Vision and Pattern Recognition (CVPR)},
  pages={91--104},
  year={2025},
  organization={IEEE}
}

@article{countqa,
  title={CountQA: How well do MLLMs count in the wild?},
  author={Tamarapalli, Jayant Sravan and Grover, Rynaa and Pande, Nilay and Yerramilli, Sahiti},
  journal={arXiv preprint arXiv:2508.06585},
  year={2025}
}

@article{holocount,
  title={HoloCount: A Holistic Visual Counting Benchmark for MLLMs},
  author={Deng, Jinhong and Qiao, Limeng and Wan, Guanglu},
  journal={arXiv preprint arXiv:2607.06420},
  year={2026}
}

@inproceedings{publaynet,
  author    = {Zhong, Xu and Tang, Jianbin and Yepes, Antonio Jimeno},
  title     = {{PubLayNet}: Largest dataset ever for document layout analysis},
  booktitle = {2019 International Conference on Document Analysis and Recognition ({ICDAR})},
  year      = {2019},
  pages     = {1015--1022},
  publisher = {IEEE},
  doi       = {10.1109/ICDAR.2019.00166},
  url       = {https://doi.org/10.1109/ICDAR.2019.00166}
}

@inproceedings{docbank,
  author    = {Li, Minghao and Xu, Yiheng and Cui, Lei and Huang, Shaohan and Wei, Furu and Li, Zhoujun and Zhou, Ming},
  title     = {{DocBank}: A Benchmark Dataset for Document Layout Analysis},
  booktitle = {Proceedings of the 28th International Conference on Computational Linguistics},
  year      = {2020},
  month     = dec,
  pages     = {949--960},
  address   = {Barcelona, Spain (Online)},
  publisher = {International Committee on Computational Linguistics},
  doi       = {10.18653/v1/2020.coling-main.82},
  url       = {https://aclanthology.org/2020.coling-main.82/}
}

@inproceedings{iiitar13k,
  author    = {Mondal, Ajoy and Lipps, Peter and Jawahar, C. V.},
  title     = {{IIIT-AR-13K}: A New Dataset for Graphical Object Detection in Documents},
  booktitle = {Document Analysis Systems},
  series    = {Lecture Notes in Computer Science},
  volume    = {12116},
  year      = {2020},
  pages     = {216--230},
  address   = {Cham},
  publisher = {Springer International Publishing},
  doi       = {10.1007/978-3-030-57058-3_16},
  url       = {https://doi.org/10.1007/978-3-030-57058-3_16}
}

@inproceedings{doclaynet,
  author    = {Pfitzmann, Birgit and Auer, Christoph and Dolfi, Michele and Nassar, Ahmed S. and Staar, Peter W. J.},
  title     = {{DocLayNet}: A Large Human-Annotated Dataset for Document-Layout Segmentation},
  booktitle = {Proceedings of the 28th ACM SIGKDD Conference on Knowledge Discovery and Data Mining},
  year      = {2022},
  pages     = {3743--3751},
  address   = {New York, NY, USA},
  publisher = {Association for Computing Machinery},
  doi       = {10.1145/3534678.3539043},
  url       = {https://doi.org/10.1145/3534678.3539043}
}

@inproceedings{m6doc,
  author    = {Cheng, Hiuyi and Zhang, Peirong and Wu, Sihang and Zhang, Jiaxin and Zhu, Qiyuan and Xie, Zecheng and Li, Jing and Ding, Kai and Jin, Lianwen},
  title     = {{$M^{6}$Doc}: A Large-Scale Multi-Format, Multi-Type, Multi-Layout, Multi-Language, Multi-Annotation Category Dataset for Modern Document Layout Analysis},
  booktitle = {Proceedings of the IEEE/CVF Conference on Computer Vision and Pattern Recognition ({CVPR})},
  year      = {2023},
  month     = jun,
  pages     = {15138--15147},
  url       = {https://openaccess.thecvf.com/content/CVPR2023/html/Cheng_M6Doc_A_Large-Scale_Multi-Format_Multi-Type_Multi-Layout_Multi-Language_Multi-Annotation_Category_Dataset_CVPR_2023_paper.html}
}

@article{docgenome,
  author  = {Xia, Renqiu and Mao, Song and Yan, Xiangchao and Zhou, Hongbin and Zhang, Bo and Peng, Haoyang and Pi, Jiahao and Fu, Daocheng and Wu, Wenjie and Ye, Hancheng and Feng, Shiyang and Wang, Bin and Xu, Chao and He, Conghui and Cai, Pinlong and Dou, Min and Shi, Botian and Zhou, Sheng and Wang, Yongwei and Wang, Bin and Yan, Junchi and Wu, Fei and Qiao, Yu},
  title   = {{DocGenome}: An Open Large-scale Scientific Document Benchmark for Training and Testing Multi-modal Large Language Models},
  journal = {arXiv preprint arXiv:2406.11633},
  year    = {2024},
  doi     = {10.48550/arXiv.2406.11633},
  url     = {https://arxiv.org/abs/2406.11633}
}

@article{paldb,
  author  = {Pe{\~n}a, Alejandro and Morales, Aythami and Fierrez, Julian and Ortega-Garcia, Javier and Puente, I{\~n}igo and Cordova, Jorge and Cordova, Gonzalo},
  title   = {Continuous document layout analysis: Human-in-the-loop {AI}-based data curation, database, and evaluation in the domain of public affairs},
  journal = {Information Fusion},
  volume  = {108},
  pages   = {102398},
  year    = {2024},
  doi     = {10.1016/j.inffus.2024.102398},
  url     = {https://doi.org/10.1016/j.inffus.2024.102398}
}

@article{diachronicdocument,
  author  = {Cl{\'e}rice, Thibault and Jan{\`e}s, Juliette and Scheithauer, Hugo and B{\'e}ni{\`e}re, Sarah and Cafiero, Florian and Romary, Laurent and Gabay, Simon and Sagot, Beno{\^i}t},
  title   = {Diachronic Document Dataset for Semantic Layout Analysis},
  journal = {arXiv preprint arXiv:2411.10068},
  year    = {2024},
  doi     = {10.48550/arXiv.2411.10068},
  url     = {https://arxiv.org/abs/2411.10068}
}

@article{doclayoutyolo,
  author  = {Zhao, Zhiyuan and Kang, Hengrui and Wang, Bin and He, Conghui},
  title   = {{DocLayout-YOLO}: Enhancing Document Layout Analysis through Diverse Synthetic Data and Global-to-Local Adaptive Perception},
  journal = {arXiv preprint arXiv:2410.12628},
  year    = {2024},
  doi     = {10.48550/arXiv.2410.12628},
  url     = {https://arxiv.org/abs/2410.12628}
}

@inproceedings{graphdoc,
  author    = {Chen, Yufan and Liu, Ruiping and Zheng, Junwei and Wen, Di and Peng, Kunyu and Zhang, Jiaming and Stiefelhagen, Rainer},
  title     = {Graph-based Document Structure Analysis},
  booktitle = {The Thirteenth International Conference on Learning Representations ({ICLR})},
  year      = {2025},
  url       = {https://proceedings.iclr.cc/paper_files/paper/2025/hash/cf3d7d8e79703fe947deffb587a83639-Abstract-Conference.html}
}

@inproceedings{indicdlp,
  author    = {Nath, Oikantik and Kukkala, Sahithi and Khapra, Mitesh and Sarvadevabhatla, Ravi Kiran},
  title     = {{IndicDLP}: A Foundational Dataset for Multi-lingual and Multi-domain Document Layout Parsing},
  booktitle = {Document Analysis and Recognition -- {ICDAR} 2025},
  series    = {Lecture Notes in Computer Science},
  volume    = {16023},
  year      = {2026},
  pages     = {23--39},
  address   = {Cham},
  publisher = {Springer Nature Switzerland},
  doi       = {10.1007/978-3-032-04614-7_2},
  url       = {https://doi.org/10.1007/978-3-032-04614-7_2},
  isbn      = {978-3-032-04614-7}
}

@article{monkeyocr,
  author  = {Li, Zhang and Liu, Yuliang and Liu, Qiang and Ma, Zhiyin and Zhang, Ziyang and Zhang, Shuo and Guo, Zidun and Zhang, Jiarui and Wang, Xinyu and Bai, Xiang},
  title   = {{MonkeyOCR}: Document Parsing with a Structure-Recognition-Relation Triplet Paradigm},
  journal = {arXiv preprint arXiv:2506.05218},
  year    = {2025},
  doi     = {10.48550/arXiv.2506.05218},
  url     = {https://arxiv.org/abs/2506.05218}
}

@inproceedings{scan,
  author    = {Ueda, Nobuhiro and Dong, Yuyang and Boros, Kriszti{\'a}n and Ito, Daiki and Sera, Takuya and Oyamada, Masafumi},
  title     = {{SCAN}: Semantic Document Layout Analysis for Textual and Visual Retrieval-Augmented Generation},
  booktitle = {Findings of the Association for Computational Linguistics: {EACL} 2026},
  year      = {2026},
  month     = mar,
  pages     = {1618--1637},
  address   = {Rabat, Morocco},
  publisher = {Association for Computational Linguistics},
  doi       = {10.18653/v1/2026.findings-eacl.82},
  url       = {https://aclanthology.org/2026.findings-eacl.82/}
}

@article{docatlas,
  author  = {Heakl, Ahmed and Mohamed, Youssef and Sohail, Abdullah and Elbadry, Rania and Nassar, Ahmed and Staar, Peter W. J. and Khan, Fahad Shahbaz and Razzak, Imran and Khan, Salman},
  title   = {{DocAtlas}: Multilingual Document Understanding Across 80+ Languages},
  journal = {arXiv preprint arXiv:2605.12623},
  year    = {2026},
  doi     = {10.48550/arXiv.2605.12623},
  url     = {https://arxiv.org/abs/2605.12623}
}

@article{team2026kwai,
  title={Kwai Keye-VL-2.0 Technical Report},
  author={Team, Kwai Keye and Wen, Bin and Liu, Changyi and Song, Chengru and Rao, Chongling and Zhang, Guowang and Li, Han and Fan, Haonan and Ju, Hengrui and Chen, Jiankang and others},
  journal={arXiv preprint arXiv:2606.10651},
  year={2026}
}

\FloatBarrier
\clearpage
\appendix
\section{\textbf{Appendix}}
\subsection{Model-Guided Data Split Construction}
\label{sec:data_split}

\noindent\textbf{Motivation.}
The released \adopds corpus contains approximately 120k densely annotated
samples.  A random split is undesirable because the dataset is intentionally
long-tailed: taxonomy labels, languages, layout styles, text density, bounding
box density, and segmentation-mask complexity are all correlated.  We therefore
construct an 80k/20k/20k train/validation/test split with two goals.  First, the
validation and test sets should be representative of the full corpus and
closely matched to each other.  Second, they should contain sufficiently
challenging samples for modern detection and segmentation systems, while
avoiding an adversarially hard subset on which models would receive little
learning signal.

\noindent\textbf{Taxonomy-balanced sampling.}
We start from the sample-level metadata and human annotations.  The fine-grained
taxonomy strings are normalized and semantically merged into 100 coarse
taxonomy groups.  All validation and test sampling is then performed around
these 100 groups rather than around raw folder names or annotation counts alone.
Within each taxonomy group, we further stratify by language, box/mask
complexity, and document profile.  The primary balancing fields are:
(i) taxonomy cluster, (ii) language group, (iii) segmentation/bounding-box
complexity tier, (iv) text-rich versus visual-rich profile, (v) dominant mask
label family, and (vi) layout aspect-ratio profile.  This procedure ensures
that the validation and test splits cover the same semantic regions of the
corpus while still preserving rare document styles.

\definecolor{splitTrain}{HTML}{1596CD}
\definecolor{splitVal}{HTML}{F32763}
\definecolor{splitTest}{HTML}{8EDDED}
\definecolor{splitCore}{HTML}{D8EEF4}
\definecolor{langEn}{HTML}{0C6FAE}
\definecolor{langZh}{HTML}{23A8D6}
\definecolor{langJa}{HTML}{7BD5E8}
\definecolor{langKo}{HTML}{064E7A}
\definecolor{langOther}{HTML}{F32763}

\newcommand{\splitpie}{%
\begin{tikzpicture}[font=\sffamily,scale=1.0]

  \def\splitseg##1##2##3{%
    \fill[##3,draw=white,line width=1.6pt]
      (##1:1.55) arc[start angle=##1,end angle=##2,radius=1.55]
      -- (##2:0.78) arc[start angle=##2,end angle=##1,radius=0.78]
      -- cycle;
  }

  \begin{scope}[shift={(0,0)}]
    \splitseg{90}{-150}{splitTrain}   %
    \splitseg{-150}{-210}{splitVal}   %
    \splitseg{-210}{-270}{splitTest}  %

    \node[align=center] at (0,0)
      {{\large\bfseries 120k}\\[-1.5pt]{\scriptsize documents}};
    \node[align=center,font=\small\bfseries] at (0,2.20) {Corpus split};

    \node[white,font=\scriptsize\bfseries,align=center] at (-30:1.16) {Train\\[-2pt]80k};
    \node[white,font=\tiny\bfseries,align=center] at (-180:1.16) {Val\\[-2pt]20k};
    \node[font=\tiny\bfseries,align=center,text=black!70] at (-240:1.16) {Test\\[-2pt]20k};
  \end{scope}

  \def\barL{5.30}       %
  \def\barR{12.95}      %
  \def\barH{0.50}       %

  \begin{scope}
    \pgfmathsetmacro\barW{\barR-\barL}   %
    \pgfmathsetmacro\pscale{\barW/100.0} %

    \node[anchor=west,font=\small\bfseries] at (\barL-0.05,2.55)
      {Language distribution matched across splits};
    \node[anchor=west,font=\scriptsize,text=gray!70] at (\barL-0.05,2.18)
      {aligned segment boundaries confirm \emph{val}\,/\,\emph{test} mirror the \emph{corpus} \& \emph{train}};

    \begin{scope}[shift={(\barL,1.55)}]
      \def\leg##1##2##3{%
        \fill[##2] (##1,0) rectangle ++(0.26,0.26);
        \node[anchor=west,font=\scriptsize] at (##1+0.34,0.13) {##3};
      }
      \leg{0.00}{langEn}{English}
      \leg{1.70}{langZh}{Chinese}
      \leg{3.45}{langJa}{Japanese}
      \leg{5.35}{langKo}{Korean}
      \leg{6.95}{langOther}{Other}
    \end{scope}

    \def\rowCorpus{1.00}
    \def\rowTrain{0.35}
    \def\rowVal{-0.30}
    \def\rowTest{-0.95}
    \pgfmathsetmacro\guideTop{\rowCorpus+\barH/2+0.10}
    \pgfmathsetmacro\guideBot{\rowTest-\barH/2}

    \def\stackrow##1##2##3##4##5##6##7##8{%
      \node[anchor=east,font=\scriptsize\bfseries,text=##3] at (\barL-0.18,##1) {##2};
      \pgfmathsetmacro\eb{##4}
      \pgfmathsetmacro\ec{##4+##5}
      \pgfmathsetmacro\ed{##4+##5+##6}
      \pgfmathsetmacro\ee{##4+##5+##6+##7}
      \pgfmathsetmacro\xEnB{\barL+\eb*\pscale}
      \pgfmathsetmacro\xZhB{\barL+\ec*\pscale}
      \pgfmathsetmacro\xJaB{\barL+\ed*\pscale}
      \pgfmathsetmacro\xKoB{\barL+\ee*\pscale}
      \pgfmathsetmacro\yA{##1-\barH/2}
      \pgfmathsetmacro\yB{##1+\barH/2}
      \fill[langEn]    (\barL,\yA) rectangle (\xEnB,\yB);
      \fill[langZh]    (\xEnB,\yA) rectangle (\xZhB,\yB);
      \fill[langJa]    (\xZhB,\yA) rectangle (\xJaB,\yB);
      \fill[langKo]    (\xJaB,\yA) rectangle (\xKoB,\yB);
      \fill[langOther] (\xKoB,\yA) rectangle (\barR,\yB);
      \draw[white,line width=0.7pt] (\xEnB,\yA)--(\xEnB,\yB);
      \draw[white,line width=0.7pt] (\xZhB,\yA)--(\xZhB,\yB);
      \draw[white,line width=0.7pt] (\xJaB,\yA)--(\xJaB,\yB);
      \draw[white,line width=0.7pt] (\xKoB,\yA)--(\xKoB,\yB);
    }
    \stackrow{\rowCorpus}{Corpus}{black!75}{50.00}{16.67}{16.67}{7.04}{9.63}
    \stackrow{\rowTrain}{Train}{splitTrain!75!black}{49.71}{16.57}{16.63}{7.08}{10.00}
    \stackrow{\rowVal}{Val}{splitVal!85!black}{50.59}{16.84}{16.73}{6.94}{8.89}
    \stackrow{\rowTest}{Test}{splitTest!55!black}{50.57}{16.86}{16.73}{6.96}{8.88}

    \def\pctrow##1##2##3##4{%
      \pgfmathsetmacro\cEn{##2/2}
      \pgfmathsetmacro\cZh{##2+##3/2}
      \pgfmathsetmacro\cJa{##2+##3+##4/2}
      \node[white,font=\tiny\bfseries] at (\barL+\cEn*\pscale,##1) {##2};
      \node[white,font=\tiny\bfseries] at (\barL+\cZh*\pscale,##1) {##3};
      \node[font=\tiny\bfseries,text=black!55] at (\barL+\cJa*\pscale,##1) {##4};
    }
    \pctrow{\rowCorpus}{50.00}{16.67}{16.67}
    \pctrow{\rowTrain}{49.71}{16.57}{16.63}
    \pctrow{\rowVal}{50.59}{16.84}{16.73}
    \pctrow{\rowTest}{50.57}{16.86}{16.73}

    \foreach \cum in {50.00,66.67,83.34,90.38}{
      \pgfmathsetmacro\gx{\barL+\cum*\pscale}
      \draw[white,opacity=0.55,dashed,dash pattern=on 1.6pt off 1.6pt,line width=0.5pt]
        (\gx,\guideTop) -- (\gx,\guideBot);
    }
  \end{scope}

  \begin{scope}[shift={(0,-1.62)}]
    \fill[splitCore!55] (-1.85,-1.02) rectangle (12.95,0.0);
    \draw[gray!25,line width=0.5pt] (-1.85,-1.02) rectangle (12.95,0.0);

    \def\colA{-1.55}  %
    \def\colB{3.30}   %
    \def\colC{6.90}   %
    \def\colD{10.40}  %

    \node[anchor=west,font=\scriptsize\bfseries,text=gray!75] at (\colA,-0.24)
      {Held-out complexity};
    \node[anchor=west,font=\scriptsize\bfseries,text=gray!75] at (\colB,-0.24) {Valid images};
    \node[anchor=west,font=\scriptsize\bfseries,text=gray!75] at (\colC,-0.24) {Inst./image};
    \node[anchor=west,font=\scriptsize\bfseries,text=gray!75] at (\colD,-0.24) {Mask vertices};
    \draw[gray!30,line width=0.5pt] (-1.78,-0.38) -- (12.88,-0.38);

    \def\statline##1##2##3##4##5##6{%
      \fill[##2] (\colA-0.20,##1-0.075) rectangle ++(0.10,0.22);
      \node[anchor=west,font=\scriptsize\bfseries,text=##2] at (\colA,##1+0.035) {##3};
      \node[anchor=west,font=\scriptsize] at (\colB,##1+0.035) {##4};
      \node[anchor=west,font=\scriptsize] at (\colC,##1+0.035) {##5};
      \node[anchor=west,font=\scriptsize] at (\colD,##1+0.035) {##6};
    }
    \statline{-0.61}{splitVal}{Validation}{19,048}{12.79}{224.09}
    \statline{-0.86}{splitTest}{Test}{19,111}{12.77}{225.97}
  \end{scope}

\end{tikzpicture}%
}

\begin{figure*}[t]
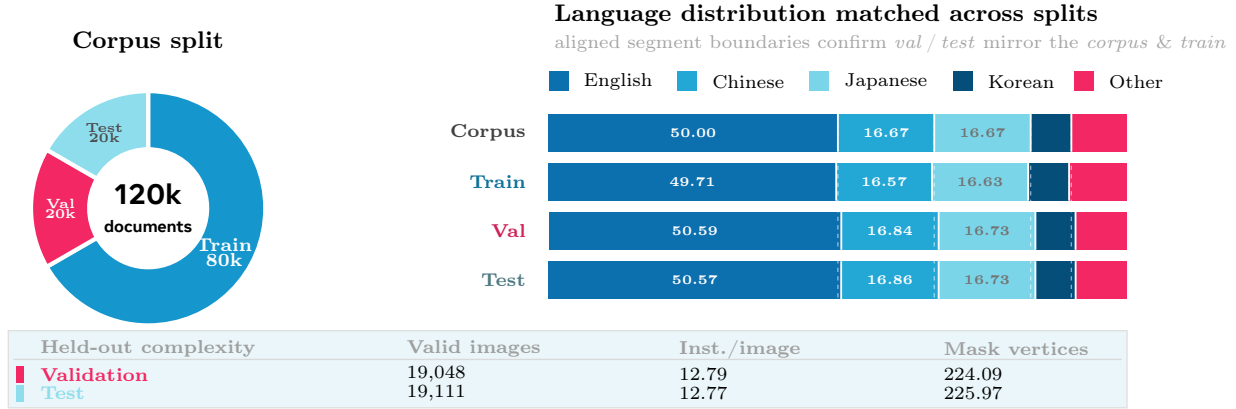

\centering
\splitpie
\caption{The \adopds corpus is partitioned into an 80k/20k/20k
train/validation/test split. The left donut shows split size; the stacked bars
give the language composition (English/Chinese/Japanese/Korean/other) of the
full corpus and of each split, where the aligned segment boundaries show that
the validation and test language distributions mirror the corpus and train; and
the bottom strip reports the matched annotation complexity (valid images,
instances per image, mask vertices) of the held-out sets.}
\label{fig:data_split_partition}
\end{figure*}

\noindent\textbf{Annotation-complexity features.}
For each sample, we compute annotation statistics from the human entity masks
and text boxes.  These include the number of valid mask instances, the number of
polygons, polygon vertices, mask-area coverage, box counts, box-area coverage,
and the balance between text-heavy and visual-heavy entities.  We group samples
into five segmentation/box complexity tiers.  Validation and test are drawn so
that every taxonomy cluster contains a mixture of low-, medium-, and
high-complexity examples.  Thus, a category is not represented only by simple
pages; for example, a financial-document group contains both sparse price cards
and dense table-like layouts.

\noindent\textbf{Candidate split generation.}
After constructing the initial balanced split, we generate five candidate
validation/test partitions from the same 40k held-out pool.  The training set
remains fixed at 80k samples.  The candidates are:
\textbf{(1)} the current taxonomy/language/complexity-balanced split,
\textbf{(2)} a taxonomy-language-tier alternating split that pairs hard and
easy samples within each bucket, \textbf{(3)} a taxonomy-language-tier split
matched by model-challenge deciles, \textbf{(4)} a complexity-profile-matched
split focused on text/visual and mask-family balance, and \textbf{(5)} a greedy
minimax split that jointly minimizes taxonomy, language, complexity, and
model-challenge gaps.  We explicitly exclude an extreme-hard stress-test split
from the final candidates because our goal is to evaluate progress on learnable
document decomposition rather than to create an adversarial benchmark.

\noindent\textbf{Model-challenge calibration.}
We evaluate the held-out 40k pool with multiple third-party codebases and
multiple inference settings.  For each model $m$ and image $i$, we compute
box recall and mask recall at IoU 0.5 over ground-truth instances and define
\begin{equation}
    c_i^m = 1 - \frac{1}{2}
    \left(R^{m,\mathrm{box}}_{i,0.5} + R^{m,\mathrm{mask}}_{i,0.5}\right).
    \label{eq:model_challenge_score}
\end{equation}
The ensemble challenge score is the average of $c_i^m$ over all completed
model/parameter runs.  We use this score only as a calibration signal: a good
validation/test split should be moderately challenging, representative, and
well matched across val and test.  It should not simply maximize failure rate.

\begin{table}[t]
\centering
\caption{Planned entity-level model suite for model-challenge calibration.
All configurations use a single class, \textit{entity}; taxonomy and language
labels are used only for split balancing and subgroup analysis.}
\label{tab:data_split_model_calibration}
\scriptsize
\setlength{\tabcolsep}{3pt}
\begin{threeparttable}
\begin{tabular*}{\textwidth}{@{\extracolsep{\fill}}llllll}
\toprule
Task &
Codebase &
Version / Variant &
Size(s) &
Checkpoint / Source &
Primary Metrics \\
\midrule
\multirow{6}{*}{Detection}
& RF-DETR & Detection & Nano, 2XL & local / downloadable & AP$_\text{box}$, AP$_{50}^\text{box}$, AP$_{75}^\text{box}$ \\
& YOLOv12 & Turbo detection & M, X & downloadable & AP$_\text{box}$, AP$_{50}^\text{box}$, AP$_{75}^\text{box}$ \\
& GroundingDINO & v1.0 SwinT & T & local & AP$_\text{box}$, prompt sensitivity \\
& GroundingDINO & v1.5 / DINO-X & API / large & API checkpoint & AP$_\text{box}$, prompt sensitivity \\
& SAM3 & text-conditioned detector & default & HF checkpoint & AP$_\text{box}$, recall@IoU \\
& SAM3.1 & text-conditioned detector & default & HF checkpoint & AP$_\text{box}$, recall@IoU \\
\midrule
\multirow{6}{*}{Segmentation}
& RF-DETR & Segmentation & Nano, 2XL & local & AP$_\text{mask}$, AP$_{50}^\text{mask}$, AP$_{75}^\text{mask}$ \\
& YOLOv12 & Segmentation release & M, X & local / downloadable & AP$_\text{mask}$, AP$_\text{box}$ \\
& Grounded-SAM2 & GDINO v1.0 + SAM2.1 & Tiny, Large & local & AP$_\text{mask}$, AP$_\text{box}$ \\
& Grounded-SAM2 & DINO-X + SAM2.1 & Large & API + local SAM2.1 & AP$_\text{mask}$, prompt sensitivity \\
& SAM3 & text-conditioned segmentation & default & HF checkpoint & AP$_\text{mask}$, recall@IoU \\
& SAM3.1 & text-conditioned segmentation & default & HF checkpoint & AP$_\text{mask}$, recall@IoU \\
\bottomrule
\end{tabular*}
\vspace{1mm}
\begin{minipage}{0.98\textwidth}
\scriptsize
\emph{Notes.}
Prediction target: every valid human entity is mapped to category id 1.
Detection and segmentation are evaluated as separate tracks.  Confidence,
box-threshold, and text-threshold sweeps are tuned on validation and frozen for
test.  During this calibration phase, foundation-model rows are evaluated
zero-shot to rank candidate splits; in the final baseline study
(Sec.~\ref{sec:experiments_2026}) the trainable architectures are fine-tuned
and additionally reported zero-shot for reference.  All results are stratified
by taxonomy, language, text richness, and visual complexity.
\end{minipage}
\end{threeparttable}
\end{table}

\noindent\textbf{Selecting the final validation and test sets.}
For each candidate split $S$, we recompute COCO metrics by filtering the
full-pool predictions to that candidate's validation and test IDs.  We then
rank candidates using a composite score:
\begin{equation}
\begin{aligned}
    \mathcal{L}(S) ={}&
    6\Delta_{\mathrm{mAP}}(S)
    + 3\Delta_{\mathrm{challenge}}(S) \\
    &+ 2\Delta_{\mathrm{feature}}(S)
    + \Delta_{\mathrm{representative}}(S) \\
    &+ 2\Delta_{\mathrm{base}}(S)
    - 0.25\overline{\mathrm{AP}}(S).
\end{aligned}
\label{eq:split_selection_score}
\end{equation}
Here $\Delta_{\mathrm{mAP}}$ is the average val/test gap in box and mask AP
over all model settings, $\Delta_{\mathrm{challenge}}$ is the val/test gap in
ensemble challenge score, $\Delta_{\mathrm{feature}}$ is the mean distribution
gap over taxonomy, language, complexity, and layout fields, and
$\Delta_{\mathrm{representative}}$ measures how close val and test are to the
full held-out pool.  The small AP reward prevents selecting a pathological
ultra-hard split when two candidates have otherwise similar balance.  The
lowest-scoring candidate is used as the final 20k validation and 20k test split.

\begin{table}[t]
\centering
\caption{Final candidate ranking after recomputing COCO mAP for each
candidate validation and test set.  The mAP gap, challenge gap, feature gap,
representativeness gap, and mean AP columns are reported in percentage points;
lower score is better.}
\label{tab:data_split_candidate_selection}
\scriptsize
\setlength{\tabcolsep}{3pt}
\begin{tabular*}{\textwidth}{@{\extracolsep{\fill}}lrrrrrr}
\toprule
Candidate & Score & mAP Gap &
Chal. Gap & Feat. Gap & Rep. Gap &
Mean AP \\
\midrule
Greedy minimax all signals & 0.0115 & 0.20 & 0.05 & 0.16 & 0.08 & 2.45 \\
Taxonomy-language challenge decile & 0.0454 & 0.29 & 0.12 & 1.17 & 0.59 & 2.36 \\
Taxonomy-language hard alternating & 0.0485 & 0.24 & 0.23 & 1.18 & 0.59 & 2.37 \\
Complexity-profile matched & 0.0671 & 0.41 & 0.16 & 1.68 & 0.84 & 2.31 \\
Initial balanced split & 0.0733 & 0.33 & 0.72 & 1.01 & 0.51 & 2.44 \\
\bottomrule
\end{tabular*}
\end{table}

\noindent\textbf{Selected split statistics.}
The selected split is the greedy minimax all-signal candidate.  It contains
80,000 training samples, 20,000 validation samples, and 20,000 test samples,
with files stored as candidate~05 in the model-challenge split root.
After removing samples without valid entity masks under the COCO conversion rules, the full-pool model
calibration evaluates 38,159 images: 19,048 from validation and 19,111 from
test, with 249,113 ground-truth entity-mask instances.  Duplicate sample
leakage between train, validation, and test is zero after sample-ID and
metadata-path checks.  The selected validation/test split is also closely
matched in annotation complexity: validation and test contain 12.79 and 12.77
boxes or polygons per image on average, 224.09 and 225.97 polygon vertices,
and nearly identical text-rich and visual-rich scores.

\subsection{Training and Inference Setup for Visual Anchor Localization}
\label{training_setup_exp1}
All supervised baselines are trained on the 80k train split, and model checkpoints used for evaluation are selected based on validation performance. YOLOv12 and RF-DETR are fine-tuned for 10 epochs with an input-resolution sweep (384/640/768/1024) and task-dependent global batches (YOLOv12 96--256; RF-DETR 32--48). SAM3/SAM3.1 use the same COCO-style data pipeline and are evaluated as mask predictors. Because their presence head becomes unstable in a naive long run with focal-loss $\gamma{=}0$, we use a numerically stable loss path. SAM3/SAM3.1 compute the ($\gamma=0$) presence-head focal loss as alpha-weighted binary cross-entropy loss instead of the unstable Triton kernel in focal-loss backward computation, preventing NaN gradients without changing the objective. After this correction, training is stable for the full 10 epochs, with mask AP peaking around epoch 3 and remaining within ${\sim}$1 point through epoch 10. For LocateAnything, we convert each page into a unified grounding target: text blocks contribute boxes, and visual entities contribute paired boxes and adaptive Douglas--Peucker polygons, simplified to at most 16 vertices. We fine-tune LocateAnything end-to-end at long edge 1024 with multi-token-prediction supervision for 50k steps ($\approx$5 epochs), and report the best validation checkpoint (epoch 3) in Table~\ref{tab:adopd2026_main_results}. The zero-shot columns evaluate off-the-shelf checkpoints without \adopds fine-tuning, mapping model outputs to the single \textit{entity} class when needed.

\subsubsection{Fine-Tuning SAM3/SAM3.1}
\label{sec:sam3_schedule_appendix}

As we mentioned in the paragraph above, SAM3/SAM3.1 require a numerically stable loss path, and training converges quickly under such a loss function. The additional training epochs did not degrade the model's performance, as shown in Table~\ref{tab:sam3_schedule}, which provides subset AP measured on a fixed 1{,}240-image validation subset. The full-split values use the same 20k validation split reported in the main results.

\begin{table}[!htbp]
\centering
\scriptsize
\setlength{\tabcolsep}{5pt}
\caption{SAM3/SAM3.1 fine-tuning schedule on \adopds (validation mask AP). After the $\gamma{=}0$ focal-loss fix, AP peaks at epoch~3 and stays on a narrow plateau through epoch~10. The full 20k-split AP is measured at the epoch-3 peak and epoch-10 end (and reported in Table~\ref{tab:adopd2026_main_results}). The per-epoch trajectory ep1$\to$ep10 is traced on a fixed 1{,}240-image subset, where the same peak-then-plateau shape is evident.}
\label{tab:sam3_schedule}
\begin{tabular*}{\linewidth}{@{\extracolsep{\fill}}l c c c c c@{}}
\toprule
& \multicolumn{4}{c}{Subset trajectory (1.24k imgs)} & Full 20k \\
\cmidrule(lr){2-5}\cmidrule(lr){6-6}
Model & ep1 & ep3 & ep5 & ep10 & ep3 / ep10 \\
\midrule
SAM3   & 45.1 & 48.1 & 47.7 & 47.6 & 57.2 / 56.3 \\
SAM3.1 & 43.8 & 49.3 & 48.9 & 49.2 & 59.4 / 58.5 \\
\bottomrule
\end{tabular*}
\\[2pt]
{\scriptsize We use a small, fixed evaluation subset solely to illustrate the trajectory's shape. Subset AP is systematically lower than full-split AP. Only the \emph{shape} (which peaks at ep3 and plateaus thereafter) is the claim. On the full split, epoch~10 remains within $\sim$1 AP of the epoch-3 peak for both models.}
\end{table}
\FloatBarrier

\subsubsection{LocateAnything Unified-Grounding Details}
\label{sec:locany_grounding_appendix}

Tables~\ref{tab:reason_locany_train}--\ref{tab:reason_locany_speed} provide the training, evaluation, and decoding details. The unified grounding target is learnable for LocateAnything, while its lower detector-style AP mainly reflects confidence ranking, tokenized geometry, polygon-mask limitations, and decoding brittleness rather than failed document-domain adaptation.

\begin{table}[!htbp]
\centering
\scriptsize
\setlength{\tabcolsep}{6pt}
\caption{LocateAnything unified-grounding training summary. Loss is cross-entropy; the selected checkpoint is step 30k (epoch 3).}
\label{tab:reason_locany_train}
\begin{tabular*}{\linewidth}{@{\extracolsep{\fill}}l c c c@{}}
\toprule
Model & Entity targets & First loss & Loss @\,30k \\
\midrule
Unified (box\,+\,polygon) & \texttt{<box>}\,+\,\texttt{<quad>} & 4.18 & 0.33 \\
\bottomrule
\end{tabular*}
\end{table}

\begin{table}[!htbp]
\centering
\scriptsize
\setlength{\tabcolsep}{5pt}
\caption{LocateAnything-3B-DP unified-grounding evaluation on the \adopds validation split (\%).}
\label{tab:reason_locany_results}
\begin{tabular*}{\linewidth}{@{\extracolsep{\fill}}l r r r r r r r@{}}
\toprule
Output & AP & AP$_{50}$ & AP$_{75}$ & AP$_S$ & AP$_M$ & AP$_L$ & mF1 \\
\midrule
Box (\texttt{<box>})    & 35.9 & 49.4 & 38.0 & 7.2 & 22.9 & 41.2 & 57.1 \\
Mask (\texttt{<quad>})  & 32.5 & 47.4 & 34.0 & 5.7 & 20.3 & 38.0 & -- \\
\bottomrule
\end{tabular*}
\\[2pt]
{\scriptsize AP uses pseudo-confidence by generation order; mF1 is F1@IoU Mean on boxes. The mask row is the LocateAnything-3B-DP entry reported in Table~\ref{tab:adopd2026_main_results}.}
\end{table}

\begin{table}[!htbp]
\centering
\scriptsize
\setlength{\tabcolsep}{4pt}
\caption{LocateAnything-3B-DP decoding speed/quality trade-off on \adopds. Speed is the mean over the image (median in parentheses); AP is the COCO AP on matched 1k-image validation subsets.}
\label{tab:reason_locany_speed}
\begin{tabular*}{\linewidth}{@{\extracolsep{\fill}}l r r r r r@{}}
\toprule
Mode & s/image & Speedup & AP & AP$_{50}$ & AP retain \\
\midrule
\multicolumn{6}{@{}l}{\emph{Detection} (text block, box-only; six-token box frame)} \\
slow (AR)       & 3.94 (3.25) & 1.0$\times$ & 41.1 & 57.5 & 100\% \\
hybrid (MTP+AR) & 1.81 (1.25) & 2.2$\times$ & 40.8 & 57.0 & 99\% \\
fast (MTP)      & 0.97 (0.92) & 4.1$\times$ & 32.9 & 47.0 & 80\% \\
\midrule
\multicolumn{6}{@{}l}{\emph{Segmentation} (entity, box+polygon; $13{+}$ tokens/region)} \\
slow (AR)       & 7.40 (4.00) & 1.0$\times$ & 23.1 & 36.2 & 100\% \\
hybrid (MTP+AR) & 1.85 (1.30) & 4.0$\times$ & 5.9 & 9.4 & 26\% \\
fast (MTP)      & 1.72 (1.02) & 4.3$\times$ & 3.1 & 5.6 & 13\% \\
\bottomrule
\end{tabular*}
\end{table}
\FloatBarrier

\subsection{Semantic Tagging Detailed Tables}
\label{sec:tagging_details_appendix}

Tables~\ref{tab:reason_locany_tag_fix}--\ref{tab:reason_vlm_closedset} provide the supporting values for the semantic-tagging analysis in Sec.~\ref{sec:tagging_experiments}.

\begin{table}[!htbp]
\centering
\scriptsize
\setlength{\tabcolsep}{4pt}
\caption{Class accuracy for the fine-tuned tagger on the class-balanced dual validation set.}
\label{tab:reason_locany_tag_fix}
\begin{tabular*}{\linewidth}{@{\extracolsep{\fill}}l c c @{}}
\toprule
& Baseline & Resampling \\
\midrule
Text Block / Content         & 53.8 & 60.2 \\
Photograph                   & 30.4 & 56.6 \\
Line / Divider               & 87.8 & 85.6 \\
Icon                         & 67.0 & 73.0 \\
Brand Logo                   & 44.4 & 49.6 \\
Table                        & 62.8 & 81.0 \\
Chart / Graph                & 74.0 & 87.8 \\
Color Block                  & 89.2 & 88.4 \\
Dialog Box                   & 56.0 & 68.8 \\
Background Image             & 55.4 & 55.0 \\
Illustration / Artwork       & 65.4 & 63.6 \\
Decorative / Pattern Graphic & 12.6 & 24.0 \\
\bottomrule
\end{tabular*}
\end{table}

\begin{table}[!htbp]
\centering
\scriptsize
\setlength{\tabcolsep}{6pt}
\caption{Closed-set decoding diagnostic for Qwen3.5-9B-VL tagger on 15 merged labels.}
\label{tab:reason_vlm_closedset}
\begin{tabular*}{\linewidth}{@{\extracolsep{\fill}}l c c@{}}
\toprule
Decoding policy & Visual Motif recall & Overall micro \\
\midrule
free-gen / summed log-prob        & 0/30 (0\%)   & 61\% \\
mean (length-normalized)          & 7/30 (23\%)  & 62\% \\
mean + logit-adjust $\tau{=}0.5$  & 13/30 (43\%) & 53\% \\
mean + logit-adjust $\tau{=}1.0$  & 10/30 (33\%) & 52\% \\
\bottomrule
\end{tabular*}
\end{table}

\FloatBarrier

\FloatBarrier

\subsection{\adopds Annotation Guidelines}
\label{sec:annotation_guidelines}

This section documents the annotation protocol used for the \adopds data collection round, which focuses on visually rich documents such as posters, advertisements, magazine pages, and infographics.
The primary task was to supplement existing pre-annotations by (i) correcting OCR boxes whose scope was too large, and (ii) assigning a semantic tag from the 30-class taxonomy (Table~\ref{tab:adopd_tag_taxonomy_appendix}) to every entity mask and every OCR text block.

\begin{table*}[t]
\centering
\small
\setlength{\tabcolsep}{6pt}
\renewcommand{\arraystretch}{1.08}
\caption{\adopds entity tag taxonomy. All 30 tags form a closed vocabulary; each annotated element receives exactly one tag.}
\label{tab:adopd_tag_taxonomy}
\label{tab:adopd_tag_taxonomy_appendix}
\begin{tabular*}{\textwidth}{@{\extracolsep{\fill}}>{\raggedright\arraybackslash}p{0.12\textwidth}
                >{\raggedright\arraybackslash}p{0.16\textwidth}
                >{\raggedright\arraybackslash}p{0.60\textwidth}}
\toprule
Mode & Group & Tags \\
\midrule
Mask & Visual (17) &
background image, prominent pattern, natural background, document content, table, illustration, brand logo, photograph, color block (borderless), icon, line element, chart, color block (bordered), background pattern, dialog box (bordered), decorative pattern, dialog box (borderless) \\
\midrule
OCR & Text (13) &
body text, title, list item, legend, table title, footer, note, header, emphasized text, chart title, hyperlink, citation, image caption \\
\bottomrule
\end{tabular*}
\end{table*}

\subsubsection{Annotation Output and Tag Distribution}

In total the corpus contains ${\sim}$2.5M tagged elements, split almost evenly between the two annotation modes (${\sim}$1.22M visual entity polygons and ${\sim}$1.25M text blocks).
At the region level, the resulting supervision remains strongly long-tailed: frequent anchors such as text/content blocks, photographs, lines, illustrations, and icons dominate, while backgrounds, dialog boxes, charts, tables, and decorative elements form the tail.
This long-tailed structure is important for \adopds because reasoning traces often depend on rare but semantically decisive anchors rather than only on the most frequent page regions.

\subsubsection{Layer Concept: Foreground and Background}

Every annotated element is conceptually assigned to one of two layers:

\noindent\textbf{Background.} Elements at the very bottom layer that span the full page width or height (full bleed, left--right bleed, or top--bottom bleed). Background elements support the foreground but do not directly attract attention. Typical examples: background image, background pattern, natural background.

\noindent\textbf{Foreground.} All other elements above the background. These are the primary visual or textual elements the viewer notices first: titles, body text, charts, icons, logos, color blocks, dialog boxes, \etc

\subsubsection{OCR Text Label Rules}

OCR boxes receive a tag from the \textbf{text group} (13 tags). Key disambiguation rules:

\begin{itemize}[leftmargin=*, itemsep=1pt, topsep=2pt]
\item \textbf{Title} -- Main and sub-titles; typically larger or bolder font. If a pre-annotated box groups title and sub-title together, label it \textit{title}.
\item \textbf{Body text} -- Main content paragraphs and bullet text. If a box mixes title and body text, label it \textit{body text}.
\item \textbf{List item} -- Elements in a numbered or bulleted list where items are visually cohesive. Split list items from body text/title if they appear in the same pre-annotated box. A single-item list still requires its own box with the \textit{list item} tag.
\item \textbf{Note} -- Supplementary comments or footnotes, typically at the end of an article or at the bottom of the document. In newspapers/magazines, side columns adjacent to articles are also notes.
\item \textbf{Header / Footer} -- Content at the very top or bottom of the page that is visually distinct from body text and unrelated to the main content (e.g., page numbers, chapter names, copyright lines).
\item \textbf{Legend} -- Small, inconspicuous text adjacent to a figure or chart that provides explanation.
\item \textbf{Emphasized text} -- Bold, italic, or underlined text within a paragraph. Annotate a separate box only if the segment stands alone; emphasized text embedded in a large body-text block is part of that block.
\item \textbf{Hyperlink} -- Underlined text linking to another resource. Annotate separately only if the hyperlink stands alone or has a visible prefix; hyperlinks embedded inside a large body-text block are part of body text.
\item \textbf{Table title / Chart title / Image caption} -- Titles or captions placed above or below the corresponding table, chart, or image.
\item \textbf{Brand logo (text)} -- Text portion of a logo, including purely text-based logos and the text component of text+graphic logos; select \textit{brand logo} in OCR mode.
\item \textbf{Citation} -- Text excerpted from another source, typically styled differently from body text.
\end{itemize}

\subsubsection{Mask Visual Label Rules}

Mask contours receive a tag from the \textbf{visual group} (17 tags). Key rules:

\begin{itemize}[leftmargin=*, itemsep=1pt, topsep=2pt]
\item \textbf{Photograph} -- Realistic images captured by a camera. Annotate only the overall boundary; do not annotate internal text or sub-objects. If a photograph is used purely as a background with text layered on top, label it as the appropriate background type and annotate the overlaid elements separately.
\item \textbf{Illustration} -- Drawings, paintings, or stylized/conceptual images. Cartoons that do not resemble icons are illustrations; cartoon-style symbolic images are icons.
\item \textbf{Icon} -- Symbols with specific semantic meaning (weather symbols, restroom signs, toolbar buttons, decorative stars/arrows/snowflakes, app logos, QR codes, barcodes). Icons are generally not the bottommost layer.
\item \textbf{Chart} -- Data visualizations (bar, pie, line, scatter). Annotate the entire chart boundary.
\item \textbf{Table} -- Table grid structures. Do not annotate any content inside the table.
\item \textbf{Brand logo (graphic)} -- The graphic component of a logo, including fully graphic logos and the graphic portion of text+graphic logos; select \textit{brand logo} in mask mode.
\item \textbf{Line element} -- Thin lines or curves used as dividers or decorations. A line-like shape that is visually thick should be labeled as a color block instead.
\item \textbf{Color block (bordered) / Color block (borderless)} -- Solid or semi-transparent colored areas, with or without a visible bounding edge. Each contiguous block is a separate instance.
\item \textbf{Dialog box (bordered) / Dialog box (borderless)} -- Speech-bubble shapes with or without a tail.
\item \textbf{Background image} -- A full-bleed or near-full-bleed non-photographic image at the bottom layer.
\item \textbf{Natural background} -- A photographic image that simultaneously functions as the full-bleed page background.
\item \textbf{Background pattern} -- A non-photographic decorative texture or pattern at the background layer (grids, dot matrices, shadow shapes). Shadows count as foreground background patterns.
\item \textbf{Prominent pattern} -- A single large, visually striking, semantically meaningful graphic that is clearly not an icon (e.g., a large standalone silhouette, a hero illustration).
\item \textbf{Decorative pattern} -- Minor ornamental graphics that are neither icons nor prominent patterns; typically few in number and purely decorative.
\item \textbf{Document content} -- An embedded document image within the main document. Do not annotate any internal content of the embedded document.
\end{itemize}

\subsubsection{Detailed Annotation Rules}

\begin{enumerate}[leftmargin=*, itemsep=8pt, topsep=2pt]

\item \textbf{Icon.}
Icons include: (a)~symbols with specific semantic meaning (weather, restroom, toolbar buttons); (b)~small decorative markers (snowflakes, arrows, stars); (c)~app logos; (d)~QR/barcodes.

\item \textbf{Icons and illustrations co-occurring.}
Elements inside a highlighted box may be icons or illustrations — inspect each individually.

\item \textbf{Prominent pattern vs.\ icon.}
A single large, visually striking standalone graphic = \textit{prominent pattern}. A small semantically specific symbol = \textit{icon}.

\item \textbf{Icon vs.\ color block.}
Thin arrow-shaped strokes = \textit{icon}. Visually thick bar-shaped elements = \textit{color block}.

\item \textbf{Natural background.}
If an image is simultaneously a natural photograph \emph{and} the full-bleed page background, label it \textit{natural background}.

\item \textbf{Titles in newspapers and magazines.}
Mastheads, article headlines, and section names = \textit{title}. Image labels beside photos = \textit{image caption}.

\item \textbf{Header / Footer.}
Content at the very top or bottom of the page clearly unrelated to the main content (page numbers, chapter names, copyright) = \textit{header} / \textit{footer}.

\item \textbf{Background pattern, borderless color block, hyperlink.}
Grid texture inside an element = \textit{background pattern}. Plain colored region without border = \textit{color block (borderless)}. Underlined link text = \textit{hyperlink}.

\item \textbf{Logo with text and graphic combined.}
Annotate the graphic part as \textit{brand logo} in mask mode; annotate the text part as \textit{brand logo} in OCR mode. Do not split the tag across modes.

\item \textbf{Mixed pre-annotated OCR boxes.}
Resolve conflicts: body text + title together → \textit{body text}; main + sub-title together → \textit{title}; list item + body text → split into two boxes. A single-item list must still be separated.

\item \textbf{Line element.}
A thin arrow whose arrowhead is small and inconspicuous = \textit{line element}, not icon.

\item \textbf{Background image vs.\ background pattern vs.\ prominent pattern.}
Full-bleed white base = \textit{background image}. Shadow overlay above it = \textit{background pattern} (foreground). Large graphic poster on top = \textit{prominent pattern} (foreground).

\item \textbf{Prominent pattern; newspaper notes and footer.}
A standalone eye-catching graphic = \textit{prominent pattern}. In newspaper layouts, side-column text adjacent to the main article = \textit{note}; text at the very bottom = \textit{footer}.

\item \textbf{Emphasized text and hyperlinks inside body text.}
Annotate separately if they stand alone. Hyperlinks or emphasis embedded in a large body-text block without their own pre-annotation belong to the body-text box.

\item \textbf{Note vs.\ image caption; table title.}
Contextual explanatory text at the top-left or bottom of an image block = \textit{note}. A centered label pointing directly to an image = \textit{image caption}. A heading above a table = \textit{table title}.

\item \textbf{Image caption in context; magazine header/footer vs.\ note.}
A label directly beside or below a photo = \textit{image caption}. Magazine top/bottom elements: judge by their relationship to the main article content.

\item \textbf{Background layer determination.}
Only the element(s) at the absolute bottom layer spanning the full page width or height = \textit{background}. All elements above = \textit{foreground}.

\item \textbf{Decorative pattern vs.\ prominent pattern; legend vs.\ chart title.}
Minor ornamental element with no direct meaning = \textit{decorative pattern}. Meaningful, conspicuous standalone graphic = \textit{prominent pattern}. Inconspicuous small text adjacent to a figure = \textit{legend}. Prominent label above a figure = \textit{chart title} / \textit{image caption}.

\item \textbf{Image caption vs.\ legend.}
Text directly below a photo pointing to it = \textit{image caption}. Text that serves as a note for multiple figures = \textit{legend}.

\item \textbf{List item.}
Numbered or bulleted list elements that are cohesive and clearly part of a list structure.

\item \textbf{Illustration vs.\ icon (cartoons).}
Cartoon resembling a recognizable icon-style symbol = \textit{icon}. Other cartoons = \textit{illustration}.

\item \textbf{Background image vs.\ background pattern (full-bleed non-solid).}
If the bottommost layer is a non-solid graphic spanning the full width or height, label it \textit{background image}, not \textit{background pattern}.

\item \textbf{Comprehensive annotation examples.}
Apply all rules above holistically: identify the bottommost background layer first, then label foreground elements from largest to smallest.
\end{enumerate}

\subsubsection{Label Refinement}
  The visual mask rules above describe the fine-grained annotation protocol used during dataset construction. For downstream model
  training and evaluation in Section~\ref{sec:tagging_experiments}, we use a coarser 12-class visual taxonomy. This coarsening removes distinctions that are useful for annotation guidance but difficult to learn reliably from pixels alone. In
  particular, \textit{color block (bordered)} and \textit{color block (borderless)} are merged into \textit{Color Block}, and
  \textit{dialog box (bordered)} and \textit{dialog box (borderless)} are merged into \textit{Dialog Box}. Several labels are
  renamed into canonical model-facing classes, e.g. \textit{line element} becomes \textit{Line / Divider}, \textit{chart} becomes
  \textit{Chart / Graph}, and \textit{brand logo}, \textit{photograph}, \textit{icon}, and \textit{table} are kept as direct
  counterparts. Background- and pattern-related labels are harmonized by visual role: broad page or section backdrops such as
  \textit{background image} and \textit{natural background} map to \textit{Background Image}, while ornamental or motif-like
  regions such as \textit{background pattern}, \textit{decorative pattern}, and parts of \textit{prominent pattern} are mapped to
  either \textit{Decorative / Pattern Graphic} or \textit{Illustration / Artwork} depending on whether the region functions mainly
  as decoration or as a standalone depicted graphic. Likewise, \textit{document content} regions are absorbed into the closest
  semantic class in the 12-class taxonomy, most often \textit{Text Block / Content} when the region functions as a readable
  content panel. Thus, the 17 visual labels should be understood as the fine annotation guideline, while the 12 labels form the
  normalized visual taxonomy used by the released mask-tag field and downstream tagging experiments.

\subsection{Qualitative Multi-Model Comparison}
\label{sec:appendix_multimodel_viz}
\begin{figure*}[t]\centering
  \includegraphics[width=\textwidth]{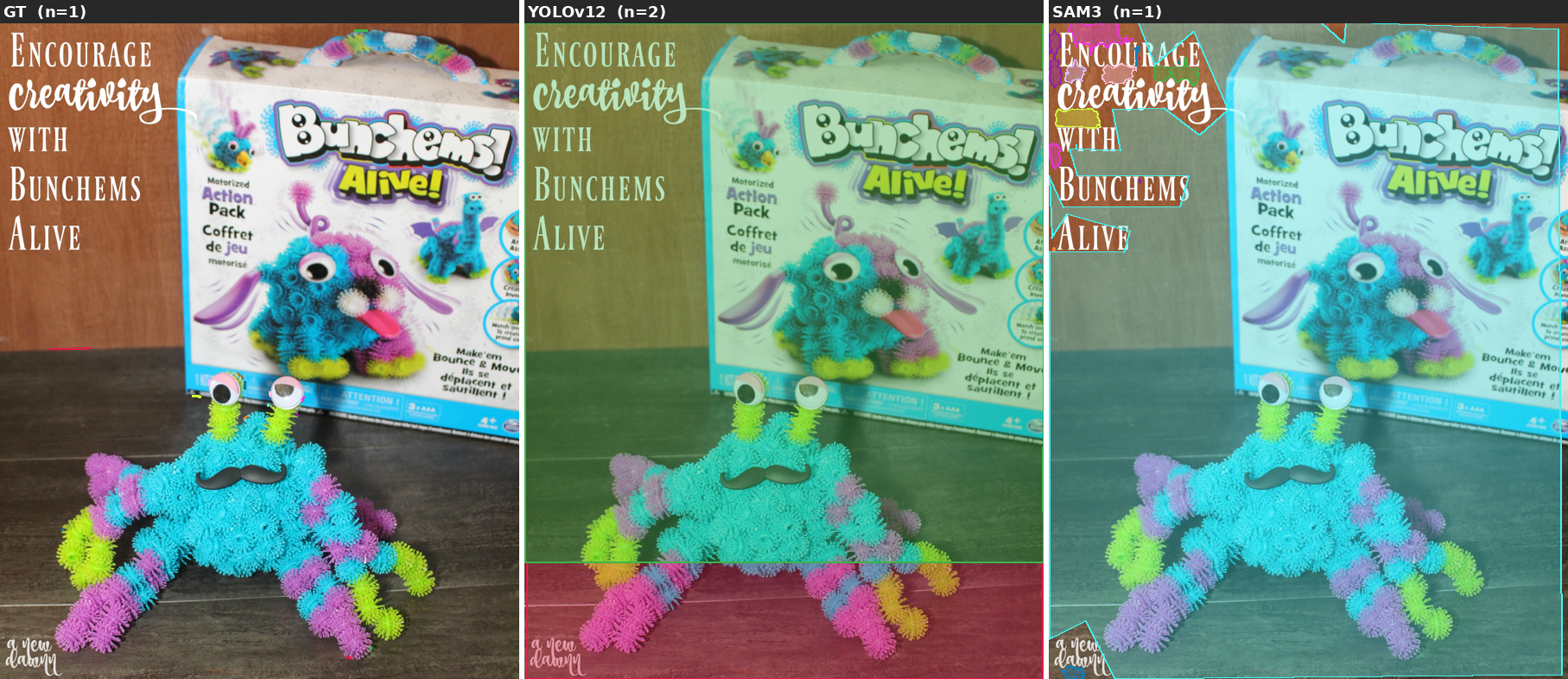}
  \caption{Qualitative comparison on a representative \emph{dense small-entity} \adopds validation page (sample \texttt{ea5c1757}; 83 OCR blocks; language \texttt{en}). Left to right: ground-truth entity masks, YOLOv12-Seg, SAM3. Per-instance polygons; panel headers give instance counts.}
  \label{fig:appx-mm-ea5c1757}
\end{figure*}

\begin{figure*}[t]\centering
  \includegraphics[width=\textwidth]{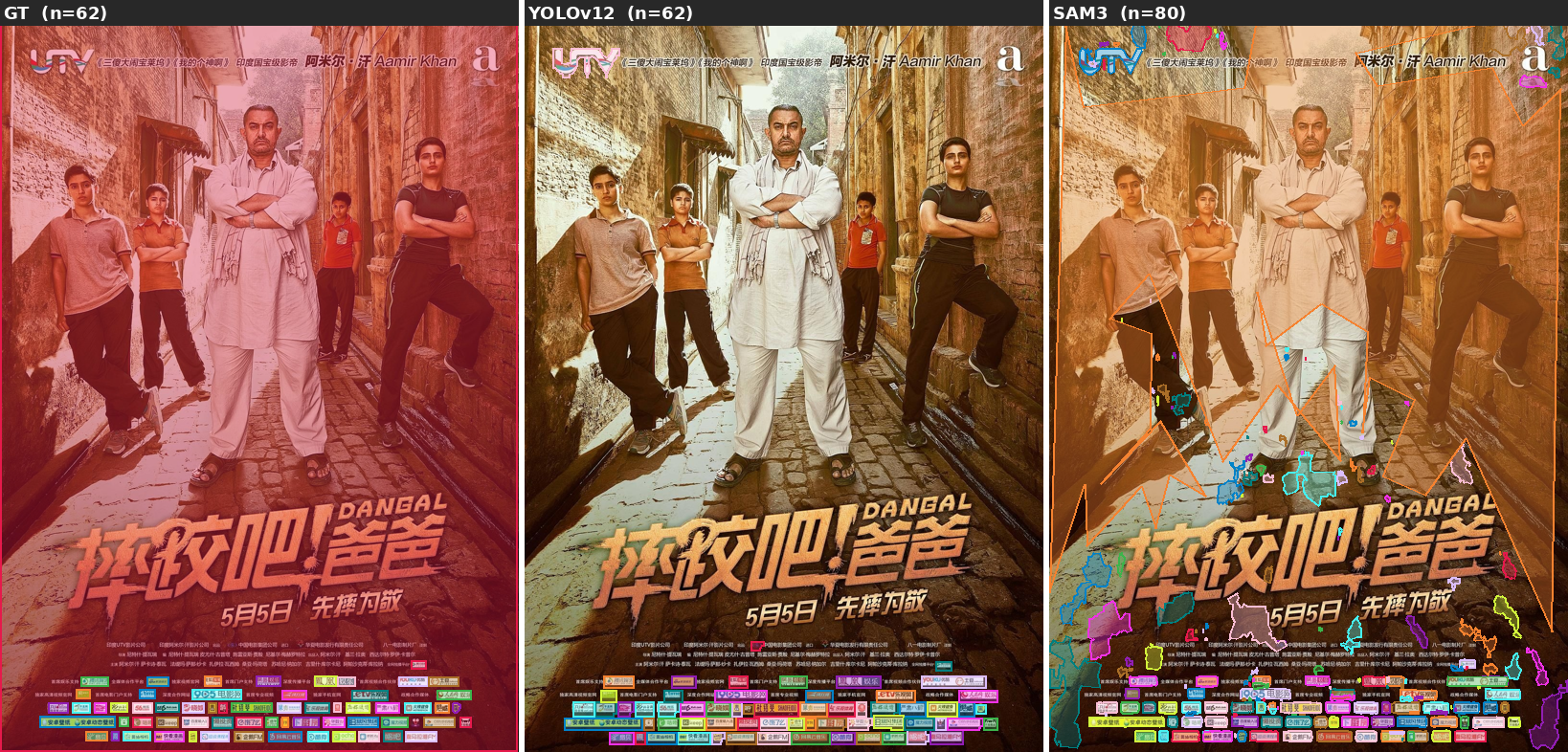}
  \caption{Qualitative comparison on a representative \emph{dense small-entity} \adopds validation page (sample \texttt{8e046669}; 63 OCR blocks; language \texttt{zh}). Left to right: ground-truth entity masks, YOLOv12-Seg, SAM3. Per-instance polygons; panel headers give instance counts.}
  \label{fig:appx-mm-8e046669}
\end{figure*}

\begin{figure*}[t]\centering
  \includegraphics[width=\textwidth]{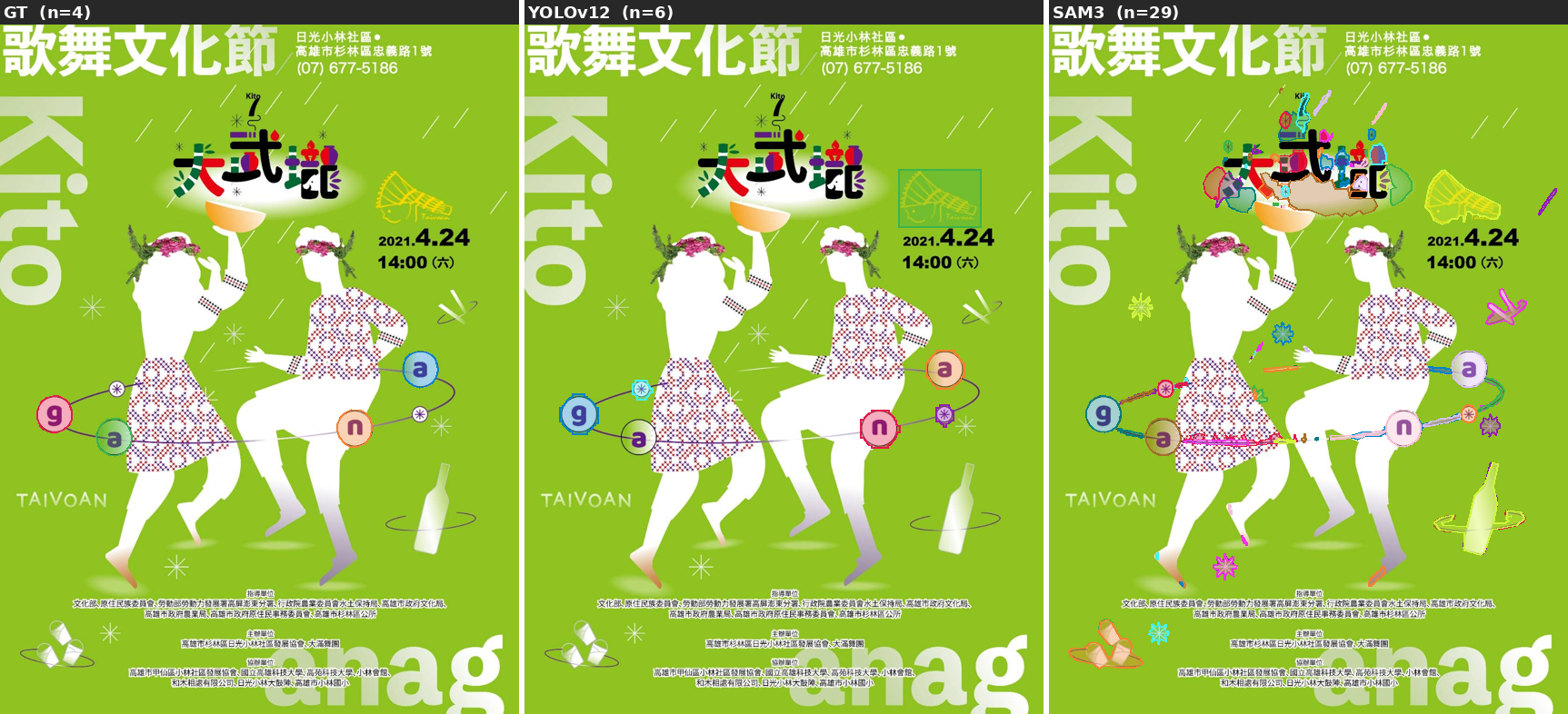}
  \caption{Qualitative comparison on a representative \emph{dense small-entity} \adopds validation page (sample \texttt{8c5bf311}; 52 OCR blocks; language \texttt{zh}). Left to right: ground-truth entity masks, YOLOv12-Seg, SAM3. Per-instance polygons; panel headers give instance counts.}
  \label{fig:appx-mm-8c5bf311}
\end{figure*}

\begin{figure*}[t]\centering
  \includegraphics[width=\textwidth]{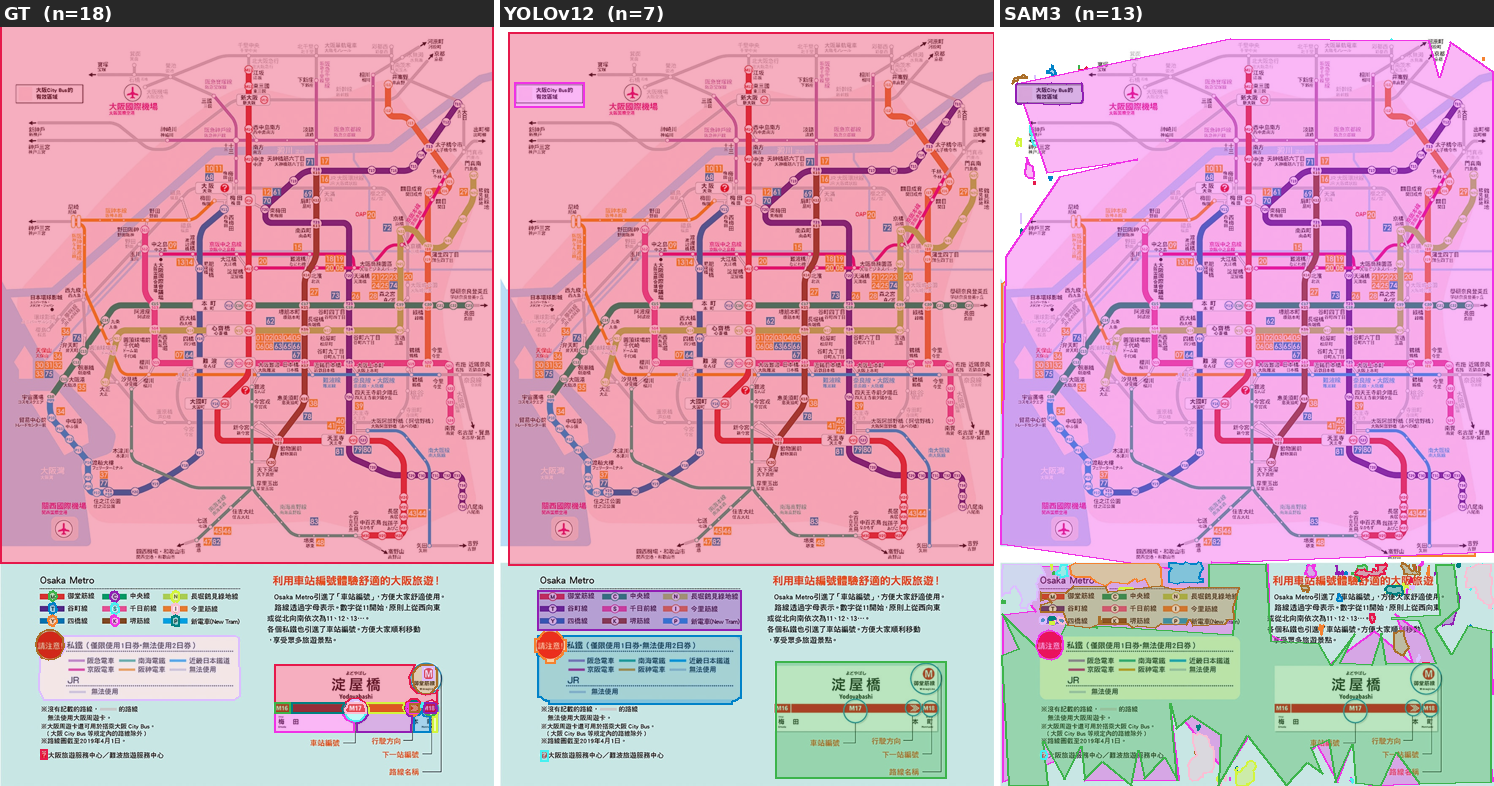}
  \caption{Qualitative comparison on a representative \emph{CJK} \adopds validation page (sample \texttt{d1919156}; 89 OCR blocks; language \texttt{zh}). Left to right: ground-truth entity masks, YOLOv12-Seg, SAM3. Per-instance polygons; panel headers give instance counts.}
  \label{fig:appx-mm-d1919156}
\end{figure*}

\begin{figure*}[t]\centering
  \includegraphics[width=\textwidth]{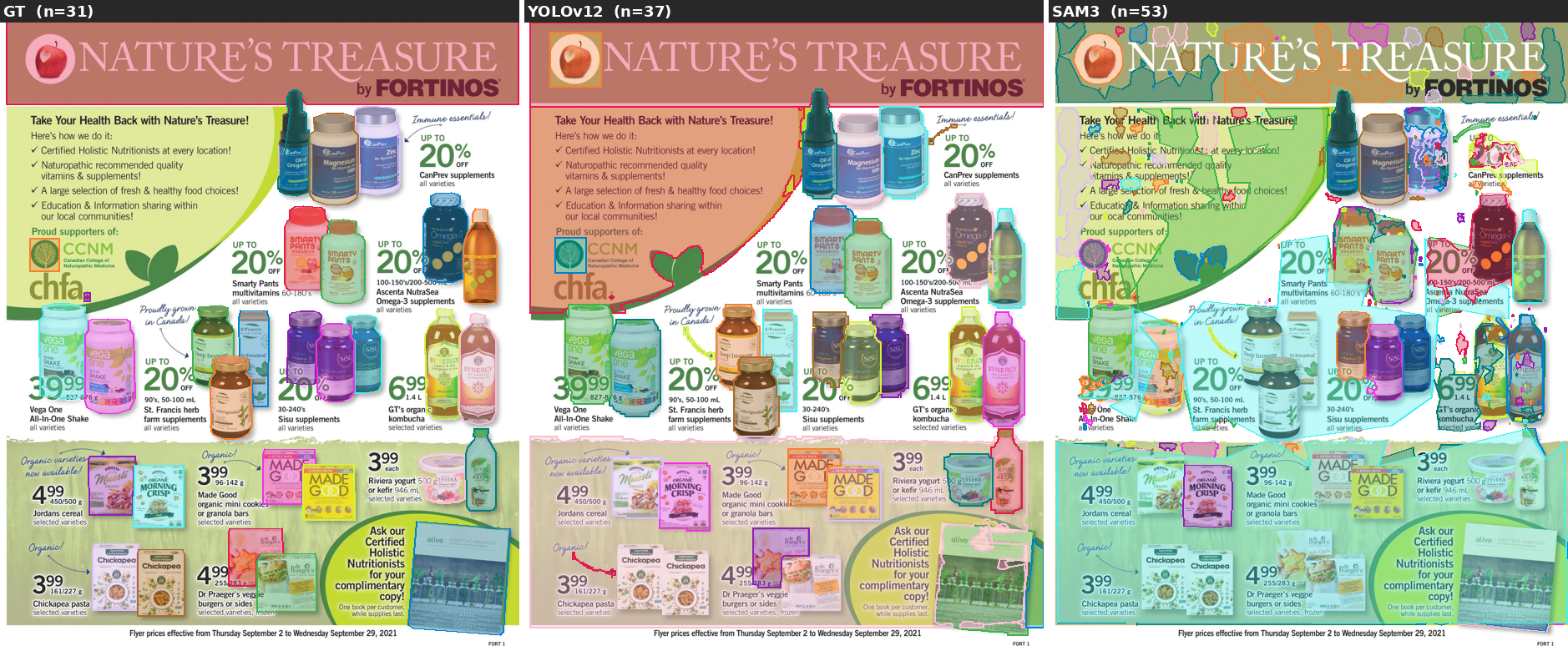}
  \caption{Qualitative comparison on a representative \emph{CJK} \adopds validation page (sample \texttt{8fe26edd}; 84 OCR blocks; language \texttt{zh}). Left to right: ground-truth entity masks, YOLOv12-Seg, SAM3. Per-instance polygons; panel headers give instance counts.}
  \label{fig:appx-mm-8fe26edd}
\end{figure*}

\begin{figure*}[t]\centering
  \includegraphics[width=\textwidth]{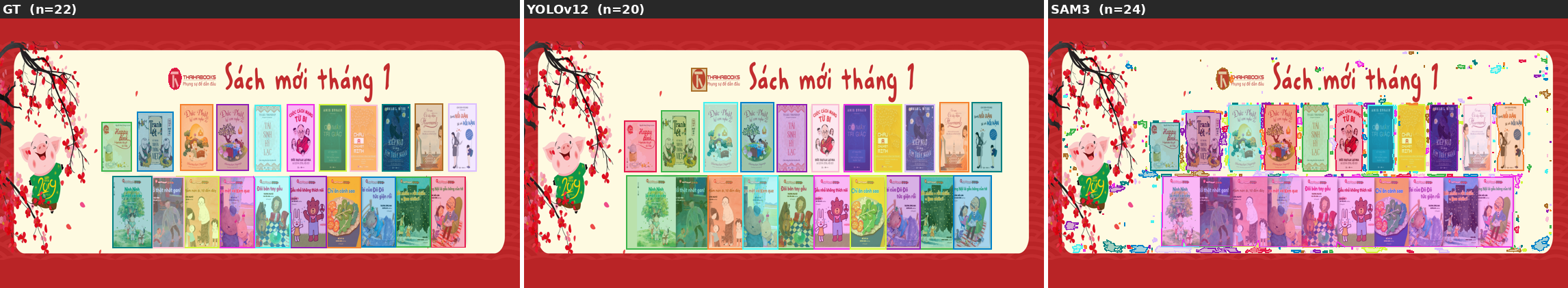}
  \caption{Qualitative comparison on a representative \emph{visual-rich} \adopds validation page (sample \texttt{62bc135c}; 69 OCR blocks; language \texttt{other}). Left to right: ground-truth entity masks, YOLOv12-Seg, SAM3. Per-instance polygons; panel headers give instance counts.}
  \label{fig:appx-mm-62bc135c}
\end{figure*}

\begin{figure*}[t]\centering
  \includegraphics[width=\textwidth]{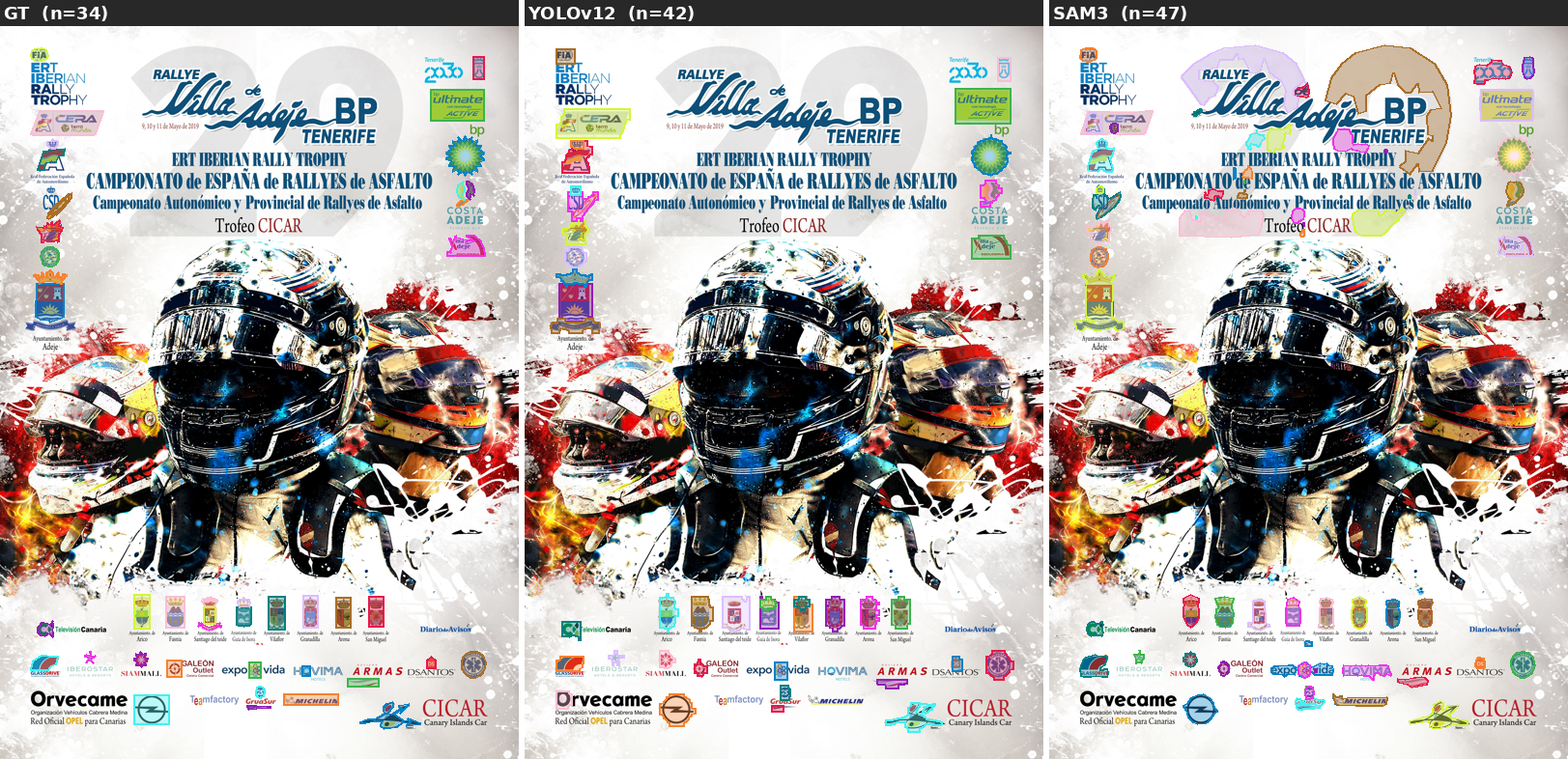}
  \caption{Qualitative comparison on a representative \emph{visual-rich} \adopds validation page (sample \texttt{568c436b}; 90 OCR blocks; language \texttt{other}). Left to right: ground-truth entity masks, YOLOv12-Seg, SAM3. Per-instance polygons; panel headers give instance counts.}
  \label{fig:appx-mm-568c436b}
\end{figure*}

\begin{figure*}[t]\centering
  \includegraphics[width=\textwidth]{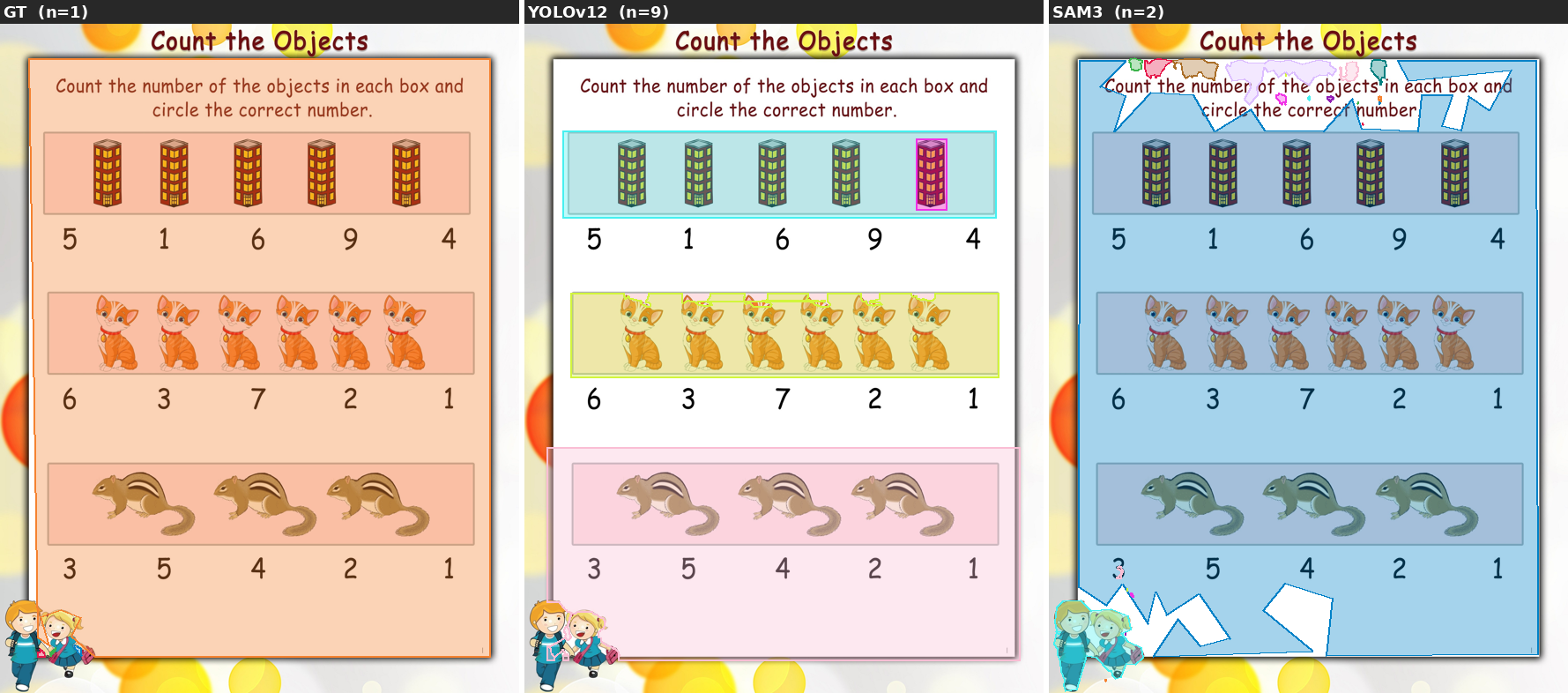}
  \caption{Qualitative comparison on a representative \emph{high model-disagreement} \adopds validation page (sample \texttt{01675a95}; 13 OCR blocks; language \texttt{zh}). Left to right: ground-truth entity masks, YOLOv12-Seg, SAM3. Per-instance polygons; panel headers give instance counts.}
  \label{fig:appx-mm-01675a95}
\end{figure*}

Figures~\ref{fig:appx-mm-ea5c1757}--\ref{fig:appx-mm-01675a95} show representative \adopds validation pages with predictions overlaid, comparing ground-truth entity masks against the fine-tuned YOLOv12-Seg and SAM3 segmenters. Each panel header lists the per-model instance count; polygons are drawn per instance. The examples span dense small-entity layouts, CJK pages, visually rich documents, and a high model-disagreement case, illustrating how mask granularity and small-entity recall differ across models on the same page.

\FloatBarrier

\subsection{LocateAnything Qualitative Comparisons}
\label{sec:locany_qual_appendix}

\begin{figure*}[!htbp]
\centering
\definecolor{mmGT}{RGB}{40,40,40}
\definecolor{mmSAM3}{RGB}{55,90,160}
\definecolor{mmRFDETR}{RGB}{150,60,40}
\definecolor{mmYOLO}{RGB}{40,110,60}
\definecolor{mmLOC}{RGB}{120,60,150}
\newcommand{\appmmchip}[2]{{\setlength{\fboxsep}{0pt}\colorbox{#1}{\makebox[0.182\linewidth][c]{\rule[-2pt]{0pt}{10pt}\normalfont\tiny\textcolor{white}{#2}}}}}
\newcommand{\appmmhdr}{%
  \appmmchip{mmGT}{GT}\appmmchip{mmSAM3}{SAM3}\appmmchip{mmRFDETR}{RF-DETR-Seg}%
  \appmmchip{mmYOLO}{YOLOv12-Seg}\appmmchip{mmLOC}{LocateAnything$^{*}$}\par}
\appmmhdr
\includegraphics[width=0.90\linewidth]{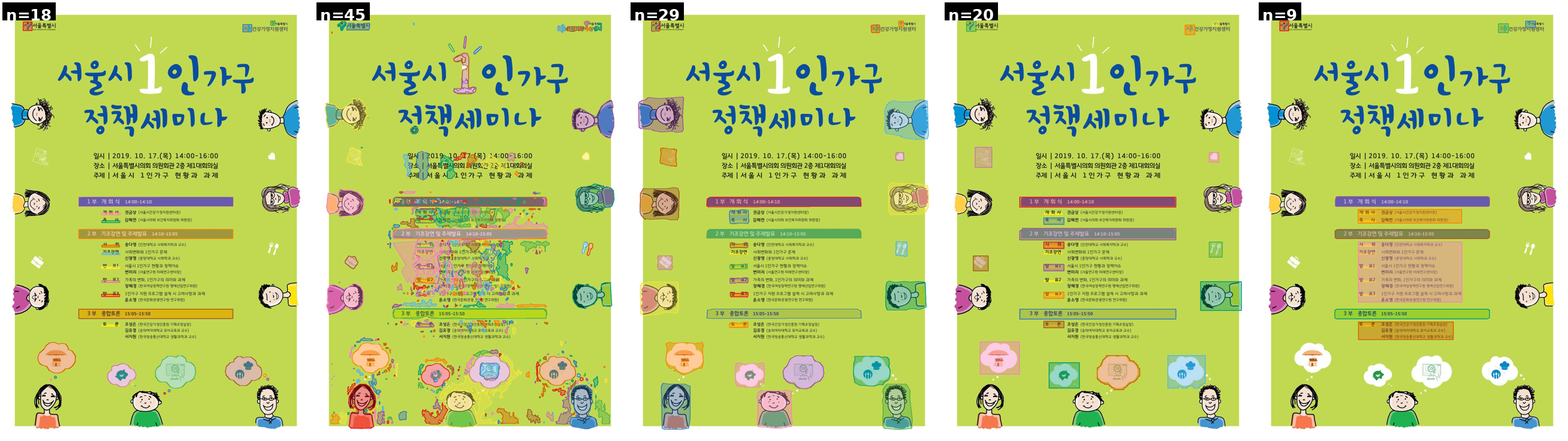}

\vspace{4pt}\appmmhdr
\includegraphics[width=0.90\linewidth]{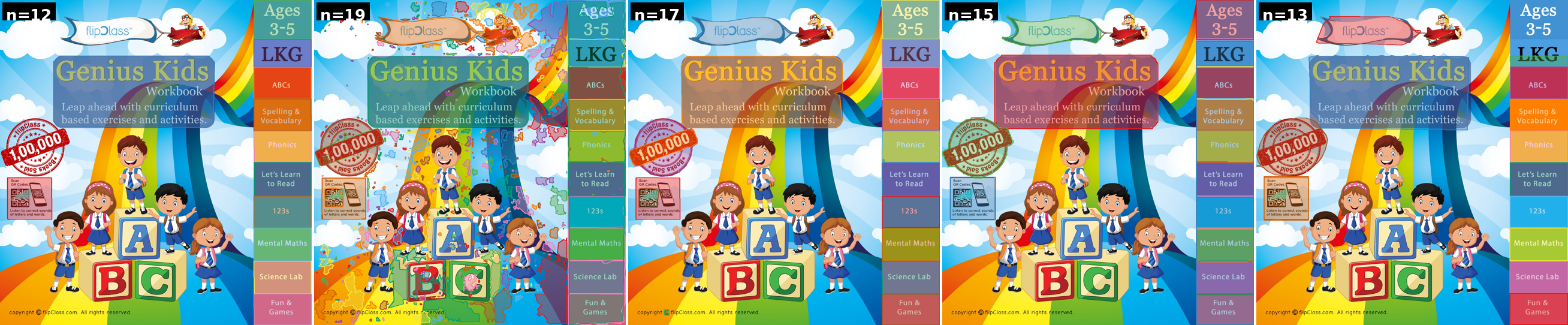}
\caption{Additional qualitative segmentation comparisons. Columns show ground truth, SAM3, RF-DETR-Seg, YOLOv12-Seg, and LocateAnything; colors denote instances and \texttt{n} gives the instance count.}
\label{fig:locany_mask_viz_appendix}
\end{figure*}

Fig.~\ref{fig:locany_mask_viz_appendix} shows the two qualitative segmentation
comparisons moved out of the main text. 
This illustrates that specialist segmenters better preserve instance
granularity after fine-tuning, while the unified grounder provides a useful but
coarser polygon reference.

\FloatBarrier

\clearpage
\newpage
\subsection{Agentic Self-Refinement: Algorithm}
\label{sec:agentic_algorithm}

The algorithm in Fig.~\ref{alg:agentic_refine} gives the full procedure behind the
workflow of Fig.~\ref{fig:agentic_workflow} (agentic self-refinement in
Sec.~\ref{sec:decomposition_analysis}). 

\begin{figure}[!htbp]
\centering
\footnotesize
\setlength{\tabcolsep}{4pt}\renewcommand{\arraystretch}{1.18}
\begin{tabular}{@{}r@{\ \ }p{0.50\columnwidth}@{\ }p{0.30\columnwidth}@{}}
\multicolumn{3}{@{}l@{}}{\textbf{Algorithm 1.} Agentic Self-Refinement (per page)}\\[1pt]
\multicolumn{3}{@{}l@{}}{\rule{\columnwidth}{0.5pt}}\\
\multicolumn{3}{@{}l@{}}{\textbf{Input:} page $I$ ($W{\times}H$), detectors $\{D_1,..,D_m\}$, VLM $\mathcal{V}$}\\
\multicolumn{3}{@{}l@{}}{\textbf{Param:} pool threshold\ $\tau{=}0.05$, NMS $\theta{=}0.6$, cap $N{=}30$}\\
\multicolumn{3}{@{}l@{}}{\rule{\columnwidth}{0.4pt}}\\
1  & $P \leftarrow \bigcup_j D_j(I,\mathrm{conf}{\ge}\tau)$           & {\scriptsize\textit{over-recall pool}}\\
2  & $P \leftarrow \textsc{Wbf}(P)$                                  & {\scriptsize\textit{fuse detectors}}\\
3  & $B \leftarrow \textsc{Nms}(P,\theta)[{:}N]$                     & {\scriptsize\textit{dedup, cap $N$}}\\
4  & \textbf{if} $|B|{<}2$ \textbf{return} $B$                       & \\
5  & $I_m \leftarrow \textsc{DrawNumbered}(I,B)$                     & {\scriptsize\textit{Set-of-Marks}}\\
6  & $t \leftarrow \mathcal{V}(I_m;\,\mathrm{stop}{=}\langle/\mathrm{think}\rangle)$ & {\scriptsize\textit{reason over marks}}\\
7  & $r \leftarrow \mathcal{V}(I_m;\,\mathrm{prefill}{=}t\,\|\,\texttt{FINAL:[[})$    & {\scriptsize\textit{structured parse}}\\
8  & $G \leftarrow \textsc{ParseGroups}(r)$                          & {\scriptsize\textit{$\emptyset$ on fail}}\\
9  & $R \leftarrow \{\},\ U \leftarrow \emptyset$                    & \\
10 & \textbf{for} $g \in G$:\ \ $S \leftarrow \{B[i]:i{\in}g\}$      & \\
11 & \quad \textbf{if} $|g|{\ge}2 \wedge \neg\,\textsc{GeoOk}(S,W,H)$: & \\
12 & \quad\quad $R \leftarrow R \cup S$                              & {\scriptsize\textit{keep separate}}\\
13 & \quad \textbf{else} $R \leftarrow R \cup \{\textsc{Union}(S)\}$  & {\scriptsize\textit{merge to 1 box}}\\
14 & \quad $U \leftarrow U \cup g$                                   & \\
15 & $R \leftarrow R \cup \{B[i]:i{\notin}U\}$                       & {\scriptsize\textit{singletons}}\\
16 & \textbf{return} $R$                                            & \\
\multicolumn{3}{@{}l@{}}{\rule{\columnwidth}{0.4pt}}\\
\multicolumn{3}{@{}p{0.97\columnwidth}@{}}{\textbf{\textsc{GeoOk}}$(S,W,H)$: \emph{reject} a ${\ge}2$-box merge if any of: (1)~some box's $x$-overlap with the union $x$-span $<0.5$ (cross-column); (2)~$\sum_b \mathrm{area}(b)/\mathrm{area}(\textsc{Union}(S)) < 0.55$ (fat-empty); (3)~union height $>0.9H$, or area $>0.6WH$, or $|S|>20$ (page-spanning); else \emph{accept}.}\\
\multicolumn{3}{@{}l@{}}{\rule{\columnwidth}{0.5pt}}\\
\end{tabular}
\caption{Agentic self-refinement. The agent ($\mathcal{V}$ = Keye-VL-2.0-30B) only \emph{groups} numbered boxes drawn on the page. It never outputs coordinates. A reason-then-parse assistant-prefill procedure yields a parseable grouping despite the reasoner's tendency to over-think, and \textsc{GeoOk} caps over-merges with a deterministic post-check.}
\label{alg:agentic_refine}
\end{figure}
\FloatBarrier

\subsection{Agentic Self-Refinement: Numerical and Qualitative Details}
\label{sec:agentic_refine_details_appendix}

\begin{figure*}[!htbp]
\centering
\begin{subfigure}{0.49\textwidth}\includegraphics[width=\linewidth]{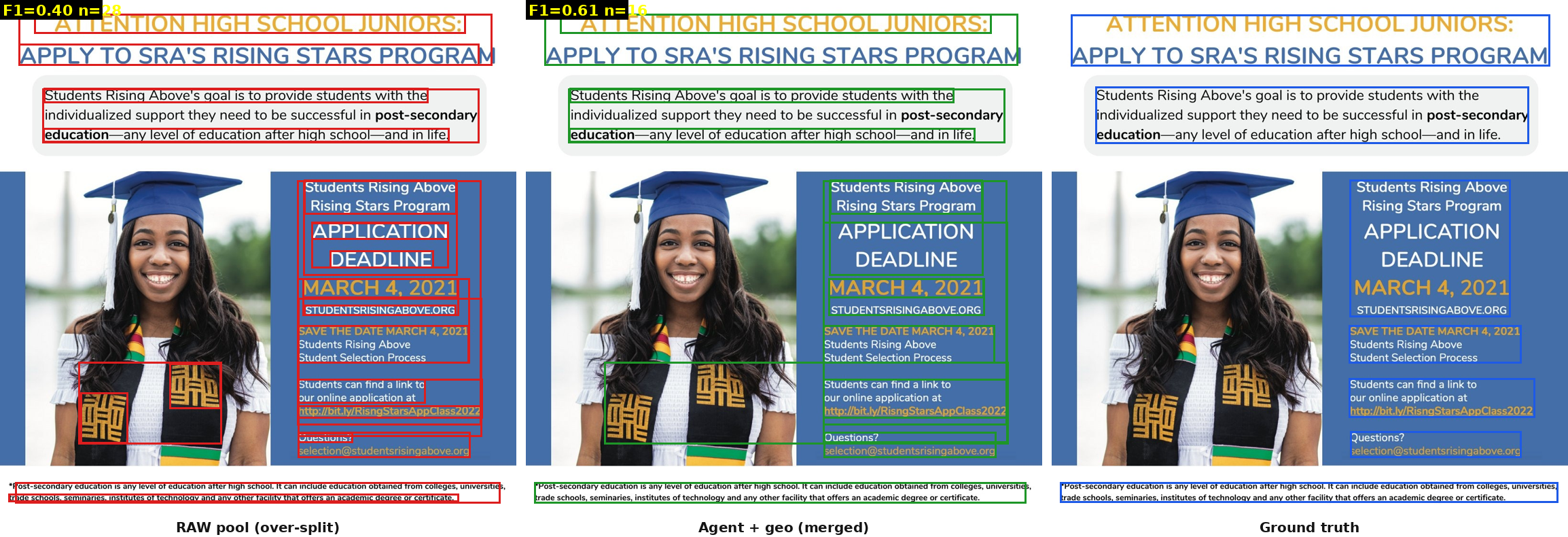}\caption{F1 $0.40\!\to\!0.61$; $28\!\to\!16$ boxes (GT 7).}\end{subfigure}\hfill
\begin{subfigure}{0.49\textwidth}\includegraphics[width=\linewidth]{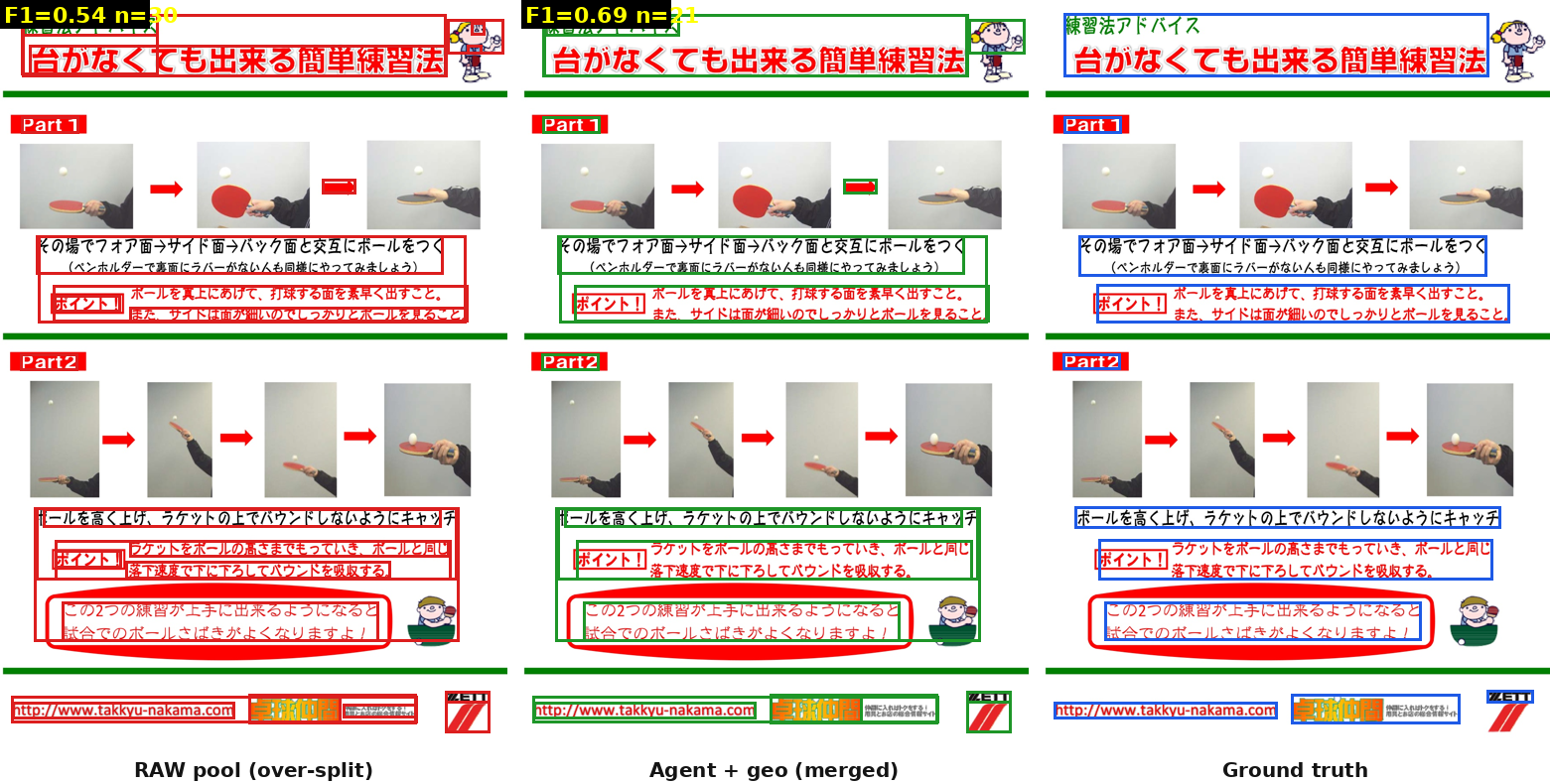}\caption{F1 $0.54\!\to\!0.69$; $30\!\to\!21$ boxes (GT 11).}\end{subfigure}

\medskip
\begin{subfigure}{0.49\textwidth}\includegraphics[width=\linewidth]{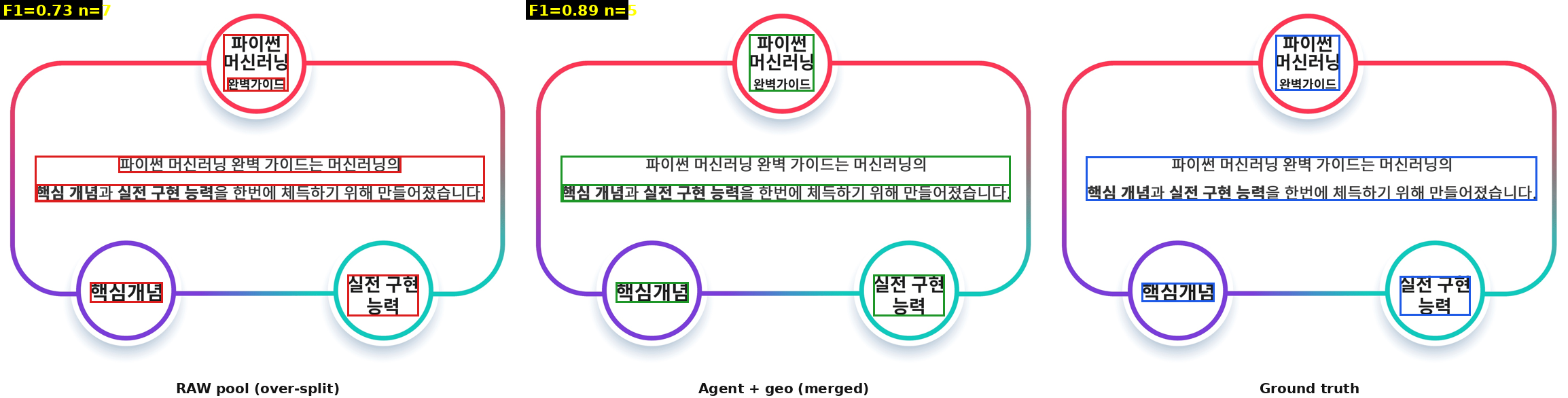}\caption{F1 $0.73\!\to\!0.89$; $7\!\to\!5$ boxes (GT 4).}\end{subfigure}\hfill
\begin{subfigure}{0.49\textwidth}\includegraphics[width=\linewidth]{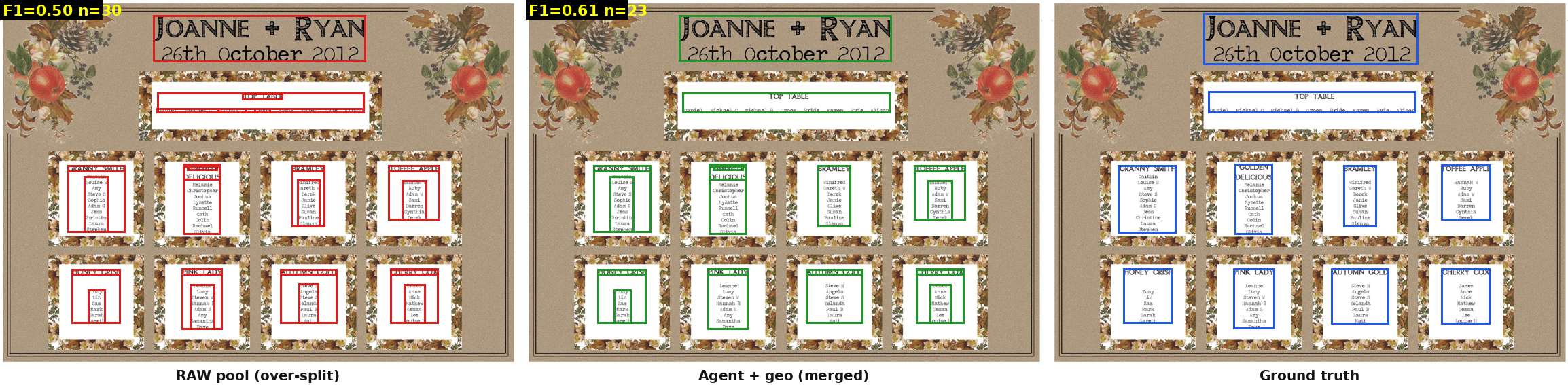}\caption{F1 $0.50\!\to\!0.61$; $30\!\to\!23$ boxes (GT 10).}\end{subfigure}
\caption{Qualitative agentic refinement (mean-F1@IoU0.5). Each panel: left = raw over-recall pool (red, over-split into many boxes), middle = after agent grouping + geo-gate (green, merged to block granularity), right = ground truth (blue).}
\label{fig:agentic_examples}
\end{figure*}

Table~\ref{tab:agentic_refine} reports the exact values behind the main-text bar trend in Fig.~\ref{fig:agentic_bar}. Fig.~\ref{fig:agentic_examples} shows the representative qualitative examples referenced in Sec.~\ref{sec:decomposition_analysis}.

\begin{table}[!htbp]
\centering
\small
\setlength{\tabcolsep}{6pt}
\caption{Agentic self-refinement on the \adopds validation split (mean-F1@IoU0.5, \%, over 19{,}382 pages). All rows share the same over-recall detection pool.}
\label{tab:agentic_refine}
\begin{tabular*}{\linewidth}{@{\extracolsep{\fill}}l c c@{}}
\toprule
Method & mean-F1 & $\Delta$ vs.\ pool \\
\midrule
RF-DETR-Large pool (raw)        & 33.6 & --- \\
\;\;+ WBF (geometric fusion)    & 34.2 & $+0.6$ \\
\;\;+ Agent grouping            & 39.6 & $+6.0$ \\
\;\;+ Agent grouping + geo      & 47.7 & $+14.1$ \\
\midrule
Merge oracle (GT upper bound)   & 61.4 & $+27.8$ \\
\bottomrule
\end{tabular*}
\end{table}

\FloatBarrier

\subsection{Agentic Self-Refinement: Additional Qualitative Examples}
\label{sec:agentic_examples_appendix}

\begin{figure*}[t]
\centering
\begin{subfigure}{0.32\textwidth}\includegraphics[width=\linewidth]{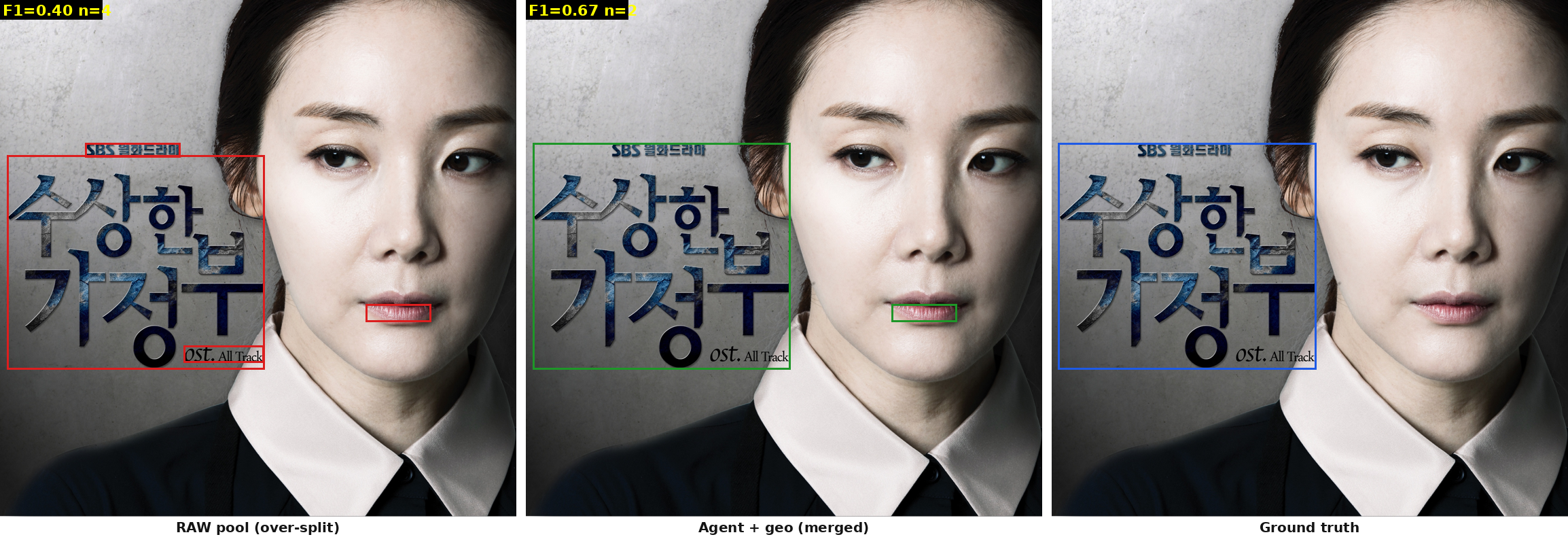}\caption{}\end{subfigure}\hfill
\begin{subfigure}{0.32\textwidth}\includegraphics[width=\linewidth]{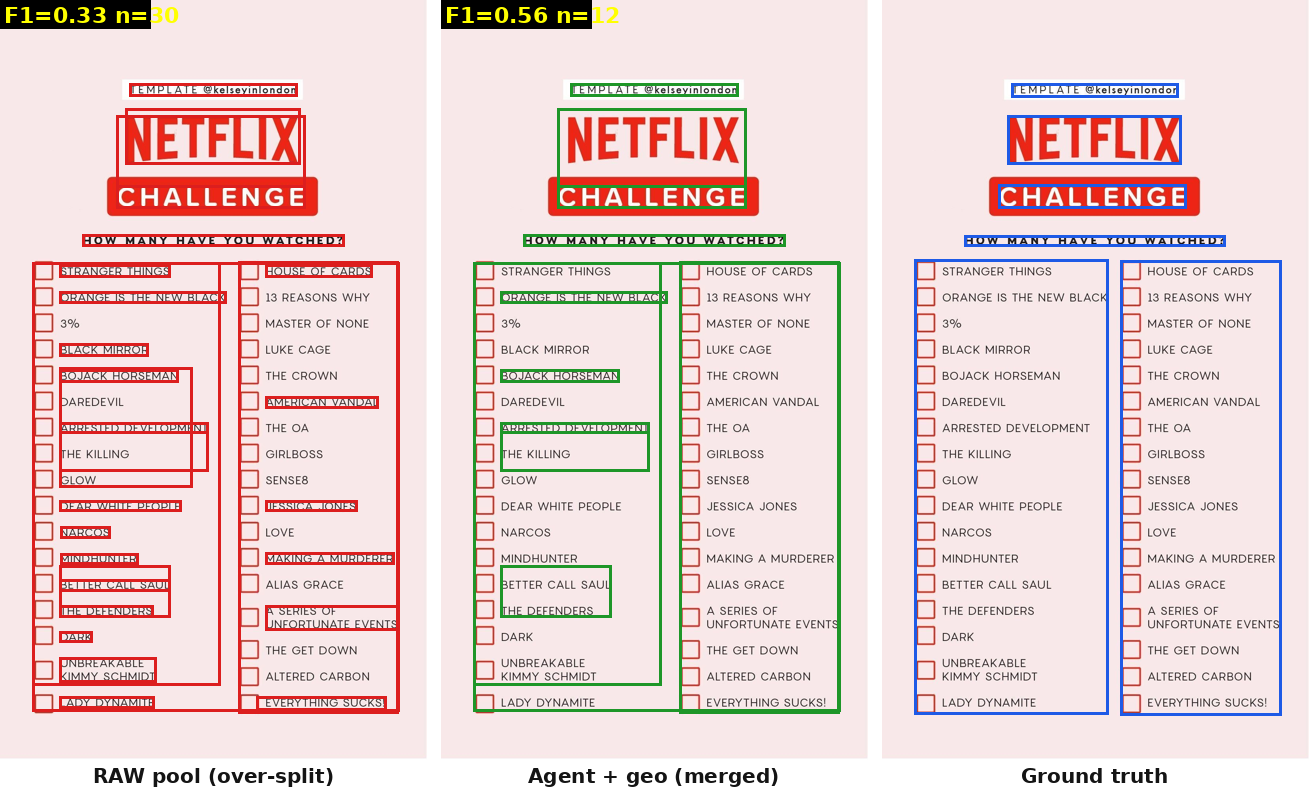}\caption{}\end{subfigure}\hfill
\begin{subfigure}{0.32\textwidth}\includegraphics[width=\linewidth]{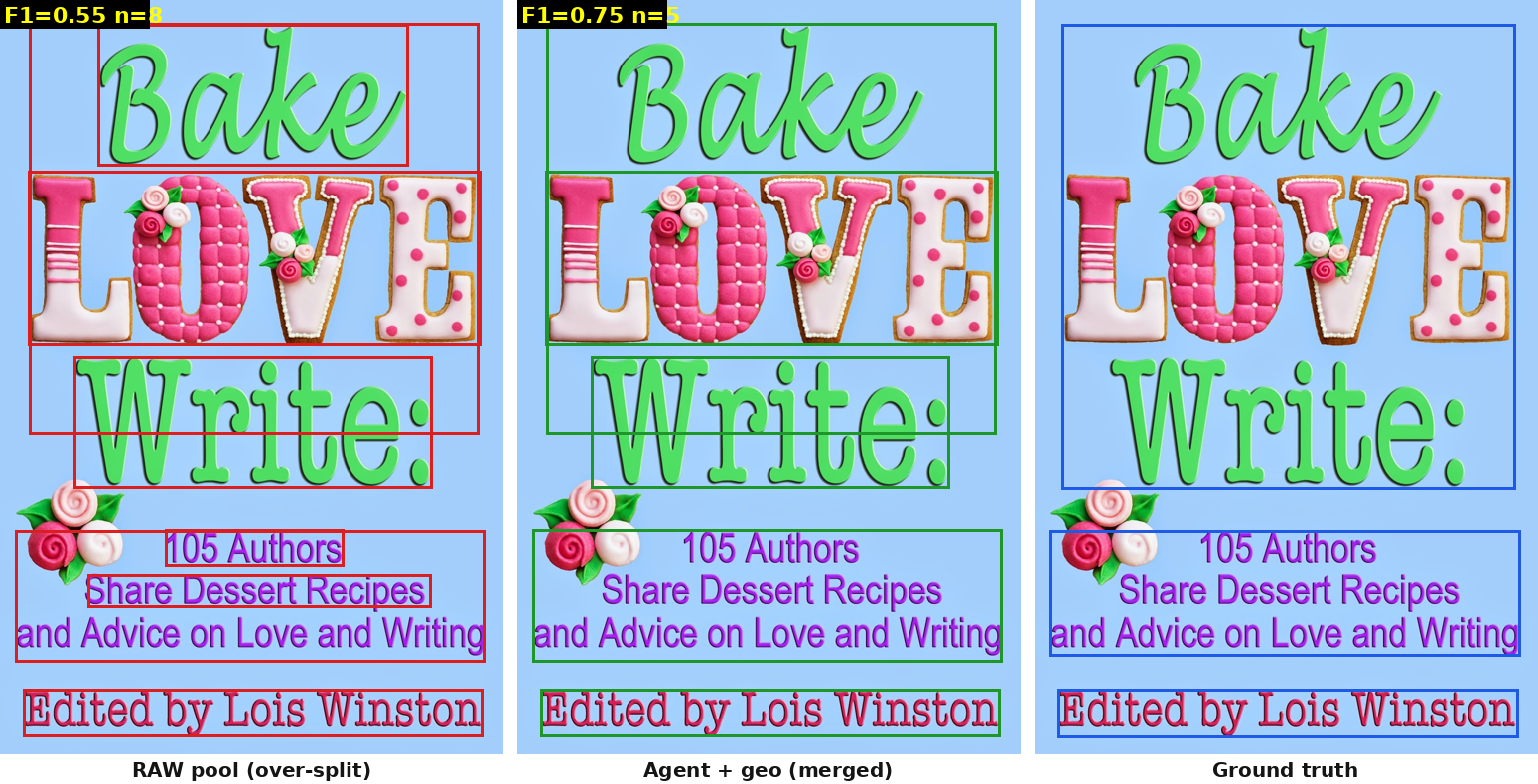}\caption{}\end{subfigure}

\medskip
\begin{subfigure}{0.32\textwidth}\includegraphics[width=\linewidth]{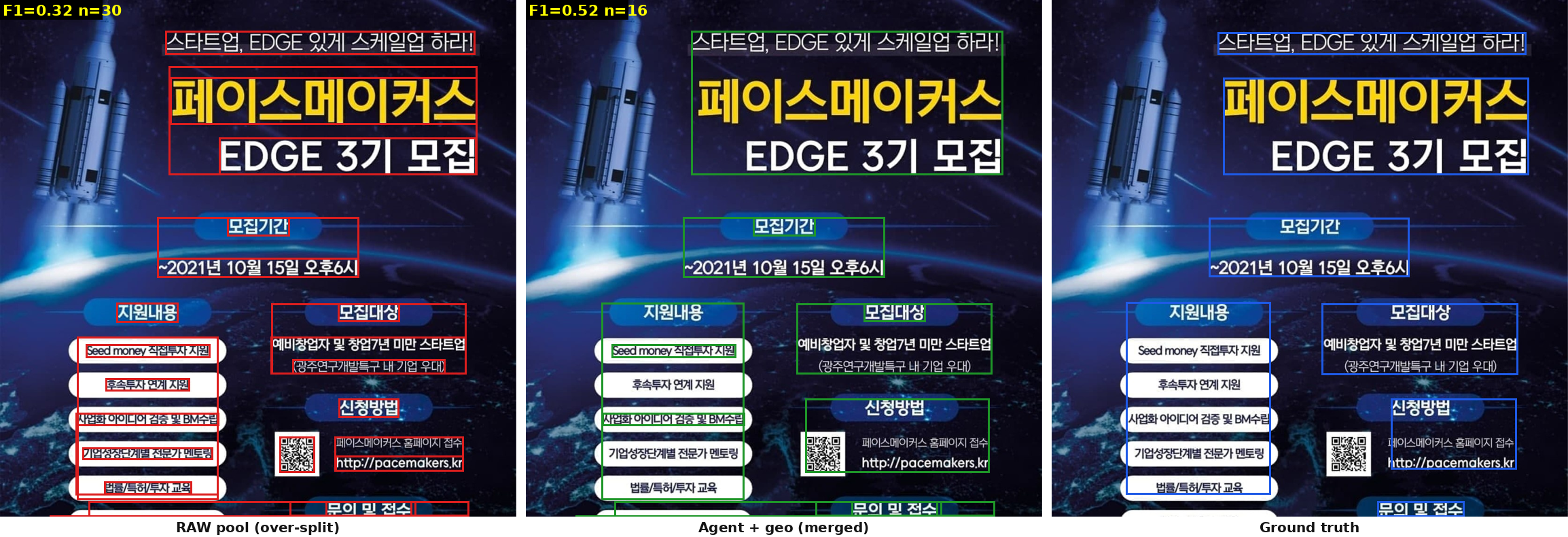}\caption{}\end{subfigure}\hfill
\begin{subfigure}{0.32\textwidth}\includegraphics[width=\linewidth]{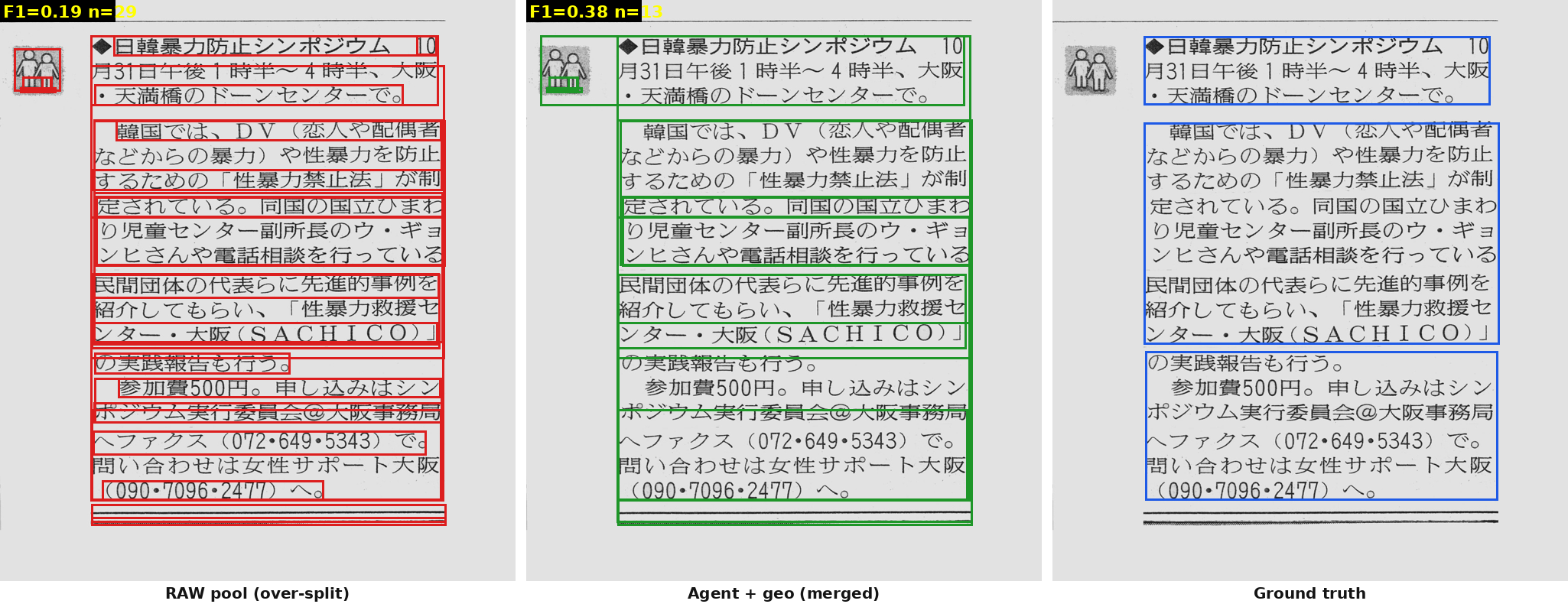}\caption{}\end{subfigure}\hfill
\begin{subfigure}{0.32\textwidth}\includegraphics[width=\linewidth]{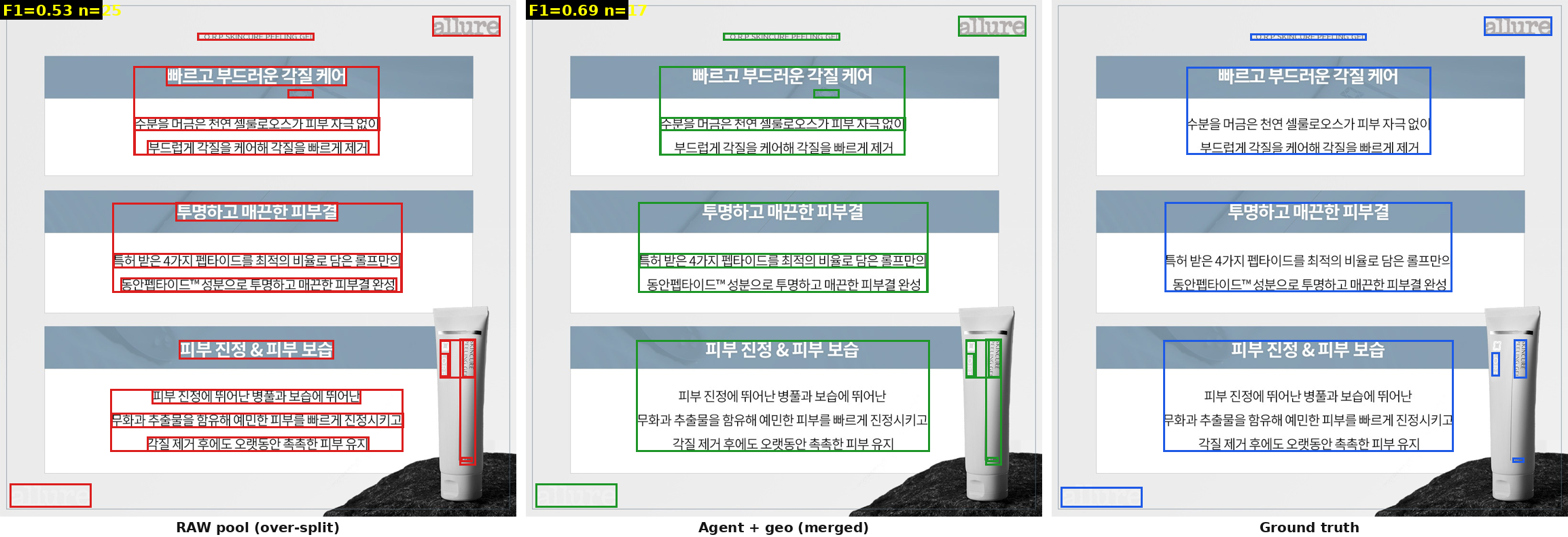}\caption{}\end{subfigure}
\caption{Additional agentic self-refinement examples (RAW over-recall pool | agent$+$geo merged | ground truth). Each triptych is annotated with its per-page mean-F1 and box count. The agent merges per-line/per-item detector fragments into coherent text blocks across diverse layouts.}
\label{fig:agentic_appendix}
\end{figure*}

Fig.~\ref{fig:agentic_appendix} shows six further pages where the agent's grouping
improves mean-F1, spanning posters, book covers, flyers, and product pages. In each
panel the detector's over-recall pool (left, red) over-splits each block into many
boxes; the agent merges them to annotation granularity (middle, green) and the
geometric guard keeps the result close to the ground truth (right, blue).

\noindent\textbf{Agent conversation traces (figure-aligned).}
Figs.~\ref{fig:agentic_trace1}--\ref{fig:agentic_trace5} pair each page's
\emph{numbered input} (left panel: each detection box a distinct color with its
index, so the box IDs in the dialogue are legible) with the model's actual exchange:
the shared instruction, the Pass-1 chain of thought (abridged), and the Pass-2
\texttt{FINAL} grouping that is parsed and merged (subject to the geometric guard).
The agent reasons over the numbered boxes it sees, a discrete decision it makes
reliably, rather than regressing coordinates. The shared prompt is:
\emph{``The image is a document page with detected boxes drawn and NUMBERED.
Detectors often OVER-SPLIT one real text block into several boxes\dots\ group the
numbered boxes so that each group is ONE text block a human would annotate
together\dots\ Output ALL groups as a JSON list of lists, on a final line:}
\texttt{FINAL: [[1,2,3],[4],[5,6]]}\emph{''}, with user turn \emph{``There are $N$
numbered boxes (0..$N{-}1$). Group them.''} plus the left-panel image.

\newcommand{\tracefig}[7]{%
  \begin{figure}[t]\centering
  \includegraphics[width=\linewidth]{data/#1}\\[3pt]
  {\footnotesize\fbox{\begin{minipage}{0.95\columnwidth}\scriptsize\setlength{\parindent}{0pt}
  \textbf{#6} (#2)\\[2pt]
  {\ttfamily\textless think\textgreater} ``#3'' {\ttfamily\textless/think\textgreater}\\[1pt]
  \textbf{FINAL:} \texttt{#4}
  \end{minipage}}}
  \caption{Aligned trace: numbered input (colored boxes) $\mid$ agent$+$geo merged $\mid$ GT, with the agent's reasoning and parsed grouping. #5}
  \label{#7}
  \end{figure}}

\tracefig{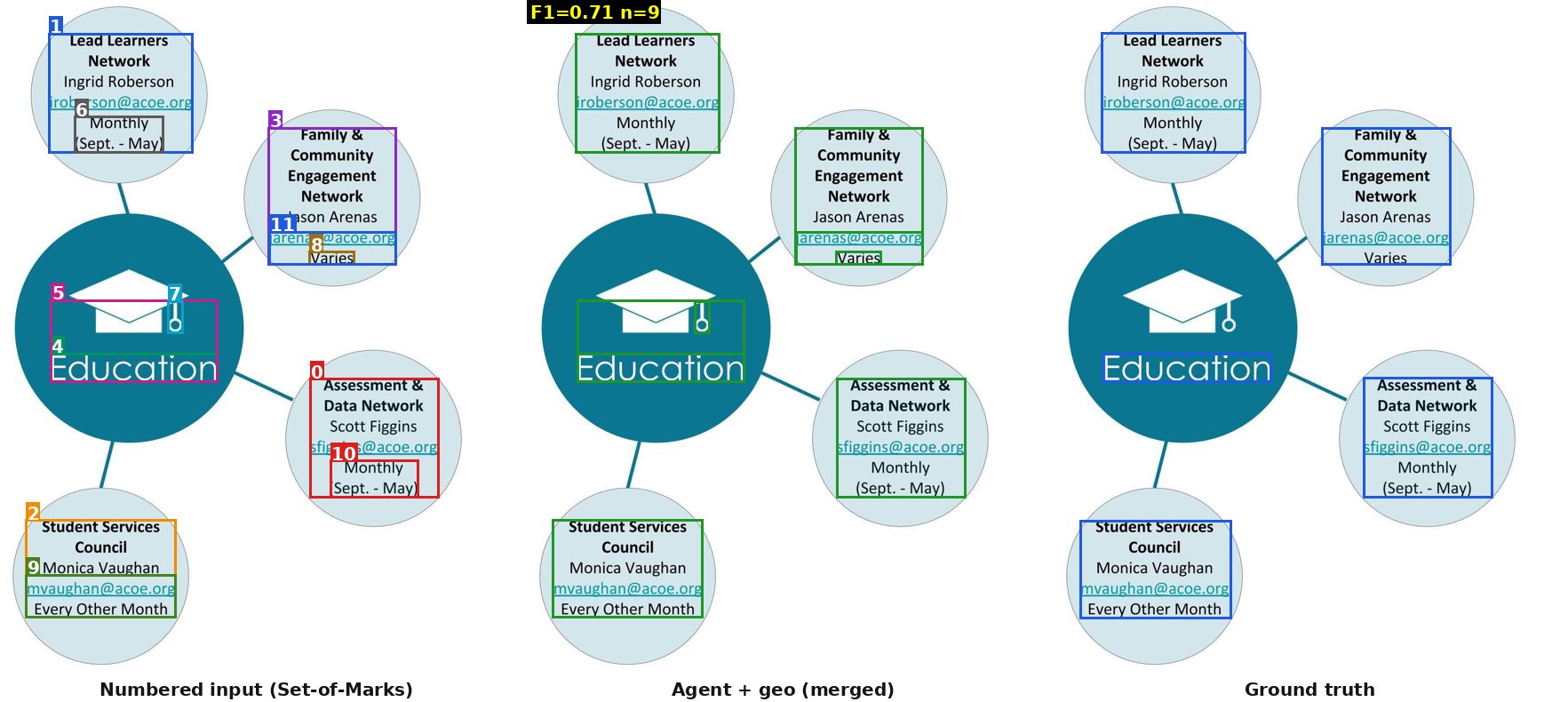}{education infographic, $12{\to}9$ boxes, F1 $0.59{\to}0.71$}{Box 1 is the ``Lead Learners Network'' title and box 6 is its ``Monthly Sept.--May'' schedule, so they are the same block $\to$ [1,6]; similarly the other three network panels each pair their title box with their schedule box \dots}{[[0,10],[1,6],[2,9],[3,7,8,11]]}{The four network panels are each recovered as one block.}{Example~A}{fig:agentic_trace1}

\tracefig{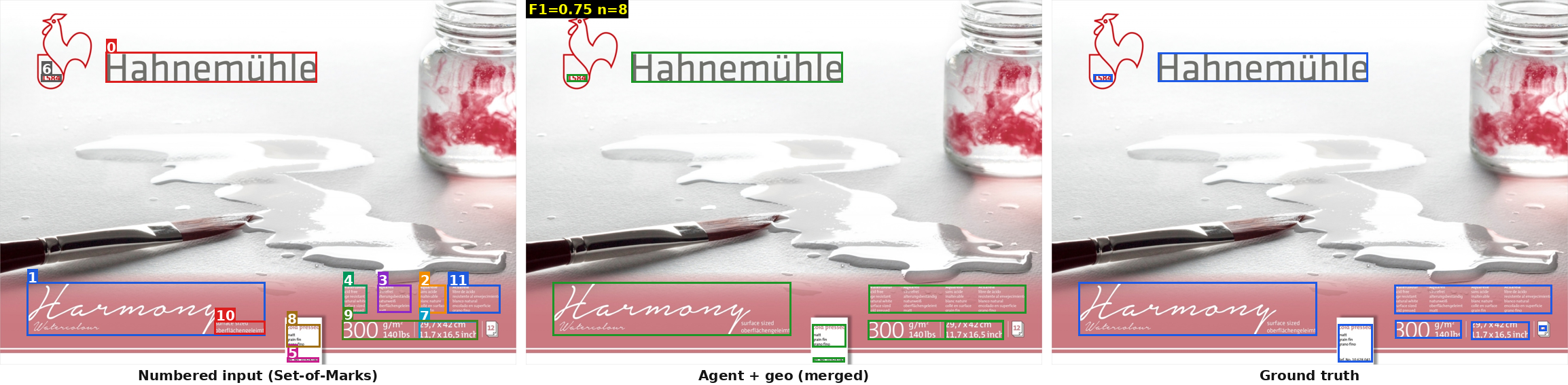}{art-supply ad, $12{\to}8$ boxes, F1 $0.50{\to}0.75$}{the ``Harmony'' product wordmark and its descriptor lines (boxes 5--9) form one block; boxes 2,3,4,11 are the spec table grouped together; box 0 is the separate brand logo \dots}{[[0],[1,10],[2,3,4,11],[5,6,7,8,9]]}{Product title, spec table, and logo are separated correctly.}{Example~B}{fig:agentic_trace5}

\tracefig{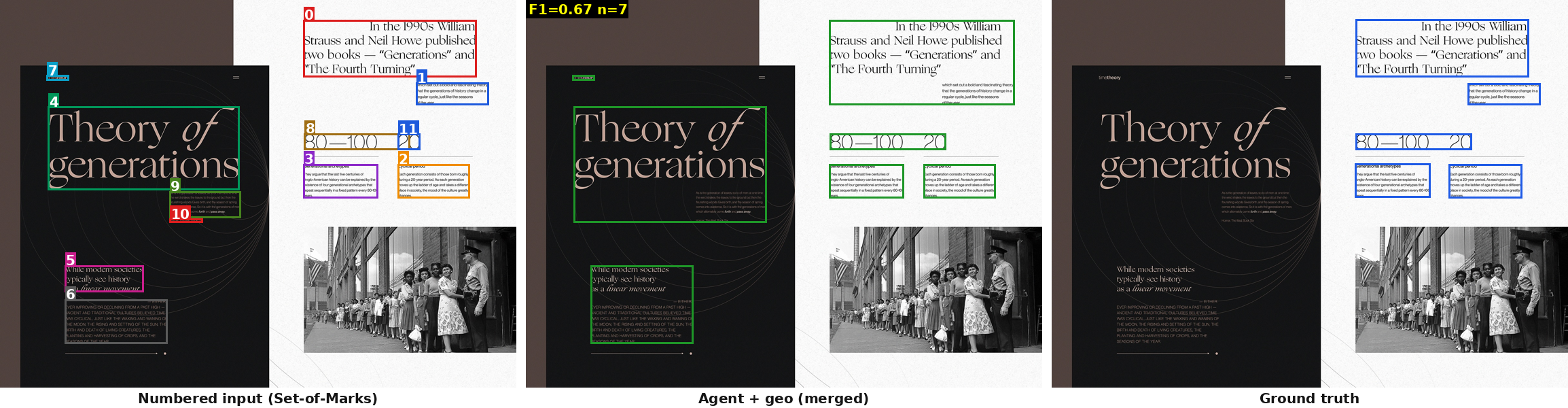}{newsletter, $12{\to}7$ boxes, F1 $0.59{\to}0.67$}{boxes 0 and 1 are the stacked heading lines $\to$ one block; boxes 4,9,10 are the body paragraph split across lines; boxes 5,6 are a sub-item pair \dots}{[[0,1],[2],[3],[4,9,10],[5,6],[7],[8,11]]}{Stacked heading and a multi-line paragraph are each merged.}{Example~C}{fig:agentic_trace3}

\tracefig{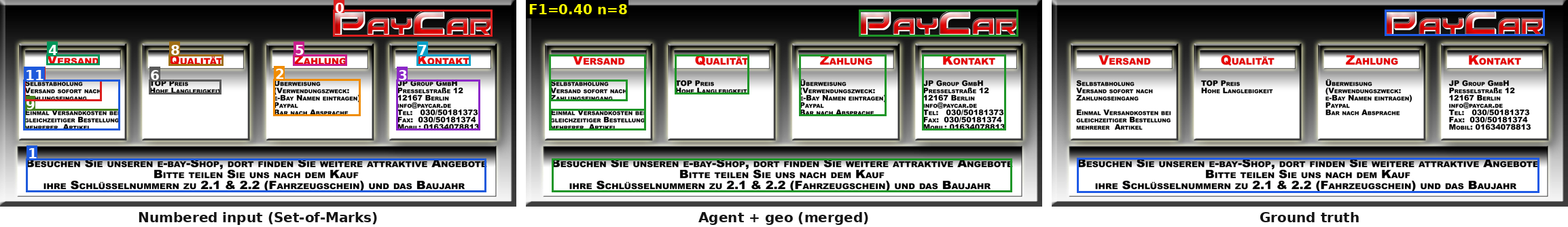}{poster, $12{\to}8$ boxes, F1 $0.29{\to}0.40$}{boxes 2 and 5 are the two lines of one title $\to$ merge; boxes 3,7 and 4,11 and 6,8 are each a fragmented caption \dots}{[[0],[1],[2,5],[3,7],[4,11],[6,8]]}{Several two-line fragments are re-joined.}{Example~D}{fig:agentic_trace4}

\tracefig{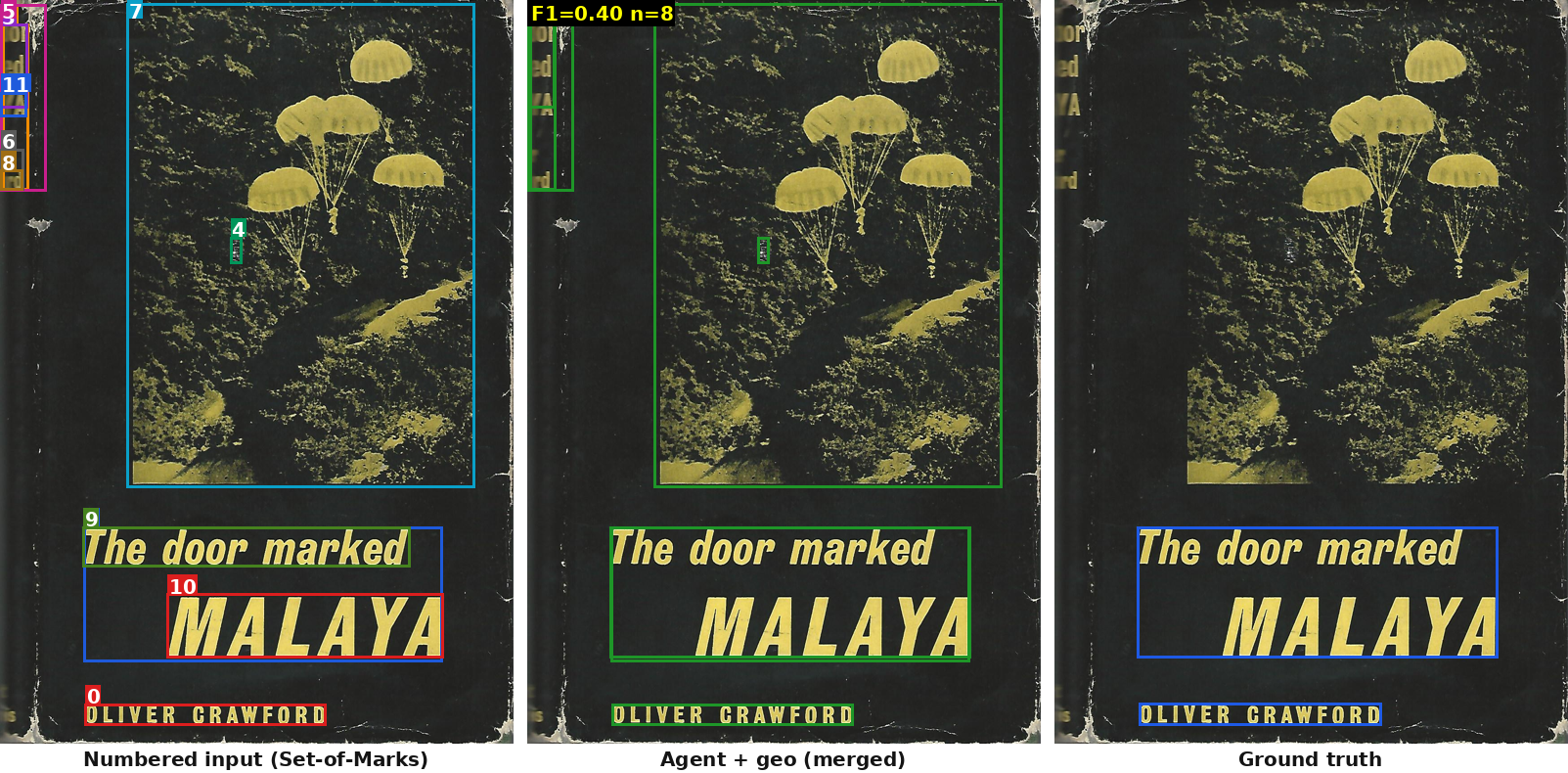}{flyer, $12{\to}8$ boxes, F1 $0.29{\to}0.40$}{boxes 5,6,8,11 are the body text broken into lines and belong together; boxes 9,10 are a footer pair; box 0 is the standalone title \dots}{[[5,11,6,8],[9,10],[0]]}{A line-split body block is recomposed.}{Example~E}{fig:agentic_trace2}

\subsection{LocateAnything Decoding: Speculative Decoding with an Autoregressive Verifier}
\label{sec:locany_specdecode}

The built-in \emph{hybrid} and \emph{fast} decode modes (the LocateAnything unified-grounding paragraph in Sec.~\ref{sec:decomposition_analysis}, Table~\ref{tab:reason_locany_speed}) accept or reject a drafted box frame with \emph{fixed heuristics} (token-confidence and well-formedness thresholds), never checking the draft against what autoregression would actually produce---which is exactly why a single bad token can derail a whole region and why \emph{fast}/\emph{hybrid} collapse on the long polygon sequences of segmentation. A principled alternative is classical speculative decoding: keep the MTP forward as the cheap \emph{drafter}, but \emph{verify} its $k$ proposed tokens with one autoregressive forward and accept token $t$ only when its verifier probability clears an acceptance threshold,
\begin{equation}
p_{\text{AR}}(t)\ge\tau\cdot\max_j p_{\text{AR}}(j).
\end{equation}
At $\tau{=}1$ this accepts a drafted token only when it equals the AR argmax, so the emitted sequence is provably AR-equivalent (quality is unchanged by construction); lowering $\tau$ accepts more aggressively, trading fidelity for the longer accepted runs that buy speed---an explicit, continuous quality knob the built-in modes lack. We implemented this verifier on top of the released model (token-level, task-agnostic) and swept $\tau$ on both tasks.

\noindent\textbf{Detection.}
On a 100-image detection subset ($k{=}6$, Table~\ref{tab:reason_locany_spec}) the verifier behaves exactly as the theory predicts: $\tau{=}1$ reproduces AR box mAP ($42.3$ vs.\ $42.2$), and box AP stays within noise down to $\tau{=}0.4$ before collapsing at $\tau{=}0$ (unconditional acceptance $=$ pure MTP). But the measured speedup is marginal ($0.98$--$1.10\times$): the MTP drafter is accepted only $37$--$46\%$ of the time, and at that acceptance rate the extra verifier forward per step cancels the parallel-draft savings.

\noindent\textbf{Segmentation.}
We expected speculative decoding to be most useful here---segmentation is where heuristic \emph{hybrid} collapses (mask AP $23.1\rightarrow5.9$, Table~\ref{tab:reason_locany_speed}), so a lossless verifier ($\tau{=}1$) should recover the full $23.1$. It does, but at no useful speed: the polygon drafter's acceptance rate ($44$--$50\%$) is no better than detection's, so $\tau{=}1$ runs at $1.02\times$ (Table~\ref{tab:reason_locany_spec_seg})---a lossless result that is exactly as slow as plain autoregression, hence pointless. A $1.24\times$ speedup appears only at $\tau{=}0.4$, where polygon AP (far less token-tolerant than boxes) would already be degraded.

\begin{table}[t]
\centering
\scriptsize
\setlength{\tabcolsep}{6pt}
\caption{True speculative decoding (MTP drafts, autoregressive verifier) on a 100-image \emph{detection} subset, $k{=}6$. The acceptance threshold $\tau$ is a continuous quality knob: $\tau{=}1$ is provably AR-equivalent (box mAP matches the AR baseline), and quality degrades gracefully until $\tau{=}0$ (accept every draft $=$ pure MTP) collapses it. The speedup is marginal because the MTP drafter's acceptance rate is only $37$--$46\%$; the verifier's extra forward per step offsets the parallel-draft savings, so the heuristic hybrid of Table~\ref{tab:reason_locany_speed} remains the better operating point on this model.}
\label{tab:reason_locany_spec}
\begin{tabular*}{\textwidth}{@{\extracolsep{\fill}}l r r r}
\toprule
Decoder & box mAP & Accept rate & Speedup \\
\midrule
autoregressive (ref)      & 42.2 & --   & 1.0$\times$ \\
spec.\ $\tau{=}1.0$       & 42.3 & 0.37 & 0.98$\times$ \\
spec.\ $\tau{=}0.7$       & 41.8 & 0.41 & 1.03$\times$ \\
spec.\ $\tau{=}0.4$       & 42.1 & 0.46 & 1.10$\times$ \\
spec.\ $\tau{=}0.0$       & 0.2  & 1.00 & 1.37$\times$ \\
\bottomrule
\end{tabular*}
\end{table}

\begin{table}[t]
\centering
\scriptsize
\setlength{\tabcolsep}{6pt}
\caption{Speculative decoding on a \emph{segmentation} subset ($k{=}6$; speed and acceptance rate). $\tau{=}1$ is lossless but offers no speedup ($1.02\times$)---it is exactly as slow as autoregression---because the polygon drafter's acceptance rate ($44$--$50\%$) is too low to amortize the verifier forward. A $1.24\times$ speedup requires $\tau{=}0.4$, where polygon AP would already be degraded. So segmentation has no useful operating point either: plain slow decoding remains the recipe.}
\label{tab:reason_locany_spec_seg}
\begin{tabular*}{\textwidth}{@{\extracolsep{\fill}}l r r}
\toprule
Decoder & Accept rate & Speedup \\
\midrule
autoregressive (ref) & --   & 1.0$\times$ (8.72\,s/img) \\
spec.\ $\tau{=}1.0$  & 0.44 & 1.02$\times$ \\
spec.\ $\tau{=}0.7$  & 0.46 & 0.95$\times$ \\
spec.\ $\tau{=}0.4$  & 0.50 & 1.24$\times$ \\
\bottomrule
\end{tabular*}
\end{table}

\noindent\textbf{Takeaway.}
Speculative decoding only pays off once the drafter's acceptance rate is high; this 3B model's MTP head reaches only $37$--$50\%$ on either task, so a correctness-preserving verifier never amortizes its extra forward. The cheap heuristic \emph{hybrid} ($2.2\times$ at $99\%$ box AP) therefore remains the best operating point for detection, and plain \emph{slow} decoding for segmentation. Closing this gap would require a higher-acceptance drafter (\eg\ a trained EAGLE/Medusa-style head) or fusing draft and verify into a single forward (self-speculation); both are out of scope here.

\clearpage
\newpage
\subsection{Complete Definitions and Counting Instructions for Target Labels in the \docount Evaluation Corpus}
\label{twvp_def_block}

\begin{tcolorbox}[
    title={Target Label Definitions},
    colback=gray!3,
    colframe=gray!55,
    boxrule=0.4pt,
    arc=1mm,
    left=2mm,
    right=2mm,
    top=1mm,
    bottom=1mm
]
\begin{description}[
    style=nextline,
    leftmargin=0pt,
    labelsep=0pt,
    font=\normalfont\bfseries
]

\item[Brand Logo]
A visual identity mark used to identify a brand, organization, product, service,
certification, standard, award, or official program. It may appear as a
logomark, emblem, seal, badge, stylized wordmark, designed text, certification
mark, quality seal, or endorsement badge. A Brand Logo can be the designed mark
alone, the mark with its associated name, or the identifying text alone when it
functions as the visual identifier. Generic decorative icons are excluded.

Count one Brand Logo as one distinct visual identity instance. If a designed
mark and its associated brand name appear together as one logo unit for the same
brand, count them as one Brand Logo. If only a designed mark appears, count that
designed mark as one Brand Logo. If a stylized wordmark is the brand identifier,
count that stylized wordmark as one Brand Logo.
\\

\item[Photograph]
Realistic or naturalistic image content depicting real-world people, objects,
places, products, scenes, textures, and related content. It may appear as a
foreground image or as a background image. Flat illustrations, icons, abstract
graphics, patterns, charts, tables, and ordinary text are excluded.
\\

\item[Table]
A grid-like arrangement of information organized into rows, columns, or cells
for comparison or lookup. Visible grid lines are not required. General
page-layout grids and decorative aligned blocks that are not meant to function
as tables are excluded.

Menu price lists count as tables when item names and prices or options are
arranged in consistent rows and aligned columns, such as item-name columns with
corresponding price columns. Menu item grids or product cards do not count as
tables when the grid mainly serves a page-layout purpose.
\\

\item[Chart / Graph]
Count each self-contained quantitative or categorical data visualization as one
chart/graph. A chart/graph represents data values using visual encodings such as
position, length, area, color scale, angle, bars, lines, points, slices, heatmap
cells, contours, map regions, or network nodes and edges.

Count one chart/graph for each independent visualization unit. A unit is
independent if it presents its own dataset, metric, coordinate, radial, or
geographic layout, or analytical result. Count small multiples as multiple
charts when each repeated panel is a self-contained visualization with its own
data display. Count them as one chart only when the repeated elements are
clearly components of a single shared comparison structure.

\end{description}
\end{tcolorbox}

\clearpage
\subsection{\docount Grounded-Narration Qualitative Examples}
\label{sec:reason_qualitative}

\definecolor{promptblue}{HTML}{F0F6FD}
\definecolor{markerorange}{HTML}{FF8C2A}
\definecolor{markergreen}{HTML}{62B33A}
\definecolor{softgray}{HTML}{777777}

\graphicspath{{data/twvp/}}

\newtcolorbox{CountingBox}[1]{
  enhanced,
  breakable=false,
  width=\linewidth,
  colback=white,
  colframe=black,
  boxrule=0.6pt,
  sharp corners,
  left=8pt,
  right=8pt,
  top=12pt,
  bottom=8pt,
  title={#1},
  fonttitle=\bfseries\large,
  coltitle=white,
  colbacktitle=black,
  boxed title style={
    sharp corners,
    boxrule=0pt,
    left=18pt,
    right=18pt,
    top=4pt,
    bottom=4pt,
    colback=black
  },
  attach boxed title to top left={xshift=-0.6pt,yshift=0.4pt}
}

\newcommand{\Trigger}{\textcolor{softgray}{\bfseries [Trigger\_Placeholder]}}
\newcommand{\vpo}[1]{\textcolor{markerorange}{\bfseries\texttt{[[#1]]}}}
\newcommand{\vpt}{\textcolor{markergreen}{\bfseries\texttt{[[T]]}}}

\newcommand{\PromptBox}[1]{%
  \begin{tcolorbox}[
    colback=promptblue,
    colframe=promptblue,
    boxrule=0pt,
    arc=5pt,
    left=10pt,
    right=10pt,
    top=10pt,
    bottom=10pt
  ]
    \Trigger\\[-1pt]
    {\bfseries #1}
  \end{tcolorbox}
}

\newcommand{\CaptionedImage}[2]{%
  \begin{tikzpicture}
    \node[inner sep=0pt,anchor=south west] (img) at (0,0)
      {\includegraphics[width=\linewidth]{\detokenize{#1}}};
    \node[
      anchor=south west,
      fill=white,
      fill opacity=0.78,
      text opacity=1,
      rounded corners=1.5pt,
      inner xsep=3pt,
      inner ysep=2pt,
      font=\scriptsize
    ] at ([xshift=2pt,yshift=2pt]img.south west) {#2};
  \end{tikzpicture}%
}

\newcommand{\ImagePair}[2]{%
  \begin{minipage}[t]{0.47\linewidth}
    \CaptionedImage{#1}{Original Image}
  \end{minipage}\hfill
  \begin{minipage}[t]{0.47\linewidth}
    \CaptionedImage{#2}{Image with Visual Anchors}
  \end{minipage}
}

\newcommand{\SectionHead}[1]{%
  \vspace{0.25em}{\fontsize{10}{12}\selectfont\bfseries #1}\par\vspace{0.1em}%
}

\newcommand{\CountingSample}[7]{%
  \begin{CountingBox}{Fine-grained Counting}
    \PromptBox{#1}
    \vspace{0.45em}

    \begin{center}
      \begin{minipage}[t]{\linewidth}
        \ImagePair{#2}{#3}
      \end{minipage}
    \end{center}

    \vspace{0.6em}

    \begin{minipage}[t]{\linewidth}
      \raggedright
      {\fontsize{10}{10}\selectfont\bfseries Anchor Thinking Reasoning \par}
      \vspace{0.25em}
      \SectionHead{1. Intent Analysis}
      #4\par
      \SectionHead{2. Grounded Object Localization}
      #5\par
      \SectionHead{3. Summation}
      #6\par
    \end{minipage}
  \end{CountingBox}
}

\CountingSample
  {How many distinct Brand Logo regions can you find on the page?}
  {0ee7650eb763148b31179393bfcb90735f7ec117981500bfb599afa4cb157251_original.jpg}
  {0ee7650eb763148b31179393bfcb90735f7ec117981500bfb599afa4cb157251_prompt_attempt0_overlay.png}
  {The task is to count every \vpt{} region in the document image by following the supplied reading order.}
  {Across the top row, \vpo{1} encloses the Tourism Malaysia mark, \vpo{2} encloses the Discovery wordmark, \vpo{3} encloses the Holiday Rail emblem, \vpo{4} encloses the KTM wordmark, and \vpo{5} encloses the Firefly airline logo. Below them on the right, \vpo{6} marks the First in Malaysia badge. Farther down, \vpo{7} marks the floral Visit Truly Asia Malaysia 2020 emblem.}
  {That makes 7 in total.}
\clearpage

\CountingSample
  {How many regions on this page should be counted as Photograph?}
  {1d9ce4244544f8d1ff2e9cdeb640b9366a7c92b5708664703386598801e8cbe6_hard_photo_original.jpg}
  {1d9ce4244544f8d1ff2e9cdeb640b9366a7c92b5708664703386598801e8cbe6_prompt_attempt0_overlay.png}
  {The task is to count every \vpt{} region in the document image by following the supplied reading order.}
  {From left to right, \vpo{1} is the large product photograph showing two girls' dresses, \vpo{2} is the upper central photograph of a child modeling an outfit, \vpo{3} is the lower central photograph of another child model, and \vpo{4} is the right-side product photograph showing two clothing sets.
}
  {I count 4.}

\clearpage

\CountingSample
  {Count the Table regions visible on this page.}
  {52f16186a14583663bf78f10ac017089b79a6e0fcf56802d9c7fdbee41fd1ed4_original.jpg}
  {52f16186a14583663bf78f10ac017089b79a6e0fcf56802d9c7fdbee41fd1ed4_prompt_attempt0_overlay.png}
  {The task is to count every \vpt{} region in the document image by following the supplied reading order.}
  {Scanning downward, \vpo{1} is the gridded table for early-childhood teaching qualifications, \vpo{2} is the separate gridded table for primary teaching qualifications, \vpo{3} is the larger gridded table for secondary teaching qualifications, and \vpo{4} is the bottom gridded table for non-teaching education qualifications.}
  {The total is 4.}

\clearpage

\CountingSample
  {What is the total number of Chart / Graph regions in the page image?}
  {ce2381ea4634893e10a8780d17027c21fe56a41c5c5075854973384d8ce10827_medium_chart_original.jpg}
  {ce2381ea4634893e10a8780d17027c21fe56a41c5c5075854973384d8ce10827_prompt_attempt0_overlay.png}
  {The task is to count every \vpt{} region in the document image by following the supplied reading order.
}
  {In the top row of small multiples, \vpo{1} is the Ben Carson time-series panel, \vpo{2} is the Ted Cruz panel, and \vpo{3} is the John Kasich panel. In the second row, \vpo{4} is the Marco Rubio panel and \vpo{5} is the Donald Trump panel; each contains plotted search and news-article trends.
}
  {I count 5 total.}

\clearpage

\CountingSample
  {What count of Brand Logo regions do you see in this page?}{f21496d399bf8082d80a541cc444f11685bbe88e5634b250869bb16af1b69796_hard_brand_logo_original.jpg}{f21496d399bf8082d80a541cc444f11685bbe88e5634b250869bb16af1b69796_prompt_attempt0_overlay.png}
  {The task is to count every \vpt{} region in the document image by following the supplied reading order.}
  {In the first awards row, \vpo{1} is the Los Angeles olive-oil gold-medal seal and \vpo{2} is the NYIOOC gold-award emblem. In the next row, \vpo{3} is the CINVE 2017 seal and \vpo{4} is the Mezquita ORG 2017 seal. The following row contains \vpo{5}, the Zurich Olive Oil Award mark, and \vpo{6}, the TerraOlivo emblem. Continuing downward on the left, \vpo{7} is the Athena competition medal, \vpo{8} is the AVPA Paris award seal, and \vpo{9} is the DOMINA olive-oil competition medal.}
  {That gives me 9.}

\clearpage
\subsection{Examples in \docount Evaluation}
\label{sec:reason_failure_cases}

\definecolor{ink}{HTML}{111111}
\definecolor{muted}{HTML}{666666}
\definecolor{promptblue}{HTML}{EEF5FF}
\definecolor{goodgreen}{HTML}{247A3D}
\definecolor{softgreen}{HTML}{ECF8F0}
\definecolor{badred}{HTML}{B42318}
\definecolor{softred}{HTML}{FFF0F0}

\newcommand{\CorrectWord}{\textcolor{goodgreen}{correct}}
\newcommand{\WrongWord}{\textcolor{badred}{wrong}}

\newtcolorbox{SampleBox}[1]{
  enhanced,
  width=\linewidth,
  colback=white,
  colframe=black,
  boxrule=0.6pt,
  sharp corners,
  left=8pt,
  right=8pt,
  top=9pt,
  bottom=8pt,
  title={#1},
  fonttitle=\bfseries,
  coltitle=white,
  colbacktitle=black,
  boxed title style={sharp corners,boxrule=0.6pt,left=16pt,right=16pt,top=6pt,bottom=5pt,colback=black!60},
  attach boxed title to top left={xshift=-0.6pt,yshift=0.4pt}
}

\newtcolorbox{PromptPanel}{
  colback=promptblue,
  colframe=promptblue,
  boxrule=0pt,
  arc=4pt,
  left=7pt,
  right=7pt,
  top=6pt,
  bottom=6pt
}

\newtcolorbox{ModelCard}[5]{
  enhanced,
  breakable,
  colback=white,
  colframe=#4,
  boxrule=0.45pt,
  arc=2pt,
  left=6pt,
  right=6pt,
  top=5pt,
  bottom=5pt,
  title={#1\hfill \quad predicted #2\quad #3},
  fonttitle=\bfseries\scriptsize,
  coltitle=ink,
  colbacktitle=#5,
  boxed title style={boxrule=0.5pt,colback=#5,sharp corners},
  attach boxed title to top left={xshift=-5pt,yshift=-5pt}
}

\newcommand{\TinyCaption}[1]{\vspace{2pt}{\scriptsize\textcolor{muted}{#1}}}
\newcommand{\ResponseImagePair}[2]{%
  \begin{minipage}[t]{0.492\linewidth}
  \centering
    \includegraphics[width=\linewidth,height=0.30\textheight,keepaspectratio]{\detokenize{#1}}\par
    \TinyCaption{Original}
  \end{minipage}\hfill
  \begin{minipage}[t]{0.492\linewidth}
  \centering
    \includegraphics[width=\linewidth,height=0.30\textheight,keepaspectratio]{\detokenize{#2}}\par
    \TinyCaption{Overlay}
  \end{minipage}%
}

\newcommand{\MetaLine}[2]{{\scriptsize\textcolor{muted}{#1}}\hspace{0.35em}{\small\bfseries #2}}

\begin{SampleBox}{Brand Logo}
  \begin{PromptPanel}
    \MetaLine{Ground Truth}{7}\quad
    \MetaLine{Tag}{brand logo}\par
    \vspace{0.25em}
    {\small\bfseries How many distinct Brand Logo regions can you find on the page?}
  \end{PromptPanel}
  \vspace{0.45em}
  \ResponseImagePair{0ee7650eb763148b31179393bfcb90735f7ec117981500bfb599afa4cb157251_original.jpg}{0ee7650eb763148b31179393bfcb90735f7ec117981500bfb599afa4cb157251_prompt_attempt0_overlay.png}
\end{SampleBox}

\vspace{0.55em}
\begin{ModelCard}{Kimi K2.5 (reasoning on)}{7}{\CorrectWord}{goodgreen}{softgreen}
{\scriptsize\raggedright
Looking at the image, I need to identify distinct visual identity marks that function as brand logos, official program marks, certification seals, or endorsement badges.\par
\par\smallskip
Here is the breakdown:\par
\par\smallskip
1.  **Tourism Malaysia Logo** (top, "Supported by" row): A designed mark with the national flag element and identifying text, functioning as the tourism board's visual identity.\par
2.  **Discovery Logo** (top row, next to Tourism Malaysia): The stylized "D" icon with the "Discovery" wordmark, representing the media/brand partner.\par
3.  **Circular Partner/Destination Logo** (top row, center): A circular badge with a landscape graphic and text, placed as a distinct sponsoring partner identity.\par
4.  **KTM Logo** (top row): The stylized "KTM" lettermark with the signature blue and yellow striped design, identifying the rail service brand.\par
5.  **Firefly Logo** (top right): Consists of the insect icon, the "firefly" stylized wordmark, and the "Beyond Convenience" tagline. Per the definition, a designed mark and its associated brand name appearing together as one unit counts as **one** distinct brand logo.\par
6.  **"FIRST IN MALAYSIA" Badge** (right side, diamond-shaped): This is an endorsement badge/seal making a specific claim about the product's exclusivity. The definition explicitly includes endorsement badges as a type of brand logo.\par
7.  **Visit Truly Asia Malaysia 2020 Logo** (right side, hibiscus and hornbill design): This is the official program/campaign logo for Malaysia's tourism year, functioning as a visual identity mark for that specific government initiative.\par
\par\smallskip
The large "FLYRAIL" text is the product/package name presented as a headline, not a distinct corporate visual identity mark. The "SPECIAL LAUNCH PROMO 15\% OFF" circle is a promotional callout graphic, not a brand logo. The "Book now!" element in the bottom right is a generic call-to-action with a decorative icon, not a brand identifier.\par
\textless{}answer\textgreater{}7\textless{}/answer\textgreater{}\par
}
\end{ModelCard}

\begin{ModelCard}{Qwen3.6 35B A3B (reasoning off)}{7}{\CorrectWord}{goodgreen}{softgreen}
{\scriptsize\raggedright
To determine the number of distinct Brand Logo regions on the page, we need to identify visual identity marks that represent brands, organizations, or official programs — excluding generic decorative icons.\par
\par\smallskip
Let’s go through the image systematically:\par
\par\smallskip
1. **Top right corner group:**\par
\hspace*{1.05em}- “Supported by TOURISM MALAYSIA” — This includes the Tourism Malaysia logo (a stylized flower emblem) with its name. This counts as **one Brand Logo**.\par
\hspace*{1.05em}- “Discovery” — The orange sun-like icon next to the word “Discovery” is the Discovery Channel brand logo. Counts as **one Brand Logo**.\par
\hspace*{1.05em}- “Malaysia Rail” — A circular emblem with a train and text “Malaysia Rail”. This is an official program/brand logo. Counts as **one Brand Logo**.\par
\hspace*{1.05em}- “KTM” — The green and blue stylized logo with “KTM” text. This is the Kuala Lumpur Transport Corporation brand logo. Counts as **one Brand Logo**.\par
\hspace*{1.05em}- “firefly Beyond Convenience” — The orange bird-like icon with “firefly” text. This is Firefly airline’s brand logo. Counts as **one Brand Logo**.\par
\par\smallskip
\hspace*{1.05em}→ That’s 5 logos in the top right.\par
\par\smallskip
2. **Right side, middle section:**\par
\hspace*{1.05em}- “FIRST IN MALAYSIA” — This is a badge/seal indicating an official certification or program. It has a distinct design (blue diamond shape with white text). Counts as **one Brand Logo** (official program mark).\par
\hspace*{1.05em}- Below it: “Visit Malaysia 2020” — This is the official tourism campaign logo for 2020, featuring a bird and hibiscus. Clearly a brand/logo for a national tourism initiative. Counts as **one Brand Logo**.\par
\par\smallskip
\hspace*{1.05em}→ That’s 2 more.\par
\par\smallskip
3. **Bottom right corner:**\par
\hspace*{1.05em}- “Book now!” with a stylized sail/boat icon — This appears to be a call-to-action button graphic, not a brand logo. Exclude.\par
\par\smallskip
4. **Airplane in sky:**\par
\hspace*{1.05em}- The airplane has “firefly” written on its side — but this is part of the aircraft livery, not a separate logo instance. Since we already counted the Firefly logo above, we don’t double-count.\par
\par\smallskip
5. **Train:**\par
\hspace*{1.05em}- No visible brand logo on the train itself (just locomotive number “2521”), so no additional logo here.\par
\par\smallskip
6. **Left side:**\par
\hspace*{1.05em}- “FLYRAIL” — This is the name of the package, styled in bold orange letters. However, it’s presented as a product/package name, not necessarily a standalone brand logo. Given the context (“3D2N Holiday Package”), it’s likely a marketing title rather than a registered brand mark. We’ll exclude it unless clearly branded — but note: if “Flyrail” is a co-branded service, it might qualify. However, since it’s not accompanied by a unique emblem or official branding signal (like a seal or trademark symbol), and it’s more of a promotional headline, we will **not count it** as a distinct brand logo per the definition.\par
\par\smallskip
7. **Circular orange promo badge:**\par
\hspace*{1.05em}- “SPECIAL LAUNCH PROMO 15\% OFF Limited time only.” — This is a promotional graphic, not a brand logo. Exclude.\par
\par\smallskip
8. **Small airplane on runway:**\par
\hspace*{1.05em}- No branding visible. Exclude.\par
\par\smallskip
9. **Bottom left corner:**\par
\hspace*{1.05em}- “*Terms \& Conditions apply.” — Text only, not a logo. Exclude.\par
\par\smallskip
Now, summing up the confirmed Brand Logos:\par
\par\smallskip
- Tourism Malaysia → 1\par
- Discovery → 1\par
- Malaysia Rail → 1\par
- KTM → 1\par
- Firefly → 1\par
- FIRST IN MALAYSIA badge → 1\par
- Visit Malaysia 2020 → 1\par
\par\smallskip
Total = **7**\par
\par\smallskip
\textless{}answer\textgreater{}7\textless{}/answer\textgreater{}\par
}
\end{ModelCard}

\begin{ModelCard}{GPT-5.5 (reasoning off)}{11}{\WrongWord}{badred}{softred}
{\scriptsize\raggedright
I identified the visible Brand Logo regions as:\par
\par\smallskip
1. Tourism Malaysia logo  \par
2. Discovery logo  \par
3. Malaysia Rail logo  \par
4. KTM logo  \par
5. Firefly logo at top right  \par
6. FlyRail stylized wordmark  \par
7. “First in Malaysia” badge  \par
8. Visit Truly Asia Malaysia 2020 logo  \par
9. Firefly branding on the aircraft tail  \par
10. Fireflyz.com.my branding on the aircraft fuselage  \par
11. Firefly-style logomark in the bottom-right orange section  \par
\par\smallskip
\textless{}answer\textgreater{}11\textless{}/answer\textgreater{}\par
}
\end{ModelCard}

\begin{ModelCard}{Claude Sonnet 4.5 (reasoning on)}{8}{\WrongWord}{badred}{softred}
{\scriptsize\raggedright
\# Detailed Analysis of Brand Logos\par
\par\smallskip
Let me systematically examine the image to identify all distinct brand logos:\par
\par\smallskip
\#\# Top Section - Sponsor Logos (Left to Right):\par
1. **Tourism Malaysia** - Official tourism organization logo with "Supported by TOURISM MALAYSIA" text\par
2. **Recovery** - Orange/yellow logo with arrow design (second sponsor)\par
3. **Circular green/teal logo** - Appears to be an environmental or certification organization (third sponsor)\par
4. **KTMB** - Blue and yellow railway company logo (Keretapi Tanah Melayu Berhad)\par
5. **firefly** - Orange/yellow airline logo with "Beyond Convenience" tagline\par
\par\smallskip
\#\# Main Content Area:\par
6. **FLYRAIL** - The large orange stylized wordmark serves as the product/service brand identifier for this fly-and-rail package\par
\par\smallskip
\#\# Right Side Elements:\par
7. **"FIRST IN MALAYSIA"** - Diamond-shaped badge/seal that functions as a program identifier and promotional emblem\par
8. **Visit Terengganu Malaysia 2021** - Floral hibiscus design logo for the tourism campaign\par
\par\smallskip
\#\# Excluded Elements:\par
- "Book now!" with arrow design - This is a call-to-action graphic element, not a brand logo\par
- The airplane and train images themselves - These are photographs, not logos\par
- Numerical identifiers on vehicles - Not brand logos\par
\par\smallskip
Each identified element meets the definition criteria as it serves as a visual identity mark for a brand, organization, product, service, or official program.\par
\par\smallskip
\textless{}answer\textgreater{}8\textless{}/answer\textgreater{}\par
}
\end{ModelCard}

\clearpage

\begin{SampleBox}{Photograph}
  \begin{PromptPanel}
    \MetaLine{Ground Truth}{4}\quad
    \MetaLine{Tag}{photograph}\par
    \vspace{0.25em}
    {\small\bfseries How many regions on this page should be counted as Photograph?}
  \end{PromptPanel}
  \vspace{0.45em}
  \ResponseImagePair{1d9ce4244544f8d1ff2e9cdeb640b9366a7c92b5708664703386598801e8cbe6_hard_photo_original.jpg}{1d9ce4244544f8d1ff2e9cdeb640b9366a7c92b5708664703386598801e8cbe6_prompt_attempt0_overlay.png}
\end{SampleBox}

\vspace{0.55em}
\begin{ModelCard}{Kimi K2.5 (reasoning on)}{6}{\WrongWord}{badred}{softred}
{\scriptsize\raggedright
Looking at this JCPenney catalog page, I need to identify all regions that are realistic photographs of real-world products or people, excluding text, illustrations, charts, and flat graphics.\par
\par\smallskip
Scanning the page layout, I can identify the following distinct photographic regions:\par
\par\smallskip
- **Region A (top left):** A realistic product photograph of a red plaid dress with a white lace collar, showing natural fabric texture, folds, and shadows.\par
- **Region B (bottom left):** A realistic product photograph of a blue jumper and white blouse set, again showing natural textile textures and dimensional draping.\par
- **Region C (center top):** A realistic photograph of a child model wearing a light blue chambray dress with a dark blue skirt, showing natural skin tones, hair texture, and fabric details.\par
- **Region D (center bottom):** A realistic photograph of a child model wearing a purple and black striped two-piece outfit, showing natural photographic qualities like hair detail and fabric sheen.\par
- **Region E (top right):** A realistic product photograph of a cream-colored top and red pants arranged on a flat surface, showing knit texture and garment shaping.\par
- **Region F (bottom right):** A realistic product photograph of a red, white, and blue dress with vest effect, laid flat or on a form, displaying natural fabric textures and stitching details.\par
\par\smallskip
The remaining areas consist of ordinary text (product descriptions, prices, item numbers), typographic headings, and catalog page background—none of which qualify as photographs under the given definition.\par
\par\smallskip
There are no illustrations, abstract graphics, or charts present; all six clothing/item displays are naturalistic photographs.\par
\par\smallskip
\textless{}answer\textgreater{}6\textless{}/answer\textgreater{}\par
}
\end{ModelCard}

\begin{ModelCard}{Qwen3.6 35B A3B (reasoning off)}{6}{\WrongWord}{badred}{softred}
{\scriptsize\raggedright
To determine how many regions on this page should be counted as Photograph, we must apply the given definition:\par
\par\smallskip
\textgreater{} Realistic or naturalistic image content depicting real-world people, objects, places, products, scenes, textures, and so on. Exclude flat illustrations, icons, abstract graphics, patterns, charts, tables, and ordinary text.\par
\par\smallskip
The page is a catalog spread from JC Penney (1970s), featuring children’s clothing. It contains both textual descriptions and visual images of garments and models.\par
\par\smallskip
Let’s break down the page into distinct visual regions:\par
\par\smallskip
1. **Region A** – Top left: Shows a red plaid dress with white collar and apron-style front. The garment is photographed realistically against a neutral background. This qualifies as a photograph.\par
\par\smallskip
2. **Region B** – Bottom left: Shows a dark blue polka-dot jumper with white blouse underneath, worn by a child (partially visible). The image is a realistic product shot — qualifies as a photograph.\par
\par\smallskip
3. **Region C** – Top middle: Features a girl modeling a denim-style dress with ruffled hem. This is a full-body photo of a real person wearing the item — qualifies as a photograph.\par
\par\smallskip
4. **Region D** – Bottom middle: Shows another girl modeling a striped dress with ruffled sleeves. Again, a realistic photo of a model in the garment — qualifies as a photograph.\par
\par\smallskip
5. **Region E** – Top right: Displays a two-piece set (top and pants) laid out flat but still presented as a photographic product shot — not illustrated, but photographed. Qualifies.\par
\par\smallskip
6. **Region F** – Bottom right: Shows a child’s dress with vest-style bodice and floral skirt, again photographed as a product image — qualifies.\par
\par\smallskip
Additionally, note that the entire layout includes photographs of actual garments and models — these are all realistic depictions of physical products and people. There are no flat illustrations, icons, charts, or abstract graphics being used to represent the products — everything shown is either a photo of a real garment or a child modeling it.\par
\par\smallskip
Therefore, each of the six labeled regions (A through F) represents a distinct photographic region.\par
\par\smallskip
There is also the JC Penney logo and page number at bottom left — but those are small textual/logo elements and do not constitute “regions” under the context of counting photographic content for product display. The question implies counting major image regions showcasing products/models.\par
\par\smallskip
Thus, we count 6 photographic regions.\par
\par\smallskip
\textless{}answer\textgreater{}6\textless{}/answer\textgreater{}\par
}
\end{ModelCard}

\begin{ModelCard}{GPT-5.5 (reasoning off)}{6}{\WrongWord}{badred}{softred}
{\scriptsize\raggedright
The page is a catalog layout with text descriptions and several product images. The regions that qualify as **Photograph** are the realistic images of clothing/products and children modeling outfits.\par
\par\smallskip
Countable photograph regions:\par
\par\smallskip
1. Red plaid dress labeled A  \par
2. Jumper-and-blouse set labeled B  \par
3. Child modeling chambray dress labeled C  \par
4. Child modeling striped skirt set labeled D  \par
5. Pants and top set labeled E  \par
6. Dress with vest-effect labeled F  \par
\par\smallskip
Text blocks, labels, and background page areas are not counted.\par
\par\smallskip
\textless{}answer\textgreater{}6\textless{}/answer\textgreater{}\par
}
\end{ModelCard}

\begin{ModelCard}{Claude Sonnet 4.5 (reasoning on)}{4}{\CorrectWord}{goodgreen}{softgreen}
{\scriptsize\raggedright
Looking at this catalog page carefully, I need to identify distinct photographic regions based on the definition provided.\par
\par\smallskip
Let me analyze each section of the page:\par
\par\smallskip
**Left Column:**\par
- Shows dress A (red plaid dress with white trim) at the top\par
- Shows dress B (navy sparkly dress with white blouse) below it\par
- These appear to be on the same continuous tan/beige background without a dividing line between them\par
- This appears to be **1 photographic region** containing both items\par
\par\smallskip
**Center Column:**\par
- Shows a model wearing dress C (blue chambray dress) in the top photo\par
- Shows a model wearing dress D (purple striped dress) in the bottom photo  \par
- There is a clear horizontal dividing line between these two photographs\par
- These are **2 separate photographic regions**\par
\par\smallskip
**Right Column:**\par
- Shows outfit E (red pants and cream top) at the top\par
- Shows dress F (red, white, and blue dress) below it\par
- These appear to be on the same continuous tan/beige background without a dividing line between them\par
- This appears to be **1 photographic region** containing both items\par
\par\smallskip
All of these images are photographs of real products and real people (the models), fitting the definition of "realistic or naturalistic image content depicting real-world people, objects, places, products."\par
\par\smallskip
The text descriptions at the top are ordinary text and should not be counted.\par
\par\smallskip
**Total count:** 1 (left column) + 2 (center photos) + 1 (right column) = 4 photographic regions\par
\par\smallskip
\textless{}answer\textgreater{}4\textless{}/answer\textgreater{}\par
}
\end{ModelCard}

\clearpage

\begin{SampleBox}{Table}
  \begin{PromptPanel}
    \MetaLine{Ground Truth}{4}\quad
    \MetaLine{Tag}{table}\par
    \vspace{0.25em}
    {\small\bfseries Count the Table regions visible on this page.}
  \end{PromptPanel}
  \vspace{0.45em}
  \ResponseImagePair{52f16186a14583663bf78f10ac017089b79a6e0fcf56802d9c7fdbee41fd1ed4_original.jpg}{52f16186a14583663bf78f10ac017089b79a6e0fcf56802d9c7fdbee41fd1ed4_prompt_attempt0_overlay.png}
\end{SampleBox}

\vspace{0.55em}
\begin{ModelCard}{Kimi K2.5 (reasoning on)}{4}{\CorrectWord}{goodgreen}{softgreen}
{\scriptsize\raggedright
Looking at the page, I can identify distinct arrangements of data organized into rows and columns for the purpose of comparing qualifications across institutions and durations.\par
\par\smallskip
**Detailed rationale:**\par
- The first table covers **Teaching (Early Childhood Education, ages 0-5)** with 4 rows listing specific bachelor's degrees, their respective institutions, and duration (3 years).\par
- The second table covers **Teaching (Primary, ages 5-13)** with 2 rows listing pasifika and general primary teaching degrees, institutions, and duration (3 years).\par
- The third table covers **Teaching (Secondary, ages 13-18)** with 4 rows listing various secondary teaching qualifications, institutions, and duration (4 years).\par
- The fourth table covers **Education (non-teaching)** with 3 rows listing non-teaching education degrees, institutions, and duration (3 years).\par
\par\smallskip
Each of these four sections functions as an independent data table with its own header row and aligned columns for qualification name, institution name, and length of study. They are separated by whitespace and have distinct categorical groupings.\par
\par\smallskip
\textless{}answer\textgreater{}4\textless{}/answer\textgreater{}\par
}
\end{ModelCard}

\begin{ModelCard}{Qwen3.6 35B A3B (reasoning off)}{1}{\WrongWord}{badred}{softred}
{\scriptsize\raggedright
The document contains one large, structured table that organizes information about undergraduate qualifications relevant to education in New Zealand. This table is divided into four main sections (or row groups), each corresponding to a different teaching qualification area:\par
\par\smallskip
1. Teaching (Early Childhood Education, ages 0–5)\par
2. Teaching (Primary, ages 5–13)\par
3. Teaching (Secondary, ages 13–18)\par
4. Education (non-teaching)\par
\par\smallskip
Each section consists of multiple rows listing specific degrees, the institutions offering them, and the duration (years). The entire structure functions as a single cohesive table with clear columnar alignment: columns for qualification type, degree name, university/institution, and duration.\par
\par\smallskip
Although there are visual separators between the four major sections, they are part of the same overarching table — not separate tables. There are no other distinct tables on the page (e.g., no menus, product grids, or side tables).\par
\par\smallskip
Therefore, only one table region is visible.\par
\par\smallskip
\textless{}answer\textgreater{}1\textless{}/answer\textgreater{}\par
}
\end{ModelCard}

\begin{ModelCard}{GPT-5.5 (reasoning off)}{4}{\CorrectWord}{goodgreen}{softgreen}
{\scriptsize\raggedright
The page contains four distinct grid-like table blocks, each separated by whitespace and organized with rows and columns:\par
\par\smallskip
1. Teaching — Early Childhood Education\par
2. Teaching — Primary\par
3. Teaching — Secondary\par
4. Education — Non-teaching\par
\par\smallskip
Therefore, the number of visible Table regions is:\par
\par\smallskip
\textless{}answer\textgreater{}4\textless{}/answer\textgreater{}\par
}
\end{ModelCard}

\begin{ModelCard}{Claude Sonnet 4.5 (reasoning on)}{4}{\CorrectWord}{goodgreen}{softgreen}
{\scriptsize\raggedright
Looking at this document about undergraduate qualifications relevant to Education, I need to identify all table regions that display information in a grid-like arrangement with rows and columns.\par
\par\smallskip
Let me examine the page systematically:\par
\par\smallskip
1. **First table**: "Teaching (Early Childhood Education, ages 0-5)" - This section contains a structured grid with columns showing qualification names, institutions, and duration. The information is organized in cells with visible borders.\par
\par\smallskip
2. **Second table**: "Teaching (Primary, ages 5-13)" - This is a separate table below the first one, with the same columnar structure showing Bachelor degrees, institutions, and years.\par
\par\smallskip
3. **Third table**: "Teaching (Secondary, ages 13-18)" - Another distinct table section showing secondary education qualifications with institutions and 4-year durations organized in rows and columns.\par
\par\smallskip
4. **Fourth table**: "Education (non-teaching)" - The final table section at the bottom showing non-teaching education qualifications with their respective institutions and durations.\par
\par\smallskip
Each of these sections is clearly separated and functions as an independent table with:\par
- A left column identifying the category\par
- Middle columns listing qualifications\par
- Columns showing institutions\par
- Right columns showing duration in years\par
\par\smallskip
All have visible grid lines forming cells for data organization and comparison purposes.\par
\par\smallskip
\textless{}answer\textgreater{}4\textless{}/answer\textgreater{}\par
}
\end{ModelCard}

\clearpage

\begin{SampleBox}{Chart / Graph}
  \begin{PromptPanel}
    \MetaLine{Ground Truth}{5}\quad
    \MetaLine{Tag}{chart / graph}\par
    \vspace{0.25em}
    {\small\bfseries What is the total number of Chart / Graph regions in the page image?}
  \end{PromptPanel}
  \vspace{0.45em}
  \ResponseImagePair{ce2381ea4634893e10a8780d17027c21fe56a41c5c5075854973384d8ce10827_medium_chart_original.jpg}{ce2381ea4634893e10a8780d17027c21fe56a41c5c5075854973384d8ce10827_prompt_attempt0_overlay.png}
\end{SampleBox}

\vspace{0.55em}
\begin{ModelCard}{Kimi K2.5 (reasoning on)}{6}{\WrongWord}{badred}{softred}
{\scriptsize\raggedright
The image contains six distinct chart/graph regions. Five of these are the self-contained small-multiple time-series panels, one for each presidential candidate (Ben Carson, Ted Cruz, John Kasich, Marco Rubio, and Donald Trump), where each panel independently plots its own dataset of relative Google searches and news articles over time. The sixth is the separate inset sparkline chart in the top-right corner, which visually encodes the two metrics (Articles on Google News and Google Searches) as a self-contained quantitative graphic distinct from the five candidate panels.\par
\par\smallskip
\textless{}answer\textgreater{}6\textless{}/answer\textgreater{}\par
}
\end{ModelCard}

\begin{ModelCard}{Qwen3.6 35B A3B (reasoning off)}{6}{\WrongWord}{badred}{softred}
{\scriptsize\raggedright
Let’s carefully analyze the image to count the number of Chart / Graph regions.\par
\par\smallskip
The image is a data visualization from FiveThirtyEight, titled:  \par
**“Trump continues to dominate both news coverage and Google searches”**\par
\par\smallskip
It includes:\par
\par\smallskip
1. A **main header section** with text and a small legend/key graphic in the top right corner — this legend shows two lines (red for “Articles on Google News”, black for “Google Searches”) over a simplified x-axis. This is a **small illustrative chart/graph** used as a key or example — it has axes, plotted lines, and data representation. So we count this as **1 chart**.\par
\par\smallskip
2. Then there are **five distinct panels**, each showing time-series graphs for a different candidate:\par
\hspace*{1.05em}- Ben Carson\par
\hspace*{1.05em}- Ted Cruz\par
\hspace*{1.05em}- John Kasich\par
\hspace*{1.05em}- Marco Rubio\par
\hspace*{1.05em}- Donald Trump\par
\par\smallskip
Each of these five panels:\par
- Has its own title (candidate name)\par
- Contains a line graph (black = Google Searches, red = Articles on Google News)\par
- Has its own x-axis (time from AUG ’15 to FEB ’16)\par
- Has its own y-axis labeled “MAX” and “0”\par
- Contains a portrait illustration of the candidate (not part of the data, but the graph itself is quantitative)\par
\par\smallskip
These are **independent visualizations** — each represents a separate dataset (for each candidate), with their own coordinate layout and analytical result. They are presented as “small multiples” — repeated structure but each panel is self-contained and presents its own data display.\par
\par\smallskip
According to the definition:  \par
\textgreater{} “Count small multiples as multiple charts when each repeated panel is a self-contained visualization with its own data display.”\par
\par\smallskip
So we count each of the five candidate panels as **separate charts**.\par
\par\smallskip
Thus:\par
- 1 small legend/chart in top right\par
- 5 candidate-specific charts\par
\par\smallskip
Total = 1 + 5 = **6**\par
\par\smallskip
Note: The entire composite figure might be considered one infographic, but the question asks for “Chart / Graph regions” — meaning individual graphical representations of data. Each panel qualifies as such.\par
\par\smallskip
\textless{}answer\textgreater{}6\textless{}/answer\textgreater{}\par
}
\end{ModelCard}

\begin{ModelCard}{GPT-5.5 (reasoning off)}{5}{\CorrectWord}{goodgreen}{softgreen}
{\scriptsize\raggedright
The image contains a set of small-multiple time-series visualizations for presidential candidates. Each candidate panel shows quantitative data over time using a red filled area and a gray line, so each panel is a distinct chart/graph.\par
\par\smallskip
The self-contained chart regions are:\par
\par\smallskip
1. Ben Carson  \par
2. Ted Cruz  \par
3. John Kasich  \par
4. Marco Rubio  \par
5. Donald Trump  \par
\par\smallskip
The small graphic in the upper-right appears to function as a legend/key explaining the red area and gray line encodings, rather than a separate independent chart.\par
\par\smallskip
\textless{}answer\textgreater{}5\textless{}/answer\textgreater{}\par
}
\end{ModelCard}

\begin{ModelCard}{Claude Sonnet 4.5 (reasoning on)}{6}{\WrongWord}{badred}{softred}
{\scriptsize\raggedright
Let me analyze this image systematically to count all chart/graph regions.\par
\par\smallskip
**Analysis:**\par
\par\smallskip
1. **Top Right Corner**: There is a small chart labeled "ARTICLES ON GOOGLE NEWS" with "GOOGLE SEARCHES" indicator. This shows two overlaid time series (one appears to be searches, one appears to be articles) - this is **1 chart**.\par
\par\smallskip
2. **Main Section - Small Multiples Grid**: The image displays 5 individual time-series charts arranged in a grid, each showing data for a different political candidate:\par
\hspace*{1.05em}- **Ben Carson** (top row, left) - 1 chart\par
\hspace*{1.05em}- **Ted Cruz** (top row, center) - 1 chart  \par
\hspace*{1.05em}- **John Kasich** (top row, right) - 1 chart\par
\hspace*{1.05em}- **Marco Rubio** (bottom row, left) - 1 chart\par
\hspace*{1.05em}- **Donald Trump** (bottom row, right) - 1 chart\par
\par\smallskip
Each of these 5 panels is a self-contained visualization with:\par
- Its own independent coordinate system and axes\par
- Its own time range (AUG '15 to FEB '16)\par
- Two overlaid data series (Google searches in dark line, articles on Google News in red/coral fill)\par
- Its own data specific to that candidate\par
\par\smallskip
According to the definition, these small multiples should be counted as separate charts since "each repeated panel is a self-contained visualization with its own data display."\par
\par\smallskip
**Total Count**: 1 (reference chart) + 5 (candidate charts) = **6 charts**\par
\par\smallskip
\textless{}answer\textgreater{}6\textless{}/answer\textgreater{}\par
}
\end{ModelCard}

\clearpage

\begin{SampleBox}{Brand Logo}
  \begin{PromptPanel}
    \MetaLine{Ground Truth}{9}\quad
    \MetaLine{Tag}{brand logo}\par
    \vspace{0.25em}
    {\small\bfseries What count of Brand Logo regions do you see in this page?}
  \end{PromptPanel}
  \vspace{0.45em}
  \ResponseImagePair{f21496d399bf8082d80a541cc444f11685bbe88e5634b250869bb16af1b69796_hard_brand_logo_original.jpg}{f21496d399bf8082d80a541cc444f11685bbe88e5634b250869bb16af1b69796_prompt_attempt0_overlay.png}
\end{SampleBox}

\vspace{0.55em}
\begin{ModelCard}{Kimi K2.5 (reasoning on)}{12}{\WrongWord}{badred}{softred}
{\scriptsize\raggedright
Rationale: The image displays an awards page with 12 distinct award entries, each accompanied by a unique visual identity mark. These marks include circular seals, rectangular certificates, and badges representing specific olive oil competitions and certifications (e.g., L.A. Olive Oil Awards, CINVE, Zurich, Athena, AVPA Paris, DOMINA, NYIOOC, Mezquita, TerraOlivo, Les Olivalies, Expo Azeite Brazil, and Japan Olive). Each of these 12 marks functions as a brand identifier, emblem, certification seal, or award badge for its respective organization or program. There are no other brand marks present, and the background sunflower is purely decorative.\par
\par\smallskip
\textless{}answer\textgreater{}12\textless{}/answer\textgreater{}\par
}
\end{ModelCard}

\begin{ModelCard}{Qwen3.6 35B A3B (reasoning off)}{12}{\WrongWord}{badred}{softred}
{\scriptsize\raggedright
Let’s carefully analyze the image to count the number of Brand Logo regions.\par
\par\smallskip
The image is titled “國際橄欖油競賽獲獎 (Awards)” and lists 12 international olive oil competition awards, each accompanied by a visual award emblem or logo on the left side of the text description.\par
\par\smallskip
We are to count **Brand Logo regions** — defined as visual identity marks used to identify brands, organizations, products, services, certifications, standards, awards, or official programs. This includes emblems, seals, badges, stylized wordmarks, etc. We exclude generic decorative icons.\par
\par\smallskip
Each award entry has one distinct emblem/logo representing that specific competition or award body. Let’s go through them:\par
\par\smallskip
Left column (6 entries):\par
1. L.A. Olive Oil Awards (USA) — Gold Medal → circular gold seal with “LOS ANGELES” and olive branch → 1 Brand Logo\par
2. CINVE Awards (Spain) — Gold Medal → black and gold emblem with “CINVE 2017” → 1 Brand Logo\par
3. Zurich Olive Oil Awards (Switzerland) — Gold Medal → white and gold circular emblem with “OLIVE OIL AWARD ZURICH 2017” → 1 Brand Logo\par
4. Athena Contest (Greece) — Gold Medal → dark circular emblem with “ATHENA CONTEST 2017” → 1 Brand Logo\par
5. AVPA Paris (France) — Gourmet Gold Medal → golden circular emblem with star design and “AVPA PARIS” → 1 Brand Logo\par
6. DOMINA Olive Oil Competition (Italy) — Gold Medal → golden circular emblem with “DOMINA IOOC Gold Medal 2017” → 1 Brand Logo\par
\par\smallskip
Right column (6 entries):\par
7. NYIOOC 2017 (USA) — Gold Medal → black and gold laurel wreath emblem with “GOLD AWARD 2017” → 1 Brand Logo\par
8. Mezquita Awards (Spain) — Gold Medal → greenish-gold circular emblem with “ORO 2017” and “MEZQUITA” → 1 Brand Logo\par
9. TerraOlivo 2017 (Israel) — Grand Prestige Gold → circular emblem with olive oil pitcher and “TERRAOLIVO ISRAEL” → 1 Brand Logo\par
10. Les Olivalies (France) — Silver Medal → rectangular certificate-style badge with “Les Olivalies Argente” → 1 Brand Logo (functions as official award seal)\par
11. Expo Azeite Brazil (Brazil) — Silver Medal → rectangular certificate-style badge with “Expo Azeite IOOC 2017” → 1 Brand Logo\par
12. Japan Olive (Japan) — Silver Medal → vertical silver badge with “SILVER” and Japanese characters → 1 Brand Logo\par
\par\smallskip
All 12 emblems are distinct visual identity marks associated with specific award competitions or organizations. None are generic decorative icons — they all serve as official seals or badges for those programs.\par
\par\smallskip
Thus, total Brand Logo regions = 12.\par
\par\smallskip
\textless{}answer\textgreater{}12\textless{}/answer\textgreater{}\par
}
\end{ModelCard}

\begin{ModelCard}{GPT-5.5 (reasoning off)}{12}{\WrongWord}{badred}{softred}
{\scriptsize\raggedright
I counted the distinct award/certification badge or seal images shown on the page, treating each medal, badge, or certificate-style award mark as one Brand Logo region.\par
\par\smallskip
- Left column: 6 award logo/badge regions\par
- Right column: 6 award logo/badge/certificate regions\par
\par\smallskip
I did not count the plain descriptive text lines or the page title as separate logos.\par
\par\smallskip
\textless{}answer\textgreater{}12\textless{}/answer\textgreater{}\par
}
\end{ModelCard}

\begin{ModelCard}{Claude Sonnet 4.5 (reasoning on)}{12}{\WrongWord}{badred}{softred}
{\scriptsize\raggedright
Let me carefully examine each award shown in this image to count the distinct Brand Logos.\par
\par\smallskip
**Left Column:**\par
1. **L.A. Olive Oil Awards** - Circular gold seal/badge\par
2. **CINVE Awards** - Black circular medal with "CINVE 2017" branding\par
3. **Zurich Olive Oil Awards** - Circular medal with olive branch design and "GOLD" label\par
4. **Athena Contest** - Circular black and gold medal badge\par
5. **AVPA Paris** - Circular gold medal with decorative sunburst design\par
6. **DOMINA Olive Oil Competition** - Circular gold medal with "IOOC Gold Medal" branding\par
\par\smallskip
**Right Column:**\par
7. **NYIOOC 2017** - Circular black badge with "GOLD AWARD 2017" text\par
8. **Mezquita Awards** - Circular gold medal with "ORO" designation\par
9. **TerraOlivo 2017** - Circular olive-green colored badge\par
10. **Les Olivalies** - Certificate/document displaying the award brand identity\par
11. **Expo Azeite Brazil** - Certificate/document displaying the award brand identity\par
12. **Japan Olive** - Rectangular badge/seal showing the award branding\par
\par\smallskip
Each of these represents a distinct visual identity mark for a different olive oil competition or award organization. Whether displayed as a medal, badge, seal, or certificate, each serves as the brand logo identifying that particular award program.\par
\par\smallskip
\textless{}answer\textgreater{}12\textless{}/answer\textgreater{}\par
}
\end{ModelCard}

\end{CJK*}
\end{document}